\documentclass[]{style}
\usepackage{titletoc}
\usepackage[toc,page,header]{appendix}

\usepackage{minitoc}
\usepackage{natbib}
\usepackage{CJKutf8}

\usepackage{xargs}  

\usepackage{todonotes}  

\usepackage{multirow}

\usepackage{cleveref}

\usepackage{amsmath}
\usepackage{dsfont}

\usepackage{svg}

\usepackage{mathrsfs}
\usepackage{adjustbox}
\usepackage{multirow}
\usepackage{multicol}
\usepackage{tcolorbox}
\usepackage{changepage}
\usepackage{enumitem}
\usepackage{graphicx}
\usepackage{amssymb}
\usepackage{xcolor}
\usepackage{float}
\usepackage{multirow}
\usepackage{threeparttable}
\usepackage{graphicx}
\usepackage{subcaption}
\usepackage{algorithm}
\usepackage{algpseudocode}
\usepackage{wrapfig}
\usepackage[table]{xcolor}  % loads xcolor with table support
\usepackage{colortbl}       % gives \columncolor, \rowcolor
\usepackage{tabularx} % Add to your preamble
\usepackage{makecell} % Add to your preamble
\usepackage[dvipsnames]{xcolor}

\newcolumntype{g}{>{\columncolor{gray!10}}c} % gray background

\definecolor{catgray}{gray}{0.9}
\definecolor{skyblue}{rgb}{0.53,0.81,0.92} % sky blue

\colorlet{skyblue!30}{skyblue!30!white} % 30% skyblue, 70% white

\definecolor{customblue}{RGB}{70,130,180}  % This is equivalent to rgb(70,130,180)

\newtcolorbox{evolbox}[2][]{%
  enhanced,
  colframe=customblue,
  colback=white,
  coltitle=white,
  rounded corners,
  boxrule=1pt,
  titlerule=0pt,
  toptitle=1mm,
  bottomtitle=1mm,
  fonttitle=\bfseries,
  width=#2\textwidth, % This takes the second parameter as the width fraction
  #1
}
\usepackage{url}

\PassOptionsToPackage{table,xcdraw}{xcolor}
\usepackage{placeins}
\usepackage{pifont}

\definecolor{RowBlue}{HTML}{E9F2FB}
\definecolor{RowRed}{HTML}{F9EAEA}
\definecolor{Top1}{HTML}{50DB4B} % 深绿
\definecolor{Top2}{HTML}{A5FFA2} % 中绿
\definecolor{Top3}{HTML}{D9FFD9} % 浅绿
\definecolor{Sub1}{HTML}{C7DBF2}
\definecolor{Sub2}{HTML}{E4E4E4}
\newcommand{\cmark}{\textcolor{green!60!black}{\checkmark}}
\newcommand{\xmark}{\textcolor{red!70!black}{\ding{55}}}

\newcommand{\myparagraph}[1]{\textbf{#1}\hspace{1.8ex}}

\renewcommand{\emph}[1]{\textit{#1}}

\title{Rethinking Video Generation Model for the Embodied World}

\author{
  Yufan Deng\texorpdfstring{$^{1,2*}$}{} \hspace{0.1cm}
  Zilin Pan\texorpdfstring{$^{1*}$}{} \hspace{0.1cm}
  Hongyu Zhang\texorpdfstring{$^{1*}$}{} \hspace{0.1cm}
  Xiaojie Li\texorpdfstring{$^{2}$}{} \hspace{0.1cm}
  Ruoqing Hu\texorpdfstring{$^{2}$}{} \hspace{0.1cm} \\
  Yufei Ding\texorpdfstring{$^{1}$}{} \hspace{0.1cm}
  Yiming Zou\texorpdfstring{$^{1}$}{} \hspace{0.1cm}
  Yan Zeng\texorpdfstring{$^{2}$}{} \hspace{0.1cm}
  Daquan Zhou\texorpdfstring{$^{1}$}{}
}

\affiliation[1]{Peking University}
\affiliation[2]{ByteDance Seed}
\abstract{
Video generation models have significantly advanced embodied intelligence, unlocking new possibilities for generating diverse robot data that capture perception, reasoning, and action in the physical world. 
However, synthesizing high-quality videos that accurately reflect real-world robotic interactions remains challenging, and the lack of a standardized benchmark limits fair comparisons and progress.
To address this gap, we introduce a comprehensive robotics benchmark, \textbf{\textcolor{seedblue}{RBench}}, designed to evaluate robot-oriented video generation across five task domains and four distinct embodiments.
It assesses both task-level correctness and visual fidelity through reproducible sub-metrics, including structural consistency, physical plausibility, and action completeness.
Evaluation of 25 representative models highlights significant deficiencies in generating physically realistic robot behaviors. Furthermore, the benchmark achieves a Spearman correlation coefficient of 0.96 with human evaluations, validating its effectiveness.
While RBench provides the necessary lens to identify these deficiencies, achieving physical realism requires moving beyond evaluation to address the critical shortage of high-quality training data. Driven by these insights, we introduce a refined four-stage data pipeline, resulting in \textbf{\textcolor{seedblue}{RoVid-X}}, the largest open-source robotic dataset for video generation with 4 million annotated video clips, covering thousands of tasks and enriched with comprehensive physical property annotations.
Collectively, this synergistic ecosystem of evaluation and data establishes a robust foundation for rigorous assessment and scalable training of video models, accelerating the evolution of embodied AI toward general intelligence.
}

\checkdata[
\raisebox{-0.2em}{\includegraphics[width=0.025\linewidth]{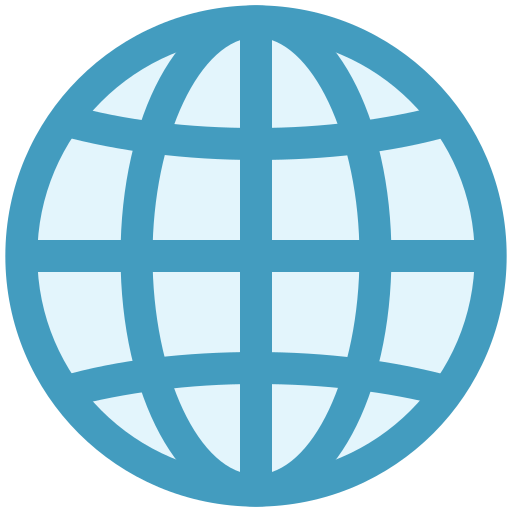}}~~Project Page]{\href{https://dagroup-pku.github.io/ReVidgen.github.io/}{\texttt{https://dagroup-pku.github.io/ReVidgen.github.io/}}
\\[-1.5ex]}

\checkdata[
\raisebox{-0.2em}{\includegraphics[width=0.025\linewidth]{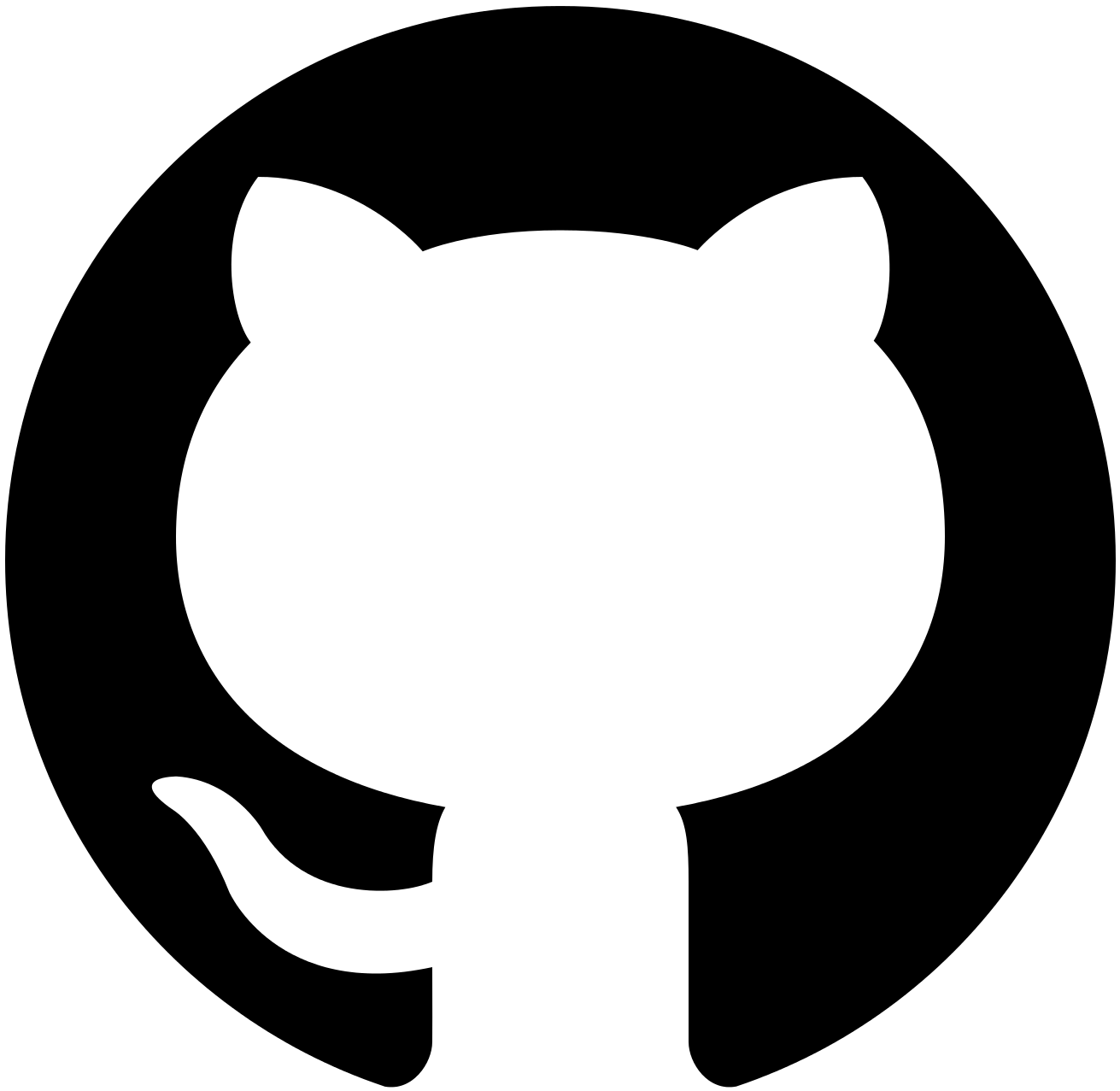}}~~GitHub Repo]{\href{https://github.com/DAGroup-PKU/ReVidgen/}{\texttt{https://github.com/DAGroup-PKU/ReVidgen/}}
\\[-1.7ex]}

\checkdata[
\raisebox{-0.3em}{\includegraphics[width=0.026\linewidth]{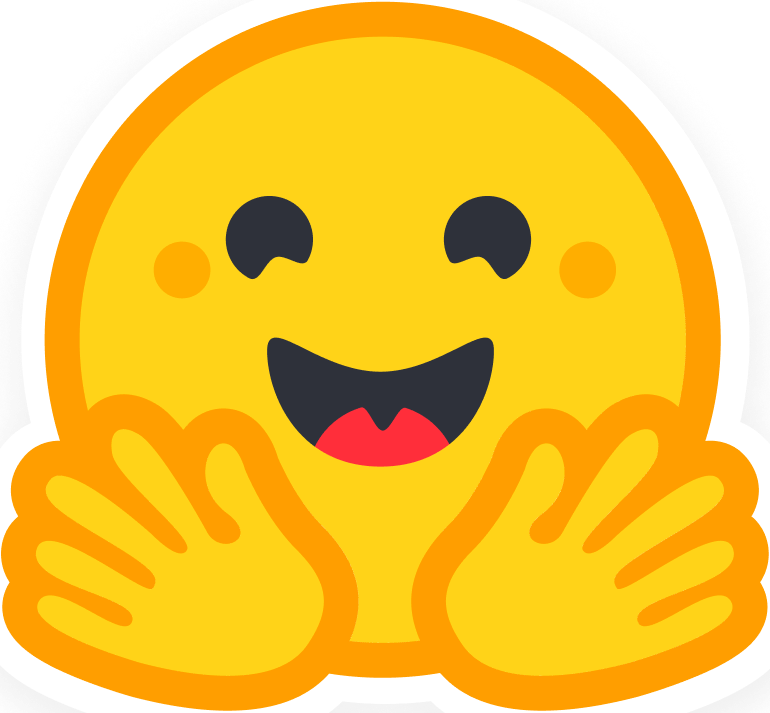}}~~HuggingFace Dataset]{\href{https://huggingface.co/datasets/DAGroup-PKU/RoVid-X/}{\texttt{https://huggingface.co/datasets/DAGroup-PKU/RoVid-X/}}}

\begin{document}
\maketitle

\begin{figure}[ht]
\centering
\includegraphics[width=\linewidth]{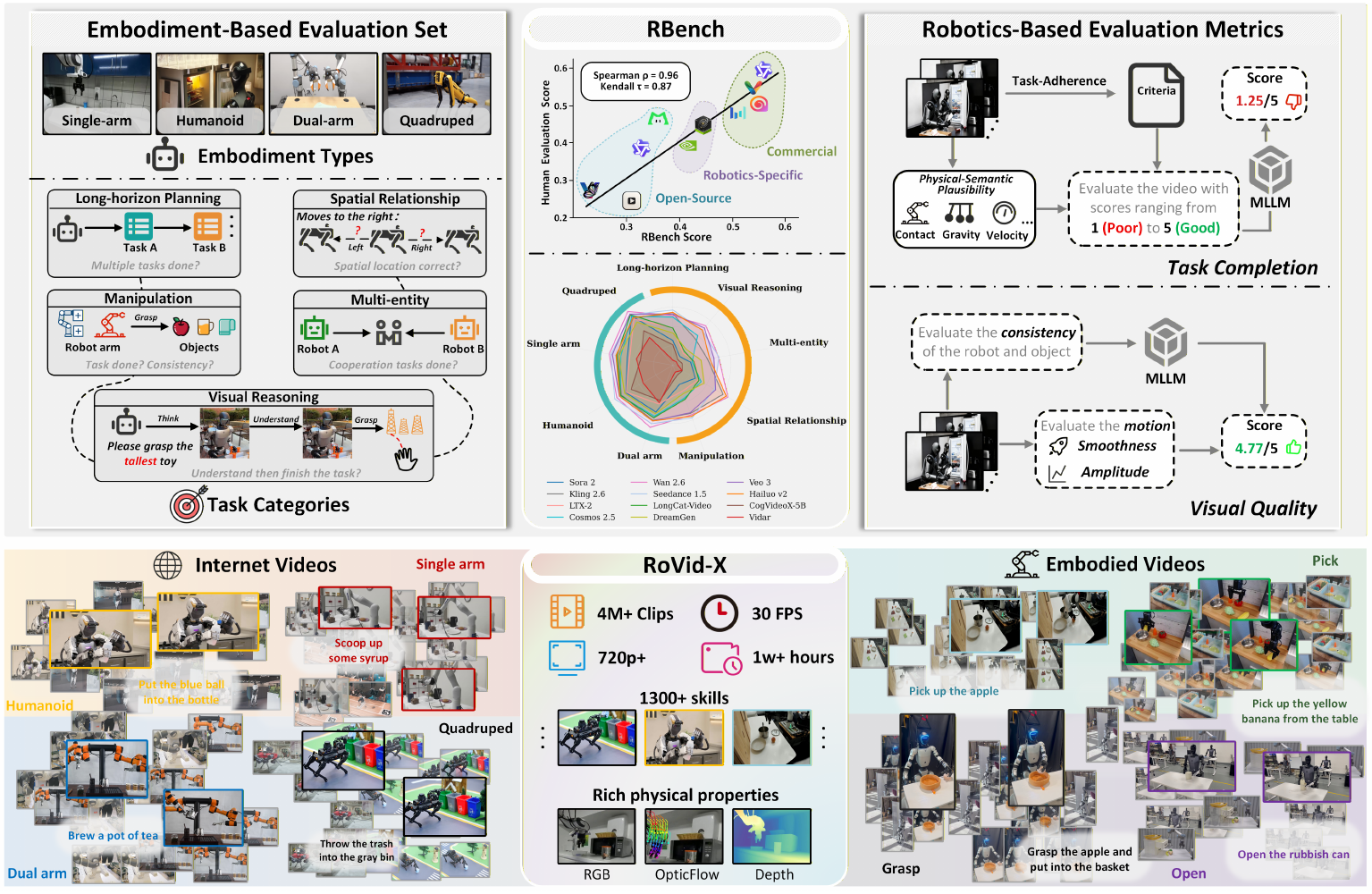}
\caption{\textbf{Overview of the comprehensive robotics benchmark and dataset for video generation.} \textbf{Top: }We present \textbf{RBench} that includes the embodiment-based evaluation set and automated evaluation metrics. Our evaluation results on 25 video models show a high level of agreement with subjective human assessments. \textbf{Bottom: }We introduce a large-scale high-quality robotic dataset (\textbf{RoVid-X}) specifically designed for training video generation models, with data sourced from internet videos and open-source embodied videos.}
\label{figure_main_results_1}
\end{figure}

\section{Introduction}
\label{sec:intro}
Recent advancements in diffusion models~\cite{ho2020denoising,song2020score,peebles2023scalable} and video generation~\cite{runway2024runway,pika2024sceneingredients,guo2024i2v,wu2025hunyuanvideo,wan2025wan} have led to significant breakthroughs, enabling applications like video editing, multi-subject generation, and motion control~\cite{jiang2025vace,ju2025editverse,deng2025cinema,deng2025magref,wang2024motionctrl}. These models have been extended to areas such as 3D scenes~\cite{kim2025videofrom3d,ren2025gen3c}, autonomous driving~\cite{gao2025magicdrive,yan2025rlgf}, and world modeling~\cite{kang2024far,genie3}, showing strong generalization across tasks. A recent study~\cite{wiedemer2025video} suggests that, similar to LLMs in natural language processing, video models are evolving into unified foundation models for machine vision. Additionally, video models are being increasingly used in robot learning and action prediction~\cite{guo2024prediction,zhu2025unified,hu2024video,zhen2025tesseract,guo2025ctrl,liang2025video}, as well as controllable simulators for synthesizing robotic video trajectories, addressing the lack of large-scale human teleoperation data~\cite{jang2025dreamgen,bjorck2025gr00t,team2025gigabrain}. These advancements highlight the growing potential of video models in the perception-reasoning-action loop of embodied agents, paving the way for more generalizable intelligent systems in the physical world.

Despite these strides, systematic evaluation for robotic video generation remains underdeveloped. Current practices rely mostly on perceptual metrics, focusing on visual quality~\cite{huang2024vbench,liu2024evalcrafter,han2025video}, while existing physics-based benchmarks often lack task-specific datasets and criteria~\cite{meng2024towards,guo2025t2vphysbench}. 
Consequently, evaluations frequently overlook critical aspects such as task completion, action-goal alignment, and physical feasibility. This leads to overly optimistic conclusions, where high scores are assigned even to videos containing unnatural movements or incomplete tasks. \textbf{The core challenge lies in rigorously assessing whether generated videos faithfully reproduce robotic behaviors.} This necessitates evaluation protocols that transcend perceptual metrics, incorporating both the physical plausibility of actions and their alignment with instructions to ensure discriminative and reproducible assessments.

To address this challenge, we propose \textbf{\textcolor{seedblue}{RBench}}, a benchmark designed to evaluate the fidelity and utility of video-based world models in robotic video generation. To the best of our knowledge, it is the first comprehensive benchmark with fine-grained metrics for robotic video generation, consisting of 650 image–text pairs across five task categories and four robot types. Evaluations are based on two dimensions: \textit{task completion} and \textit{visual quality}, incorporating sub-metrics like structural consistency, physical plausibility, and execution completeness.
Based on RBench, we conduct qualitative and quantitative assessments of 25 representative models. The results highlight that general video foundation models still have significant room for improvement in physical robot video generation, revealing a persistent gap between these models and the requirements of embodied robotic tasks. This underscores the need for systematic advancements in both robotic video data and training methodologies.

Advancing general robotic video generation with human-like capabilities and adaptability requires diverse, scalable, and comprehensive training data~\cite{o2024open,bu2025agibot}. However, unlike computer vision and natural language processing, which can leverage vast web-scale datasets, robotic interaction data has long been constrained by both scale and diversity~\cite{brohan2022rt,zhao2025humanoid,fourier2025actionnet}. Even the largest existing collections are smaller and less varied than those for vision or language. More critically, many datasets have narrow distributions along key axes such as environment, object set, task spectrum, and robot morphology~\cite{yang2025physics,wang2023robogen}, often confined to specific robot types, low-resolution recordings, or limited task ranges, which hampers the generalization of video foundation models. To address these gaps, we integrate over 20 open-source datasets and multi-source video platforms, creating a four-stage end-to-end data pipeline. The stages include robot video collection, video quality filtering, task segmentation and captioning, and physical property annotation, resulting in \textbf{\textcolor{seedblue}{RoVid-X}}, a large-scale, high-quality embodied robotic video dataset (see Table~\ref{tab:dataset_comparison}). To our knowledge, RoVid-X is currently the largest dataset specifically designed for embodied video generation models, covering a broad range of robot morphologies and task types. It aims to enhance video foundation models with physical interaction priors and task semantic diversity, driving further advancements in the field. Overall, the main contributions are summarized as follows:
\begin{itemize}[leftmargin=*, topsep=0pt]
\item A systematic benchmark tailored for robotic video generation. We propose RBench, which comprehensively evaluates the performance of video foundation models across five robotic tasks and four robot types with 650 meticulously curated evaluation samples, while introducing reproducible automated evaluation metrics.
\item Key insights into robotic video generation for embodied research. We conduct a systematic evaluation of 25 representative video models, including open-source, commercial, and robotics-specific ones, revealing the limitations of current video foundation models and potential directions for improvement, offering new perspectives for researchers exploring the embodied domain using video world models.
\item A large-scale, high-quality robotic video dataset. We construct RoVid-X, a dataset containing approximately 4 million curated robotic videos with standardized task descriptions and physical property annotations, providing essential support for the training and evaluation of embodied video models.
\end{itemize}

\section{Related Work}
\label{sec:related work}
\subsection{Video World Modeling for Robotics}
The latest breakthroughs in video generation technology have led to the development of powerful models capable of generating high-quality videos from text or image prompts~\cite{runway2024runway,pika2024sceneingredients,openai2024sora,kong2024hunyuanvideo,wan2025wan}. With the advancement of these technologies, an increasing number of studies have begun applying them to the field of embodied intelligence~\cite{feng2025vidar,unifolm-wma-0,ali2025world,bruce2024genie}. Video provides a rich source of information for robot training~\cite{cheang2024gr}. On one hand, video generation models can be used to synthesize robot trajectories~\cite{jang2025dreamgen,bjorck2025gr00t,bharadhwaj2024gen2act}, serving as an alternative to the time-consuming and labor-intensive process of human teleoperation data. And executable actions can be extracted through inverse dynamics models (IDM)~\cite{tianpredictive,baker2022video,du2023learning,zhou2024robodreamer} or latent action models~\cite{tharwat2025latent,yelatent}. On the other hand, recent studies have explored using video to simulate task dynamics and predict future states, thereby assisting in policy learning. Specifically, video generation models are used to initialize robot policies for training~\cite{liao2025genie,jiang2025rynnvla,wuunleashing} or to simultaneously train policies and inverse dynamics models, enabling co-training with robot data~\cite{guo2024prediction,zhu2025unified,li2025unified}. These efforts demonstrate the significant potential of video generation models in the field of embodied robotics, highlighting their important value in improving training efficiency and task performance.

\subsection{Datasets for Robot Learning}
A core challenge in robot learning is the lack of large-scale, diverse datasets that facilitate the training of general-purpose robots with physical interaction capabilities~\cite{o2024open,bjorck2025gr00t}. Currently, datasets used in the embodied intelligence community for robot learning can be broadly classified into three categories: real-world robot data~\cite{brohan2022rt,o2024open,zhao2025humanoid,fourier2025actionnet,mao2024learning}, human video data~\cite{damen2018scaling,goyal2017something,grauman2022ego4d}, and synthetic robot data~\cite{yang2025physics,wang2023robogen,nasiriany2024robocasa,mandlekar2023mimicgen,jiang2025dexmimicgen,tian2025interndata}. As a key element in training physical AI models, most existing real-world robot datasets are collected through robotic teleoperation~\cite{wu2024gello,fu2024mobile,aldaco2024aloha} or by teams of human operators~\cite{bu2025agibot,lynch2023interactive,black2024pi}, which leads to high collection costs and limited data scale. Furthermore, these datasets predominantly focus on similar types of robots, resulting in issues of limited diversity and restricted environments~\cite{ebert2021bridge,khazatsky2024droid}. Additionally, inconsistent data collection and storage methods across different datasets make it difficult to enable effective cross-dataset co-training. Our focus is on collecting robot data for video generation that spans various robot morphologies and entities, and providing a unified set of physical attributes for all data sources, thereby advancing cross-entity research in robot learning.

\begin{figure*}[t]
    \centering
    \includegraphics[width=\linewidth]{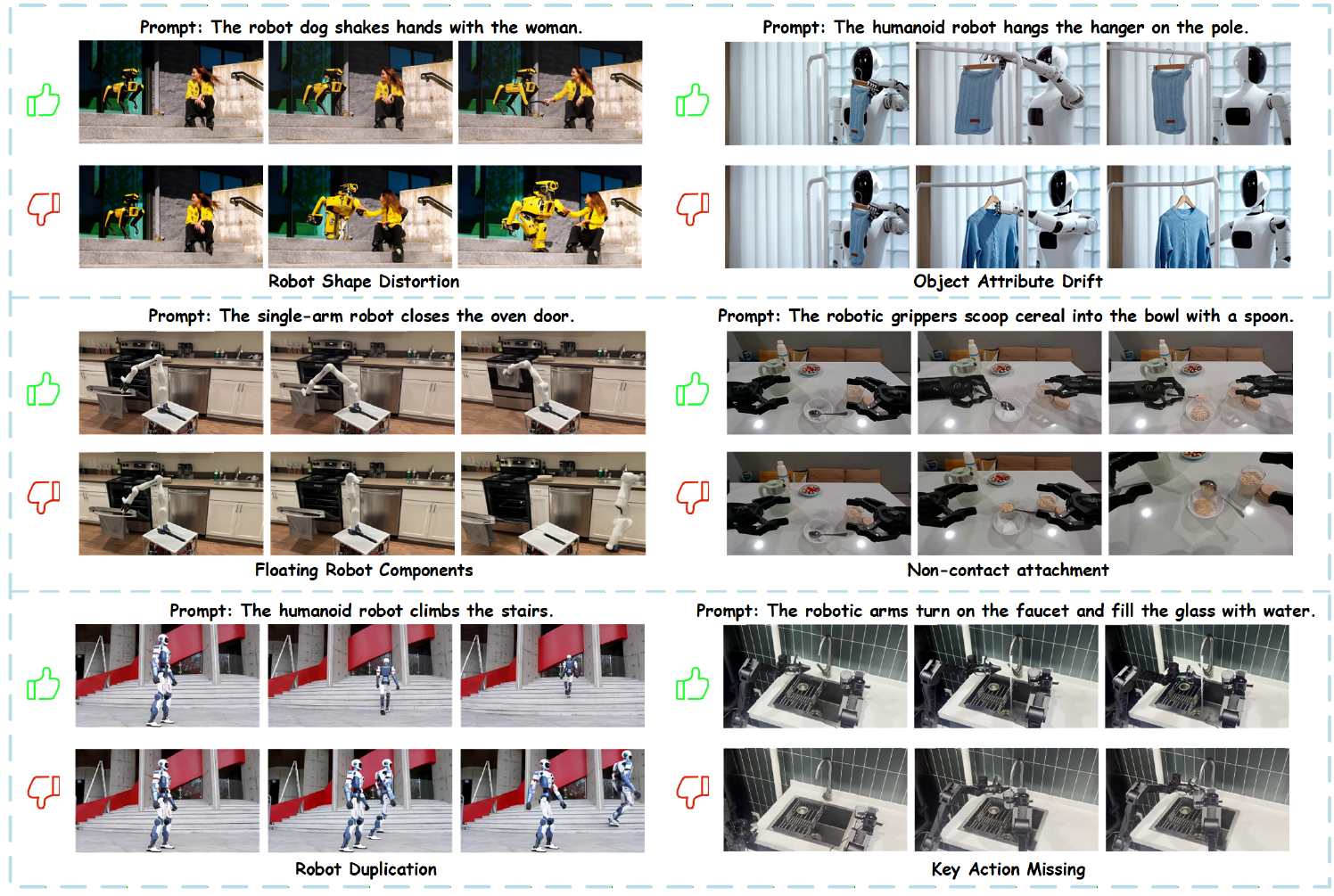}
    \vspace{-2mm}
    \caption{
    \textbf{Qualitative illustration of failure modes captured by RBench.}
    Unlike conventional metrics that focus primarily on pixel-level fidelity, RBench provides a granular evaluation across multiple dimensions, including physical plausibility and task-level consistency. These results highlight persistent challenges in robotic video generation, such as structural distortion, floating components, and key action omission, which are accurately identified by our proposed sub-metrics. More cases are shown in the Appendix~\ref{subsec:evaluation_metrics}.
    }
    \label{fig:benchmark_analysis}
    \vspace{-3pt}
\end{figure*}

\subsection{Benchmarks for Video Generation} 
Establishing robust evaluation frameworks is essential for measuring the progress of video generation models. Currently, evaluation methodologies can be categorized into three primary streams: visual fidelity and semantics, which assess basic clarity and text-video alignment~\cite{liu2023fetv,sun2025t2v,yuan2025opens2v}; temporal dynamics, focusing on motion consistency and long-range narrative coherence~\cite{Ji_2024_CVPR,liao2024evaluation,bugliarello2023storybench}; and physical plausibility, which examines adherence to fundamental laws such as inertia and collision dynamics~\cite{meng2024towards,bansal2024videophy,meng2024phybench,wang2025wisa}. While these benchmarks provide valuable insights into general video quality, they are largely decoupled from the specific requirements of embodied AI. Specifically, existing frameworks often rely on isolated physical constraints or local visual metrics, failing to capture the complex interplay between robotic actions and environmental responses. Furthermore, there is a distinct lack of systematic evaluation for task-level correctness and spatial constraints in multi-embodiment scenarios. To bridge this gap, we propose a comprehensive benchmark specifically tailored for robotic video generation, introducing reproducible metrics that unify physical realism with task-oriented action completeness.
\section{RBench}
\begin{figure*}[t]
    \centering
    \includegraphics[width=\linewidth]{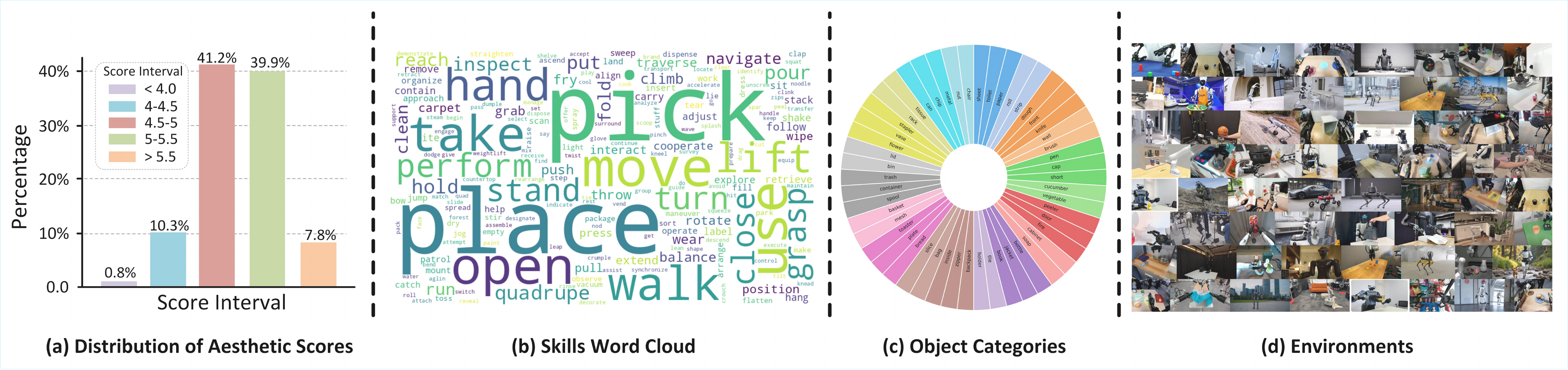}
    \vspace{-2mm}
    \caption{
    \textbf{Statistics in RBench.} The benchmark covers diverse tasks, object categories, and environments, demonstrating the high quality and comprehensiveness of the evaluation set, highlighting its high applicability to a wide range of robotic video generation scenarios.}
    \label{fig:benchmark}
    \vspace{-3mm}
\end{figure*}

Existing video generation benchmarks primarily focus on evaluating model performance in general scenes~\cite{huang2024vbench,han2025video}, while other benchmarks specifically designed for physical scenarios mainly assess models' capabilities in physical reasoning~\cite{meng2024towards,guo2025t2vphysbench}. In this paper, we design a benchmark tailored for robotic physical scenarios, aimed at comprehensively evaluating the performance of video generation models in robotic tasks. This benchmark differs from existing general scene benchmarks by focusing on evaluating video generation models' capabilities in robotic physical environments. As shown in Figure~\ref{fig:benchmark_analysis}, our benchmark highlights common failure modes in robotic video generation, including issues such as robot shape distortion, object attribute drift, non-contact attachment, and others. Section~\ref{benchmark_construction} outlines the process of benchmark construction, while Section~\ref{automatic_metrics} discusses the automatic metrics used for evaluation.

\subsection{Benchmark Construction}
\label{benchmark_construction}
To comprehensively evaluate the capabilities of video generation models in robotic scenarios, the designed evaluation dimensions must cover a wide range of task scenarios and embodiment types, ensuring that these scenarios reflect realistic robotic action semantics. To this end, we design a diversified benchmark from two aspects: task categories and embodiment types, containing a total of 650 evaluation cases. The task-oriented categories include five representative tasks: \emph{Common Manipulation}, \emph{Long-horizon Planning}, \emph{Multi-entity Collaboration}, \emph{Spatial Relationship}, and \emph{Visual Reasoning}, with a total of 250 image-text pairs, 50 samples for each task. The embodiment-specific categories cover four mainstream embodiment types: \emph{Dual-arm robots}, \emph{Humanoid robots}, \emph{Single-arm robots}, and \emph{Quadruped robots}, with a total of 400 image-text pairs, 100 samples for each embodiment type.

The benchmark includes a variety of text prompts and high-quality robot reference images. Each sample image is a keyframe extracted from high-quality videos sourced from public datasets or online sources, and each image is manually verified to ensure its accuracy. 
To avoid overlap with the training data, we ensure that the selected videos in the evaluation set do not appear in the subsequent training database, and we redesign new task prompts for each reference image, effectively avoiding the risk of content overlap. All samples are verified and filtered by human annotators to ensure that the generated task prompts align with realistic logic. Figure~\ref{fig:benchmark} illustrates the high aesthetic quality of the reference images (a), the broad range of testing scenarios including various objects, tasks, and action skills (b, c), and the diversity of environments in our evaluation set (d). Additionally, we record the metadata for each sample, such as the manipulated object, embodiment type, and camera viewpoint (first-person/third-person), to support more detailed analysis. See more details in the Appendix~\ref{sec:dataset_details}.

\subsection{Automatic Metrics}
\label{automatic_metrics}
Existing video generation evaluation protocols, such as the representative VBench~\cite{huang2024vbench}, primarily focus on perceptual quality, assessing aspects like frame clarity, texture fidelity, and motion smoothness. However, they lack task-specific criteria tailored to robotic scenarios. Recently, several studies~\cite{sun2025t2v,gu2025phyworldbench,wang2025your} have utilized multimodal large language models (MLLMs) as zero-shot evaluators for generated videos. Building upon this, we extend this approach to the domain of robotic video evaluation and propose a set of automatic evaluation metrics, incorporating manually designed indicators to assess both the visual realism and task-level validity of generated robotic videos. Following previous practices, we select the open-source Qwen3-VL~\cite{Qwen3-VL} and the closed-source GPT-5~\cite{achiam2023gpt} as our MLLM evaluation models. In the following sections, we introduce the evaluation methods for task completion and visual quality, respectively. Further details on metrics design and mathematical definitions are provided in the Appendix~\ref{subsec:evaluation_metrics}.
\subsubsection{Task Completion}
\noindent\myparagraph{Physical-Semantic Plausibility.}This metric targets everyday physical and semantic plausibility violations that standard perception scores often miss. As shown in Figure~\ref{figure_main_results_1}, we evaluate temporal grids of uniformly sampled frames with a VQA-style protocol using MLLM. Beyond assessing physical-semantic plausibility, we place special emphasis on the following frequent failure modes:
(i) \emph{Floating/Penetration}: parts of the robot or objects are not grounded or interpenetrate with solid objects; 
(ii) \emph{Spontaneous emergence}: entities appear/disappear without causal motion; 
(iii) \emph{Non-contact attachment/Incorrect grasp}: objects move with the robot without visible contact or with improper gripper closure.

\noindent\myparagraph{Task-Adherence Consistency.}This metric evaluates whether the video aligns with the intent and sequence defined by the prompt. Typical deviations include missing actions (e.g., approach without grasping or placing), incorrect order (e.g., placing before grasping), semantic drift (e.g., "wiping" becomes "touching"), and non-responsiveness. We construct temporal grids and apply an MLLM-based VQA checklist, which covers:
(i) \emph{Task responsiveness}, ensuring the goal state is reached without premature interruption;
(ii) \emph{Key actions}, verifying that required actions (e.g., grasp, place, open/close) occur and align with the prompt.

\subsubsection{Visual Quality}
\noindent\myparagraph{Motion Amplitude}This metric measures the motion amplitude of the robotic subject while discounting apparent movement caused by camera motion, thereby penalizing videos that appear smooth but lack meaningful subject activity.
Following VMBench~\cite{ling2025vmbench}, active subjects are localized with GroundingDINO~\cite{liu2023grounding}, temporally stable masks are produced by GroundedSAM~\cite{ren2024grounded}, and salient points are tracked via CoTracker~\cite{karaev24cotracker3}. Let $\bar{D}_t$ be the mean displacement of tracked points on the subject at frame $t$. The Motion Amplitude Score (MAS) is
\begin{equation}
\mathrm{MAS} \,=\, \frac{1}{T}\sum_{t=1}^{T} \min\!\bigl(\bar{D}_t,\,1\bigr),
\end{equation}
where a lower MAS indicates insufficient subject motion and complements smoothness by revealing smooth-but-inactive failure modes.

\noindent\myparagraph{Robot-Subject Stability.}This metric assesses the stability of robot morphology and target object attributes over time. Typical failures include gripper/hand shape drifting into non-mechanical forms, extra/missing manipulators, link-length/topology changes, joint inversion, object misidentification or attribute drift (class, color, position), and impossible deformation of rigid items. We adopt a contrastive VQA setup based on MLLM, which compares a reference frame and a generated frame and assigns a consistency score targeting the above failures.

\noindent\myparagraph{Motion Smoothness}This metric quantifies temporal continuity and natural dynamics, targeting artifacts from low-level aliasing to high-level jitter/blur. Following VMBench~\cite{ling2025vmbench}, we measure frame-to-frame quality stability with the Q-Align aesthetic score~\cite{wu2023qalign}. For frames $\{f_t\}_{t=1}^{T}$ and per-frame score $Q(f_t)$, define:
\begin{equation}
\Delta Q_t \,=\, Q(f_{t-1}) - Q(f_t).
\end{equation}
A temporal anomaly is flagged when $\Delta Q_t$ exceeds an adaptive threshold $\tau_s(t)$ determined by the robotic subject’s motion. The Motion Smoothness Score (MSS) is
\begin{equation}
\mathrm{MSS} \,=\, 1 - \frac{1}{T}\sum_{t=2}^{T} \mathbb{I}\!\bigl(\Delta Q_t > \tau_s(t)\bigr),
\end{equation}
where $\mathbb{I}(\cdot)$ is the indicator function. A higher MSS indicates smoother motion.

\begin{figure*}[t]
    \centering
    \includegraphics[width=\linewidth]{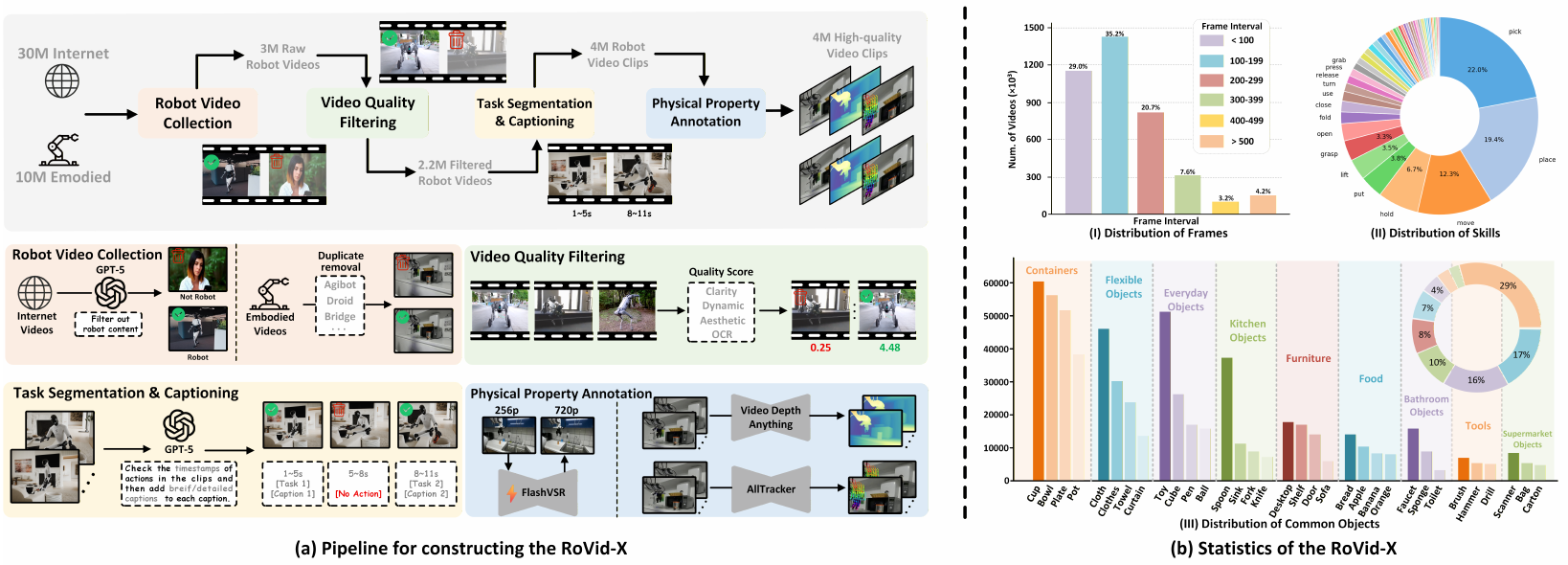}
    \vspace{-2mm}
    \caption{
    \textbf{Overview of RoVid-X Construction and Descriptive Statistics.} (a) shows the four-stage pipeline for constructing the RoVid-X. (b) presents descriptive statistics, covering frame intervals, skill distribution, and common objects, highlighting the dataset's diversity and suitability for robotic task training and video generation.}
    \label{fig:data_cruation}
    \vspace{-3pt}
\end{figure*}
\section{RoVid-X}
In this section, we introduce the construction of a high-quality robotic video dataset, resulting in RoVid-X. The dataset is developed through a refined four-stage pipeline, as shown in Figure~\ref{fig:data_cruation} (a). The dataset primarily comes from internet-sourced robotic videos that are public domain or non-copyrighted, as well as open-source embodied video datasets, all of which are licensed for use. We then introduce the construction process of the dataset and provide statistical information.
\subsection{Dataset Construction}
Our data processing workflow consists of four distinct stages, each designed to ensure the quality, diversity, and relevance of the collected data. These stages are outlined as follows:

\noindent\myparagraph{Robot Video Collection.}In the first stage, we collect raw robotic videos from large-scale internet video platforms and over 20 open-source embodied video datasets. These datasets cover a variety of robot types and task scenarios, ensuring the breadth and diversity of the data. To improve dataset relevance and quality, we employ the GPT-5 model~\cite{achiam2023gpt} to automatically filter the content of each video and remove low-quality or irrelevant video clips that do not align with the research objectives. During the filtering process, GPT-5 identifies videos related to robotic tasks and actions based on visual content and subtitles, ensuring that all collected videos effectively support the training and evaluation of robotic tasks. After this filtering process, we identify approximately 3 million raw robotic video clips, covering different actions, tasks, and robot types.

\noindent\myparagraph{Video Quality Filtering.}In this stage, we perform a rigorous filtering procedure on the collected videos to remove low-quality and irrelevant video clips that do not align with the research objectives. First, we apply scene segmentation detection to remove all video data unrelated to robots. Then, we use a video quality scoring system to assess the videos from multiple dimensions, including clarity, dynamic effects, aesthetic performance, and optical character recognition (OCR), among other metrics. Each video clip is assigned a quality score based on these criteria, ensuring that the videos retained in the final dataset meet high-quality standards.
\begin{table*}[t]
\centering
\footnotesize
\setlength{\tabcolsep}{6pt}
\renewcommand{\arraystretch}{1.15}
\caption{\textbf{Comparison of representative robotic video datasets.}}
\resizebox{\textwidth}{!}{

\begin{tabular}{lcccccccc}
\toprule
\textbf{Dataset} & \textbf{Year} & \textbf{\#Videos} & \textbf{\#Skills}  & \textbf{Resolution} & \textbf{Optical Flow} & \textbf{Diverse Robotic Forms} & \textbf{Diverse Captions} \\
\midrule
RoboTurk~\cite{mandlekar2018roboturk}            & 2018  & 2.1k  & 2 & 480P & \xmark  & \xmark & \xmark \\
RoboNet~\cite{dasari2019robonet}                 & 2019  & 162k & N/A & 240P & \xmark  & \xmark & \xmark \\
BridgeData~\cite{ebert2021bridge}                & 2021  & 7.2k & 4 & 480P & \xmark  & \xmark & \xmark \\
RH20T~\cite{fang2023rh20t}                       & 2023  & 13k & 33 & 720P & \xmark  & \xmark & \xmark \\
DROID~\cite{khazatsky2024droid}                  & 2024  & 76k & 86 & 720P & \xmark  & \xmark & \xmark \\
Open X-Embodiment~\cite{o2024open}               & 2024  & 1.4M & 217 & 64P---720P  & \xmark  & \cmark & \xmark\\
RoboMIND~\cite{wu2024robomind}    & 2024  & 107k & 38 & 480P & \xmark  & \xmark & \xmark \\
RoboCOIN~\cite{wu2025robocoin}    & 2025  & 180k & 36 & 480P & \xmark  & \xmark & \xmark \\
Galaxea~\cite{jiang2025galaxea}    & 2025  & 100k & 58 & 720P & \xmark  & \xmark & \xmark \\
InternData-A1~\cite{tian2025interndata}    & 2025  & 630k & 18 & 480P & \xmark  & \xmark & \xmark \\
Fourier ActionNet~\cite{fourier2025actionnet}    & 2025  & 13k & 16 & 800P & \xmark  & \xmark & \xmark \\
Humanoid Everyday~\cite{zhao2025humanoid}        & 2025  & 10.3k & 221 & 320P---720P & \xmark  & \xmark & \xmark \\
Agibot World~\cite{bu2025agibot}                 & 2025  & 1M & 87 & 480P & \xmark  & \xmark & \xmark \\
\midrule
\rowcolor{RowBlue} \textbf{RoVid-X (Ours)}     & 2026  & 4M & 1300+ & 720P & \cmark  & \cmark & \cmark\\
\bottomrule
\end{tabular}
}

\label{tab:dataset_comparison}
\vspace{-5pt}
\end{table*}

\noindent\myparagraph{Task Segmentation and Captioning.}In this stage, we use a video understanding model~\cite{guo2025seed1} and a specially designed prompt template to automatically analyze the robot actions within the videos. The system segments the videos into different task segments based on timestamps, generating short subtitles for each task segment that accurately describe the robot's actions and operational details in that task.

The action recognition and description process for each task segment follows these steps: First, the system identifies all dynamic actions within the video and excludes static scenes or irrelevant actions (e.g., waiting or remaining still). The time range for each action (start and end times) is precisely labeled to ensure accuracy. Next, using the MLLM model~\cite{guo2025seed1}, textual descriptions of each action are automatically generated, including the action subject (e.g., "right arm" or "left gripper"), the object being manipulated (e.g., "nameplate" or "box"), and the specific operation details (e.g., "grasp and move" or "remove from the table"). Finally, the subtitles for each task segment are output in a standardized format, ensuring that the action descriptions for each video clip are clear, concise, and aligned with the task requirements.

\noindent\myparagraph{Physical Property Annotation.}To ensure consistency and realism of robot actions within physical space, we apply physical attribute enhancement to the videos. Specifically, we use FlashVSR~\cite{zhuang2025flashvsrrealtimediffusionbasedstreaming} to improve the video resolution, making the images clearer and enhancing the details of the actions. Then, using the AllTracker tool~\cite{harley2025alltracker}, we annotate a unified optical flow for the subjects in the videos, ensuring consistency in tracking and recording robot actions across different scenes. Additionally, using Video Depth Anything~\cite{video_depth_anything}, we generate relative depth maps to accurately describe the spatial relationships and depth information of objects in the scene. The goal of these physical attribute annotations is to provide researchers with more precise reference data, aiding in the training and evaluation of robot video generation models and offering richer physical data support for future research.

%---------------------------------------------Benchmark Table----------------------------------------------------
\begin{table*}[t]
\centering
\renewcommand{\arraystretch}{1.15}
\setlength{\tabcolsep}{3.8pt}
\caption{\textbf{RBench quantitative results.} Evaluations across task-oriented and embodiment-specific dimensions for 25 models from open-source, commercial, and robotics-specific families. The "Avg." column shows the mean score across nine indicators, with task performance in the left block and embodiment performance in the right block. In the table, a "\#" next to the Sora2 model in the top right corner indicates review limitations from the official Sora API, where approximately 50 out of 650 videos could not be generated. The scores derived from sub-metrics are reported in the Appendix~\ref{subsec: Quantitative}.}
\resizebox{\textwidth}{!}{%
\begin{tabular}{
l c c|
>{\centering\arraybackslash}p{2.1cm} % Manipulation
>{\centering\arraybackslash}p{2.1cm} % Long-horizon
>{\centering\arraybackslash}p{2.1cm} % Multi-entity
>{\centering\arraybackslash}p{2.1cm} % Spatial
>{\centering\arraybackslash}p{2.1cm}|% Reasoning
>{\centering\arraybackslash}p{2.0cm} % Single arm
>{\centering\arraybackslash}p{2.0cm} % Dual arm
>{\centering\arraybackslash}p{2.0cm} % Quad robot
>{\centering\arraybackslash}p{2.0cm}  % Humanoid
}
\toprule
\textbf{Models} & \textbf{Rank} & \textbf{Avg.} &
\multicolumn{5}{c|}{\textbf{Tasks}} &
\multicolumn{4}{c}{\textbf{Embodiments}} \\
 &  &  &
\textbf{Manipulation} & \textbf{Spatial} & \textbf{Multi-entity} & \textbf{Long-horizon} & \textbf{Reasoning} &
\textbf{Single arm} & \textbf{Dual arm} & \textbf{Quadruped} & \textbf{Humanoid} \\
\midrule
\rowcolor{RowBlue}
\multicolumn{12}{l}{\textbf{\textit{Open-source}}} \\

Wan2.2\_A14B~\cite{wan2025wan} & 8 & 0.507 &
0.381 & 0.454 & 0.373 & 0.501 & 0.330 &
0.608 & 0.582 & 0.690 & 0.648 \\

HunyuanVideo 1.5~\cite{wu2025hunyuanvideo} & 10 & 0.460 &
0.442 & 0.316 & 0.312 & 0.438 & 0.364 &
0.513 & 0.526 & 0.634 & 0.595 \\

LongCat-Video~\cite{meituanlongcatteam2025longcatvideotechnicalreport} & 11 & 0.437 &
0.372 & 0.310 & 0.220 & 0.384 & 0.186 &
0.586 & 0.576 & 0.681 & 0.621 \\

Wan2.1\_14B~\cite{wan2025wan} & 14 & 0.399 &
0.344 & 0.268 & 0.282 & 0.335 & 0.205 &
0.464 & 0.497 & 0.595 & 0.599 \\

LTX-2~\cite{hacohen2026ltx} & 15 & 0.381 &
0.284 & 0.304 & 0.233 & 0.386 & 0.164 &
0.453 & 0.424 & 0.622 & 0.555 \\

Wan2.2\_5B~\cite{wan2025wan} & 16 & 0.380 &
0.331 & 0.313 & 0.142 & 0.318 & 0.234 &
0.436 & 0.448 & 0.590 & 0.607 \\

SkyReels~\cite{chen2025skyreelsv2infinitelengthfilmgenerative} & 18 & 0.361 &
0.203 & 0.276 & 0.203 & 0.254 & 0.234 &
0.507 & 0.477 & 0.586 & 0.509 \\

LTX-Video~\cite{HaCohen2024LTXVideo} & 19 & 0.344 &
0.302 & 0.176 & 0.210 & 0.280 & 0.241 &
0.440 & 0.456 & 0.526 & 0.464 \\

FramePack~\cite{zhang2025framepackv1} & 20 & 0.339 &
0.206 & 0.258 & 0.173 & 0.169 & 0.170 &
0.440 & 0.464 & 0.626 & 0.548 \\

HunyuanVideo~\cite{kong2024hunyuanvideo} & 21 & 0.303 &
0.177 & 0.180 & 0.108 & 0.147 & 0.035 &
0.454 & 0.480 & 0.625 & 0.524 \\

CogVideoX\_5B~\cite{yang2024cogvideox} & 23 & 0.256 &
0.116 & 0.112 & 0.098 & 0.212 & 0.079 &
0.338 & 0.385 & 0.465 & 0.496 \\

\midrule
\rowcolor{RowBlue}
\multicolumn{12}{l}{\textbf{\textit{Commercial}}} \\

Wan 2.6~\cite{wan2025wan} & \cellcolor{Top1}1 & 0.607 &
0.546 & \cellcolor{Sub1}0.656 & 0.479 & 0.514 & \cellcolor{Sub1}0.531 &
0.666 & \cellcolor{Sub1}0.681 & 0.723 & 0.667 \\

Seedance 1.5 pro~\cite{chen2025seedance} & \cellcolor{Top2}2 & 0.584 &
\cellcolor{Sub1}0.577 & 0.495 & \cellcolor{Sub1}0.484 & \cellcolor{Sub1}0.570 & 0.470 &
0.648 & 0.641 & 0.680 & \cellcolor{Sub1}0.692 \\

Wan 2.5~\cite{wan2025wan} & \cellcolor{Top3}3 & 0.570 &
0.527 & 0.576 & 0.402 & 0.496 & 0.437 &
\cellcolor{Sub1}0.680 & 0.634 & \cellcolor{Sub1}0.726 & 0.654 \\

Hailuo v2~\cite{hailuo2024} & 4 & 0.565 &
0.560 & 0.637 & 0.386 & 0.545 & 0.474 &
0.594 & 0.611 & 0.640 & 0.635 \\

Veo 3~\cite{GoogleDeepMind2025Veo3} & 5 & 0.563 &
0.521 & 0.508 & 0.430 & 0.530 & 0.504 &
0.634 & 0.610 & 0.689 & 0.637 \\

Seedance 1.0~\cite{gao2025seedance} & 6 & 0.551 &
0.542 & 0.425 & 0.448 & 0.454 & 0.442 &
0.622 & 0.641 & 0.698 & 0.686 \\

Kling 2.6 pro~\cite{kling2025} & 7 & 0.534 &
0.529 & 0.598 & 0.364 & 0.530 & 0.358 &
0.570 & 0.605 & 0.637 & 0.613 \\

Sora v2 Pro$^{\#}$~\cite{openai2025sora2} & 17 & 0.362 &
0.208 & 0.268 & 0.186 & 0.255 & 0.115 &
0.476 & 0.513 & 0.664 & 0.561 \\

Sora v1~\cite{openai2024sora} & 22 & 0.266 &
0.151 & 0.223 & 0.111 & 0.166 & 0.139 &
0.314 & 0.324 & 0.544 & 0.419 \\

\midrule
\rowcolor{RowBlue}
\multicolumn{12}{l}{\textbf{\textit{Robotics-specific}}} \\

Cosmos 2.5~\cite{ali2025world} & 9 & 0.464 &
0.358 & 0.338 & 0.201 & 0.496 & 0.399 &
0.544 & 0.560 & 0.658 & 0.626 \\

DreamGen(gr1)~\cite{jang2025dreamgen} & 12 & 0.420 &
0.312 & 0.372 & 0.297 & 0.334 & 0.215 &
0.564 & 0.532 & 0.579 & 0.575 \\

DreamGen(droid)~\cite{jang2025dreamgen} & 13 & 0.405 &
0.358 & 0.348 & 0.214 & 0.316 & 0.339 &
0.499 & 0.476 & 0.542 & 0.556 \\

Vidar~\cite{feng2025vidar} & 24 & 0.206 &
0.073 & 0.106 & 0.050 & 0.054 & 0.050 &
0.382 & 0.410 & 0.374 & 0.357 \\

UnifoLM-WMA-0~\cite{unifolm-wma-0} & 25 & 0.123 &
0.036 & 0.040 & 0.018 & 0.062 & 0.000 &
0.268 & 0.194 & 0.293 & 0.200 \\

\bottomrule
\end{tabular}

}
\label{tab:magicbench_combined}
\end{table*}

\subsection{Dataset Analysis}
RoVid-X is the first open-source large-scale robotic video dataset specifically designed for training video generation models, containing 4 million robotic video clips. This dataset is designed to address the physical challenges that video generation models face when generating robotic videos, providing high-quality data for both training and evaluation. RoVid-X aims to bridge the gap between traditional video generation tasks and the unique demands of embodied robot learning, where physical interaction, spatial relationships, and real-world dynamics play a crucial role.

The dataset includes a diverse range of robotic actions, tasks, and robot types, ensuring its applicability across different robotic domains. By incorporating videos from various robot types and scenarios, RoVid-X provides comprehensive coverage of the physical properties and task requirements needed for robot training. As shown in Figure~\ref{fig:data_cruation} (b), detailed statistics of the dataset are provided, illustrating the variety in terms of action skills, task types, and interaction objects. The wide-ranging data distribution of RoVid-X is critical for supporting the development of robust video generation models that can simulate realistic robot behaviors in dynamic environments.

\section{Experiment}
\begin{figure*}[t]
    \centering
    \includegraphics[width=\linewidth]{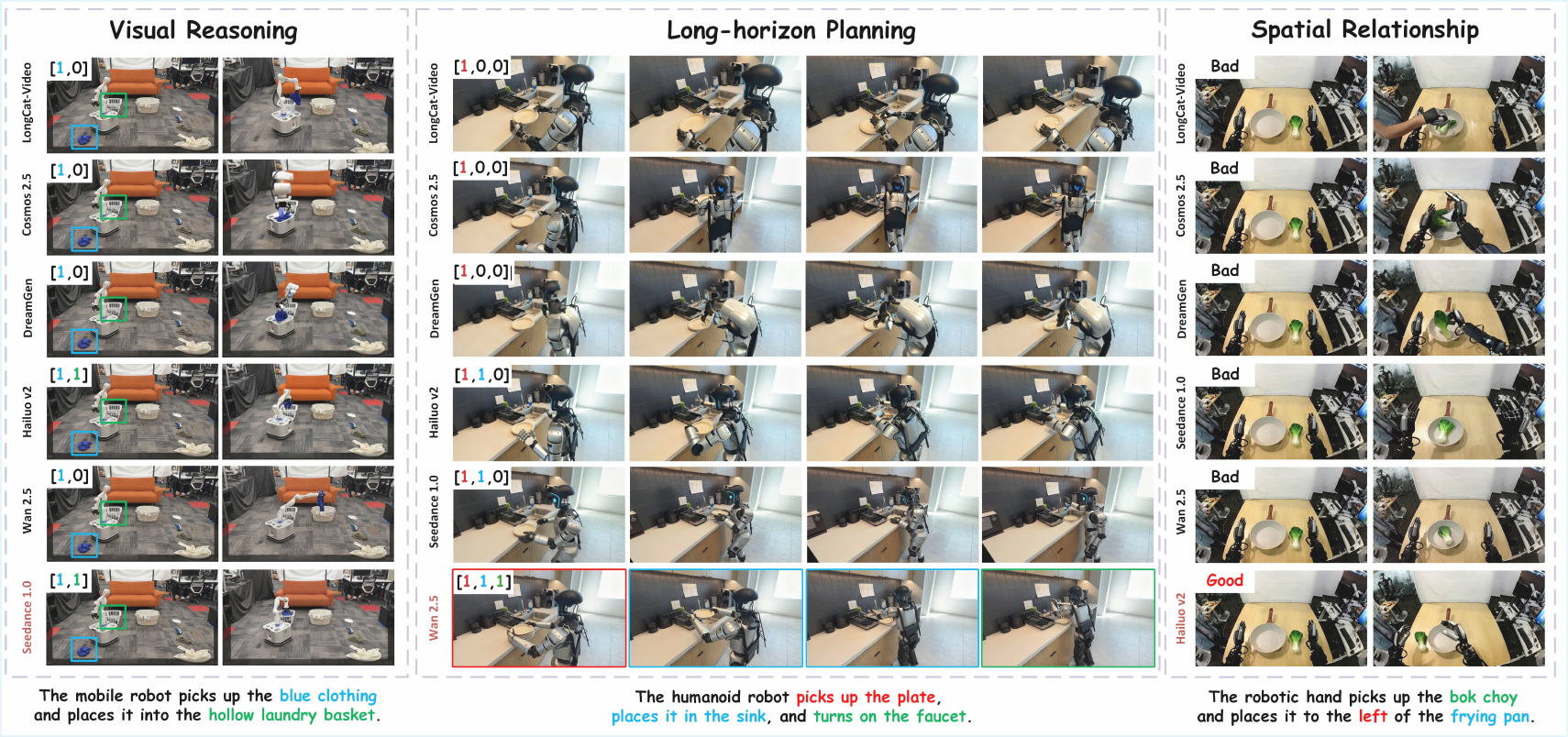}
    \vspace{-2mm}
    \caption{
    \textbf{Qualitative comparison across representative tasks.}
    We visualize the generated results for three representative tasks: \textbf{Visual Reasoning}, \textbf{Long-horizon Planning}, and \textbf{Spatial Relationship}, across six models. Each row displays temporally sampled frames from the same generated video, with captions below indicating the corresponding task instruction. More cases are shown in the Appendix.
    }
    \label{fig:qualitative_analysis}
    \vspace{-3pt}
\end{figure*}

\subsection{Evaluation Setups}
\noindent\myparagraph{Evaluation Models.}We evaluate 25 state-of-the-art video generation models, grouped into three types. Specifically, the closed-source models include Hailuo~\cite{hailuo2024}, Wan~\cite{wan2025wan}, Veo 3~\cite{GoogleDeepMind2025Veo3}, Sora~\cite{openai2024sora,openai2025sora2}, Kling~\cite{kling2025}, Seedance~\cite{gao2025seedance,chen2025seedance}, and others, while the open-source models include several representative models such as HunyuanVideo~\cite{kong2024hunyuanvideo,wu2025hunyuanvideo}, LTX~\cite{hacohen2026ltx,HaCohen2024LTXVideo} and CogVideoX~\cite{yang2024cogvideox}. Additionally, we assess models specifically designed for robotic tasks, such as DreamGen~\cite{jang2025dreamgen}, Vidar~\cite{feng2025vidar}, and Cosmos 2.5~\cite{ali2025world}. The evaluations of these models cover various types of embodiments and multiple tasks, providing a comprehensive perspective on model performance.

\noindent\myparagraph{Implementation Details.}To ensure a fair comparison, all open-source models generate videos using their official default configurations to ensure consistency with the model’s preset settings. For closed-source video models, we use their official APIs, strictly following the methods recommended by the developers for invoking and using the models. In the benchmark testing, we generate the videos for each image-text pair. To minimize errors, we generate three videos for each model sample and take the average as the final score for that sample.
These generated videos are evaluated using the automated evaluation metrics that we propose, which are designed to comprehensively assess multiple aspects of the generated videos, including task completion, action consistency, physical plausibility, and more. Further details on the model setup and configuration parameters are provided in the Appendix~\ref{subsec: Setups}.
\begin{table*}[t]
\centering
\small
\setlength{\tabcolsep}{6pt}

% ================= Left table =================
\begin{minipage}[t]{0.48\textwidth}
\caption{\textbf{Comparison between human preference scores and RBench scores.} This table demonstrates a high correlation between the two sets of scores, as reflected in the similar ranking orders.} 
\label{tab:human-rbench}
\centering
\vspace{0pt}
\begin{tabular}{lccccc}
  \toprule
  \textbf{Model} & \textbf{Human} & \textbf{RBench} & $r_h$ & $r_b$ & $\Delta r$ \\
  \midrule
  Wan 2.5           & 0.573 & 0.570 & 1  & 1  & 0  \\
  Veo 3             & 0.540 & 0.563 & 2  & 3  & 1  \\
  Hailuo v2         & 0.513 & 0.565 & 3  & 2  & -1 \\
  Seedance 1.0      & 0.505 & 0.551 & 4  & 4  & 0  \\
  Cosmos 2.5        & 0.500 & 0.464 & 5  & 5  & 0  \\
  DreamGen          & 0.482 & 0.420 & 6  & 7  & 1 \\
  LongCat-Video     & 0.480 & 0.437 & 7  & 6  & -1  \\
  Wan2.1-14B        & 0.378 & 0.399 & 8  & 8  & 0  \\
  CogVideoX-5B      & 0.333 & 0.256 & 9  & 10 & 1  \\
  LTX-Video         & 0.246 & 0.344 & 10 & 9  & -1 \\
  \bottomrule
\end{tabular}
\end{minipage}
\hfill
% ================= Right tables (stacked vertically) =================
\begin{minipage}[t]{0.48\textwidth}
\caption{\textbf{RoVid-X effectiveness validation experiment.} The experimental results using different models for finetuning show stable improvements across various dimensions, validating the effectiveness of the dataset.} 
\label{tab:validation}
\centering
\setlength{\tabcolsep}{4pt}
\renewcommand{\arraystretch}{1.21}
\vspace{0pt}
% ---- Top right table ----
\resizebox{\linewidth}{!}{%
\begin{tabular}{lccccc}
  \toprule
  \textbf{Model} & \textbf{Manip.} & \textbf{Long.} & \textbf{Multi.} & \textbf{Spatial.} & \textbf{Reason.} \\
  \midrule
  Wan2.1\_14B            & 0.344 & 0.335 & 0.282 & 0.268 & 0.205 \\
  Wan2.1\_14B+Ours     & \textbf{0.376} & \textbf{0.389} & \textbf{0.295} & \textbf{0.314} & \textbf{0.298} \\
  Wan2.2\_5B             & 0.331 & 0.318 & 0.142 & 0.313 & 0.234 \\
  Wan2.2\_5B+Ours      & \textbf{0.373} & \textbf{0.387} & \textbf{0.221} & \textbf{0.403} & \textbf{0.284} \\
  \bottomrule
\end{tabular}}

\vspace{6pt}

% ---- Bottom right table ----
\resizebox{\linewidth}{!}{%
\begin{tabular}{lccccc}
  \toprule
  \textbf{Model} & \textbf{Single} & \textbf{Dual} & \textbf{Quad.} & \textbf{Humanoid} & \textbf{Total} \\
  \midrule
  Wan2.1\_14B            & 0.464 & 0.497 & 0.595 & 0.599 & 0.399 \\
  Wan2.1\_14B + Ours     & \textbf{0.526} & \textbf{0.546} & \textbf{0.639} & \textbf{0.628} & \textbf{0.446} \\
  Wan2.2\_5B             & 0.436 & 0.448 & 0.590 & 0.607 & 0.380 \\
  Wan2.2\_5B + Ours      & \textbf{0.514} & \textbf{0.503} & \textbf{0.628} & \textbf{0.641} & \textbf{0.439} \\
  \specialrule{0.9pt}{0pt}{0pt}
\end{tabular}}

\end{minipage}

\vspace{-4pt}
\end{table*}

\subsection{Main Analysis}
\subsubsection{Quantitative Results}
Table~\ref{tab:magicbench_combined} presents a comprehensive quantitative evaluation across varying model architectures, tasks, and embodiments. Beyond standard performance metrics, the results reveal a pivotal paradigm shift in the video generation landscape.

\vspace{2pt}
\noindent\textbf{From Visual Fidelity to Physical Intelligence.}
The most significant trend observed is the transition of video generation models from pursuing high-fidelity visualization to addressing the complex dynamics of the physical world. While traditional metrics prioritize pixel-level quality, our benchmark highlights that top-tier commercial models (e.g., Wan 2.6, Seedance 1.5 Pro) are beginning to emerge as effective \textit{World Simulators}." This indicates that the field is moving towards a new stage: \textit{Physical AI}, where models must understand and simulate interaction-rich, physically challenging real-world scenarios rather than merely generating aesthetically pleasing videos.

\noindent\textbf{Iterative Scaling Unlocks Physical Capabilities.}
Analyzing model evolution reveals a strong correlation between model iteration and physical reasoning capabilities. For instance, the \textit{Wan} series exhibits a dramatic performance leap: from Wan 2.1 (Rank 14, 0.399) to Wan 2.6 (Rank 1, 0.607). Similarly, Seedance evolves from 1.0 to 1.5 Pro, climbing from Rank 6 to Rank 2. These substantial gains suggest that scaling laws and iterative optimization are not just improving visual quality but are actively refining the model's understanding of physics, distinct motion patterns, and control logic.

\noindent\textbf{The "Media-Simulation" Gap in Consumer Models.}
Surprisingly, widely recognized consumer-oriented models like the Sora series perform sub-optimally on this benchmark (Sora v2 Pro at Rank 17, Avg 0.362). This counter-intuitive result highlights a critical "domain gap": models optimized for media consumption prioritize visual smoothness and cinematic transitions, often at the expense of physical fidelity and precise motion control. This discrepancy suggests that proficiency in creative video generation does not naturally transfer to Embodied AI tasks, underlining the necessity for physically-grounded training data.

\noindent\textbf{Closed-source Models Lead in Performance.}
Commercial closed-source models occupy the top 7 positions in our benchmark, demonstrating a clear and consistent advantage over open-source counterparts. The significant performance margin between the state-of-the-art commercial model (Wan 2.6) and the leading open-source model (Wan 2.2) highlights a substantial capability gap. This disparity underscores a critical urgency for the open-source community: to democratize high-capability foundation models, more concerted efforts are needed in scaling physical training data and optimizing architectures for embodied video tasks.

\noindent\textbf{The Dilemma of Specialization: Domain Data vs. World Knowledge.}
While General Foundation Models lead the leaderboard, the robotics-specific model Cosmos 2.5 demonstrates remarkable resilience. Despite trailing top-tier commercial models, it outperforms significantly larger open-source video models, confirming that training with physical data yields stable gains in robotic tasks. Conversely, models fine-tuned on specific robot entities (e.g., Vidar, UnifoLM) struggle significantly, ranking at the bottom of the benchmark. This contrast highlights a critical trade-off: while domain-specific data is valuable for control precision, it cannot fully compensate for the deficit in "World Knowledge" provided by large-scale pretraining. Balancing proprietary robot data with generalizable representations remains a pivotal challenge for future research.

\noindent\textbf{Cognitive and Fine-grained Control Bottlenecks.}
A consistent trend across all model families is that tasks requiring high-level logic or precise interaction represent the most significant performance bottlenecks. 
First, regarding cognitive capabilities, we observe a substantial "Cognitive Gap": while top-tier models like Wan 2.6 excel in execution-oriented tasks, their performance drops sharply in \textit{Visual Reasoning} (0.531). Furthermore, analyzing specific embodiments reveals a "Manipulation Gap": models consistently score higher on coarse-grained locomotion tasks (Quadruped, Humanoid) than on fine-grained manipulation. This implies that for current video generators, mastering the fine-grained contact dynamics required for object interaction is physically more challenging than generating the rhythmic patterns of legged locomotion.

\subsubsection{Qualitative Results}
We conduct a qualitative analysis of representative tasks, and the partial results are shown in Figure~\ref{fig:qualitative_analysis}. For the visual reasoning task, Seedance 1.0~\cite{gao2025seedance} and Hailuo~\cite{hailuo2024} correctly identify the blue clothing and the hollow basket, while Wan 2.5~\cite{wan2025wan} mistakenly identifies the woven basket as the hollow basket. In the long-horizon planning task, Wan 2.5 successfully completes all actions in the correct sequence, while Hailuo lacks the "turn-on" action, leading to a violation of physical logic. In the spatial relationship task, Hailuo correctly places the bok choy to the left of the pan, whereas other models mistakenly place it inside the pan. Notably, LongCat-Video introduces an unrealistic human arm intervention, disrupting physical plausibility. More detailed analysis and qualitative results can be found in the Appendix~\ref{subsec: Qualitative}. 

These models each have their strengths, but there is still significant room for improvement in their overall performance. This further highlights the necessity of designing such a benchmark to advance video generation models in robotic tasks.

\subsection{Human Preference Study}
We conduct a human preference study to assess how well automatic metrics align with human perception. Thirty participants are invited to participate. For each comparison, two model outputs for the same prompt and video instance are presented side-by-side, and annotators choose from three options: ``A is better,'' ``B is better,'' or ``Tie''. Votes are aggregated into per-model scores: a win contributes \(5\), a loss contributes \(1\), and a tie contributes \(3\) to both models.
We then compare these model-level human scores with the corresponding RBench benchmark scores. On the ten-model subset used in the study, the Spearman rank correlation between human scores and RBench scores is \( \rho = 0.96 \) (two-sided \(p < 10^{-3}\)).
Table~\ref{tab:human-rbench} presents the human scores, RBench scores, and ranks for the ten selected models, where the \(\Delta r\) column denotes the rank difference \((r_b - r_h)\).
Overall, models that rank highest under the benchmark largely match human judgments, while the remaining small discrepancies highlight opportunities to further refine the metric for improved human alignment. The high degree of consistency further demonstrates the validity and effectiveness of our automated metrics in evaluating video generation models, indicating that the metrics accurately reflect human perception and thus provide a reliable evaluation standard for robotic video generation tasks. Please refer to the Appendix~\ref{sec:human_preference} for more details.

\subsection{Validation of RoVid-X}
To assess the effectiveness and robustness of RoVid-X, we finetune models initialized with Wan2.1 14B and Wan2.2 5B weights, using MSE loss exclusively. Due to computational constraints, we randomly sample 200k instances from the original RoVid-X dataset. The results, shown in Table~\ref{tab:validation}, highlight that our dataset significantly enhances performance across five task domains and four distinct embodiments. These improvements validate both the proposed dataset and our data collection pipeline.

\section{Conclusion}
In this work, we rethink video generation models for the embodied world and introduce RBench, a new benchmark designed to fill a critical gap in evaluating robot-oriented video generation models. Unlike previous methods that primarily rely on perceptual metrics, RBench incorporates both task-level accuracy and visual fidelity, using a comprehensive evaluation suite with detailed sub-metrics such as structural consistency and physical plausibility. The evaluation of 25 models highlights that current video generation models still require significant improvements to generate physically realistic robot behaviors. The strong correlation between RBench scores and human evaluations further validates the benchmark’s effectiveness. Additionally, RoVid-X overcomes the limitations of existing robotic datasets by offering a large-scale and diverse resource for video generation tasks. Together, RBench and RoVid-X provide a robust foundation for advancing video generation models in robotics. Our findings highlight the shortcomings of current video foundation models and suggest possible avenues for improvement, providing researchers with fresh perspectives for exploring the embodied domain through video world models.

\textbf{Future Work}. We aim to bridge the gap between video generation and actionable robot policy. We plan to employ Inverse Dynamics Models (IDM) to recover executable actions from generated videos, enabling closed-loop control experiments in both simulation environments and on real-world hardware. Furthermore, we intend to develop more automated and physically grounded evaluation metrics to rigorously assess the kinematic and dynamic feasibility of generated behaviors. Additionally, we will focus on training video generation models with improved physical capabilities, enabling the generation of robot videos that perform high-fidelity actions. Ultimately, these efforts will accelerate the development of a comprehensive solution for video-driven embodied intelligence.

\clearpage

\bibliographystyle{plainnat}
\setlength{\bibhang}{0pt}
\setlength\bibindent{0pt}
\bibliography{main}

\clearpage

% \beginappendix
\clearpage
\appendix
\startcontents[chapters]
\setcounter{page}{1}

\begin{center}
  \textbf{\Large Rethinking Video Generation Model for the Embodied World} \vspace{0.5cm} \\ 
  {\Large Appendix}
  \vspace{0.5cm}
\end{center}

\printcontents[chapters]{}{1}{}

\section{Evaluation Set Details}
\label{sec:dataset_details}

\subsection{Task-Oriented Evaluation Set}
\label{subsec:task_oriented_dataset}

To systematically evaluate the multi-dimensional task execution capabilities of video generation models in robotic scenarios, RBench constructs a task-oriented evaluation set with five core task dimensions: \emph{Common Manipulation}, \emph{Long-Horizon Planning}, \emph{Multi-Entity Collaboration}, \emph{Spatial Relationship}, and \emph{Visual Reasoning}. For each task, we collect 50 images as initial frames from open-source datasets or public web sources. Human annotators then create and verify corresponding text prompts to ensure both correctness and diversity of language descriptions. Together, these image-text pairs define a diverse evaluation corpus that covers a wide range of everyday manipulation, complex planning, multi-entity interaction, spatial reasoning, and visual-semantic reasoning scenarios.

\subsubsection{Common Manipulation}
\label{subsec:dataset_common_manipulation}

This task evaluates the ability of video generation models to produce diverse manipulation behaviors in basic object interaction scenarios. The scenes cover single-arm, dual-arm, and humanoid robots performing typical manipulation actions such as grasping, placing, pushing, rotating, and pressing. The dataset focuses on whether the model can generate physically plausible, temporally coherent, and natural manipulation behaviors that achieve the specified goals for everyday object handling.

\subsubsection{Long-Horizon Planning}
\label{subsec:dataset_long_horizon_planning}

This task evaluates the capability of video generation models to understand and generate long-horizon robotic behaviors that involve multi-stage action planning. Each instance in the evaluation set is composed of multiple sequential sub-actions, including:
\begin{itemize}
\item \textbf{Object Manipulation Sequences:} e.g., The robot opens the refrigerator door, takes the green box out of the refrigerator, and then closes the refrigerator door,'' which requires clearly delineated action stages and physically reasonable transitions. \item \textbf{Multi-Step Spatial Planning:} e.g., The humanoid robot picks up the bag, turns around, climbs up the stairs, and walks across the wooden plank,'' emphasizing continuous spatial transitions and modeling of multi-step action chaining.
\item \textbf{Physical Motion and Body Coordination:} e.g., ``The quadruped robot performs a front flip, lands steadily, then leans forward and balances upside down on its front legs,'' which assesses temporal coherence and physical plausibility in complex motion and body control.
\end{itemize}

Overall, the dataset spans a broad range of tasks from everyday interactions to dynamic control, focusing on a model's capability in action decomposition, stage transitions, and cross-time reasoning for comprehensive long-horizon planning.

\subsubsection{Multi-Entity Collaboration}
\label{subsec:dataset_multi_entity_collaboration}

This task focuses on evaluating the capability of video generation models to depict collaborative and interactive behaviors in multi-entity robotic scenarios. Each scene contains a \emph{Primary Entity} and a \emph{Secondary Entity}. The Primary Entity can be a single-arm robot, a dual-arm robot, a humanoid robot, or a quadruped robot, while the Secondary Entity can be a human, an animal, or another robot. The task covers diverse interaction types such as object handover and usage, dressing assistance, collaborative task completion, following, and guidance. The dataset is designed to assess whether the model can generate natural, temporally coherent, and task-consistent multi-entity collaboration behaviors at semantic, temporal, and physical levels.

\subsubsection{Spatial Relationship}
\label{subsec:dataset_spatial_relationship}

This task evaluates the ability of video generation models to understand and express spatial relationships in generated videos. We construct scenes where humanoid robots, single-arm robots, and quadruped robots interact with clearly defined objects while satisfying various spatial relations, such as above/below, left/right, and front/behind. The dataset requires models to correctly present relative positions, orientations, and motion trajectories between entities, revealing their competence in spatial understanding and geometric reasoning. Consistent spatial layouts and motion patterns across time are essential to correctly reflect the described spatial relations.

\subsubsection{Visual Reasoning}
\label{subsec:dataset_visual_reasoning}

This task aims to evaluate the visual-semantic reasoning capabilities of video generation models in complex scenes. The evaluation set includes a wide range of visual concepts and multi-level semantic logic, such as:

\begin{itemize}
\item Color recognition (e.g., pick up the sky-blue book); \item Numerical and ordering reasoning (e.g., the robot places the apple, water bottle, and Rubik's cube into the bag in that order);
\item Attribute and category matching (e.g., the robot gripper places the white bottle of baby powder onto the shelf, aligning it with other identical bottles in the same column); \item Geometric and object-property understanding (e.g., the robot picks up the tallest orange object and places it into the basket);
\item Text and semantic understanding (e.g., the left manipulator places the cup under the white dispenser labeled `Jasmine Tea,' and the right manipulator opens the dispenser to pour jasmine tea into the cup); \item Visual feature understanding (e.g., the robot grasps the book with a portrait of a person on its cover).
\end{itemize}

This dataset is designed to emphasize fine-grained visual grounding, logical consistency, and the ability to align robot actions with high-level visual-semantic reasoning requirements.

\subsection{Embodiment-Specific Evaluation Set}
\label{subsec:embodiment_specific_dataset}

Different types of robots exhibit substantial variations in morphology, degrees of freedom, control modes, and task objectives. These factors directly influence the modeling complexity and generalization challenges faced by video generation models in robotics contexts. To more systematically analyze model performance across heterogeneous robot embodiments, RBench constructs embodiment-specific evaluation subsets that encompass four representative robot categories: dual-arm robots, humanoid robots, single-arm robots, and quadruped robots. For each embodiment, 100 initial-frame images are sourced from open-access or publicly available datasets, and human annotators create and verify the corresponding prompts for accuracy and linguistic diversity.

Each subset includes a diverse range of robot models, action types, manipulated objects, scene environments, and both first-person and third-person perspectives. This design introduces embodiment-specific challenges: dual-arm robots emphasize coordinated bimanual manipulation, humanoid robots prioritize tool use and natural full-body postures, single-arm robots focus on precise object interactions, and quadruped robots predominantly test terrain adaptation and motion continuity. Along with the task-oriented splits, these embodiment-specific subsets provide a comprehensive and structured dataset for benchmarking video generation models in the embodied world.

Evaluating models across these four robot categories further uncovers the current biases and capability preferences of image-to-video generation models. For instance, due to extensive pre-training on large-scale human activity datasets, many models tend to exhibit higher task completion rates and better visual quality in humanoid-robot scenarios, while they often struggle with fine-grained single-arm manipulation. By systematically comparing performance across different embodiments, RBench makes such imbalances explicit and offers a principled framework to identify where current models excel or fail.

More broadly, embodiment-aware datasets like RBench are pivotal for advancing video foundation models in robotics. They promote the development of architectures and training strategies capable of generalizing beyond human-centric motion priors, enabling models to learn and adapt to a broader distribution of robot-specific motion patterns, rather than relying solely on human demonstrations. Furthermore, they facilitate fair and transparent comparisons between models, allowing evaluation results to clearly identify which models perform best for specific robot embodiments. Finally, such datasets help bridge the gap between generic video generation and physically grounded embodied intelligence, fostering the transition from visually appealing but brittle outputs toward robust, controllable, and deployment-ready generative models for real-world robotic systems.
\section{Automatic Metrics Details}
\label{subsec:evaluation_metrics}

To quantitatively assess the core capabilities of different models in robot video generation, we design five fine-grained metrics: \emph{Physical-Semantic Plausibility}, \emph{Task-Adherence Consistency}, \emph{Motion Amplitude}, \emph{Robot-Subject Stability}, and \emph{Motion Smoothness}. These metrics are evaluated using an MLLM-based, VQA-style protocol applied to grid images composed of key frames sampled from each generated video. Additionally, the evaluation is supplemented by low-level, non-MLLM computational indicators that capture pixel-level motion statistics and temporal dynamics. Together, these two layers of evaluation provide a comprehensive assessment of both task completion and visual quality in robotic video generation.

\subsection{Physical-Semantic Plausibility}
\label{subsubsec:physical_semantic}
\begin{figure*}[htbp]
    \centering
    \includegraphics[width=1.0\textwidth]{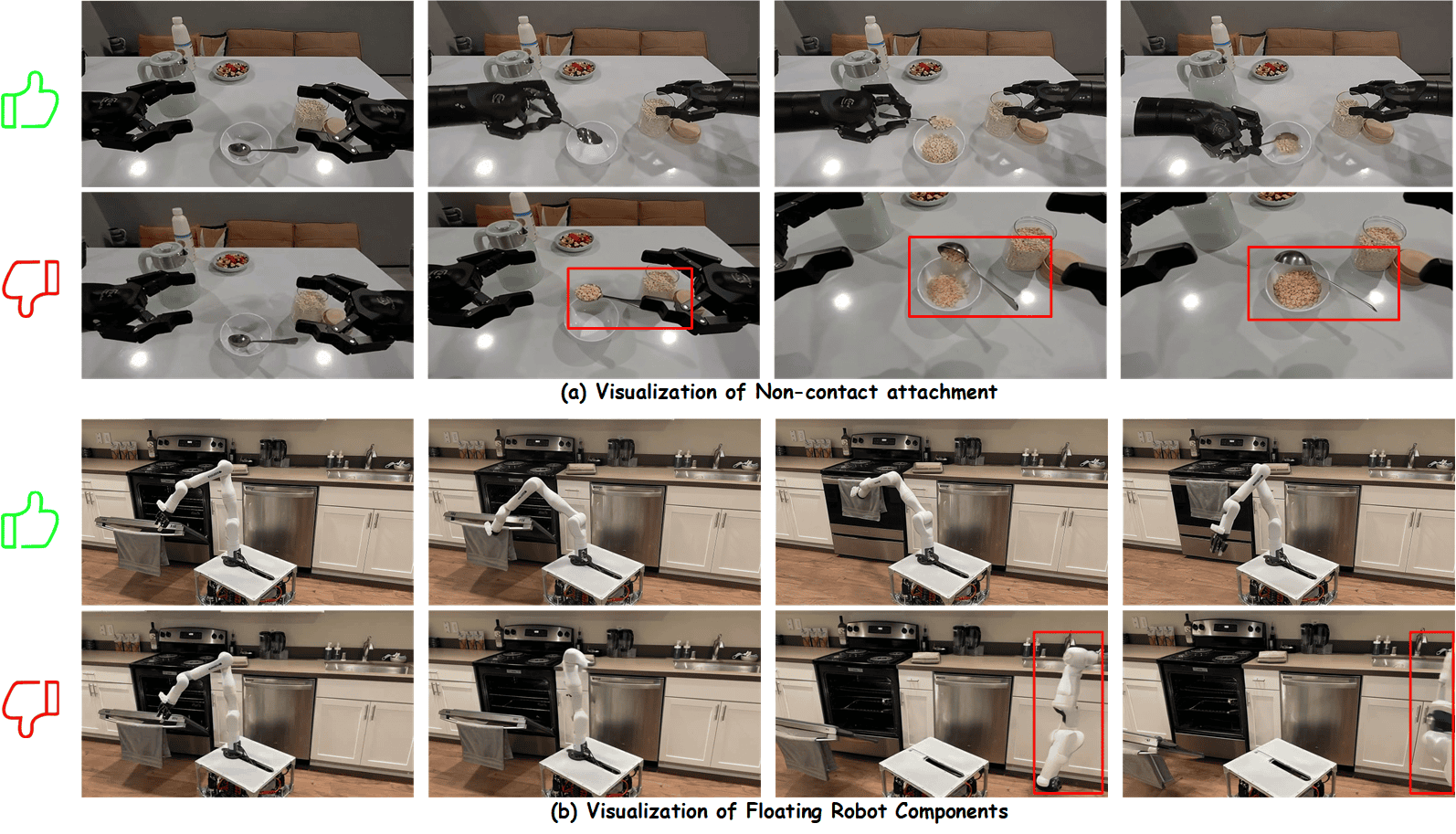}
    \caption{\textbf{Visualization of robot and subject floating.}}
    \label{fig:physical_01}
\end{figure*}

In robotics video generation, models often produce physically implausible or commonsense-violating artifacts, such as grippers passing through objects, floating objects, or the sudden appearance of irrelevant entities. These errors are typically undetectable by standard visual perception metrics, yet they directly highlight limitations in a model's understanding of physical laws and semantic causality.

\begin{figure*}[htbp]
    \centering
    \includegraphics[width=1.0\textwidth]{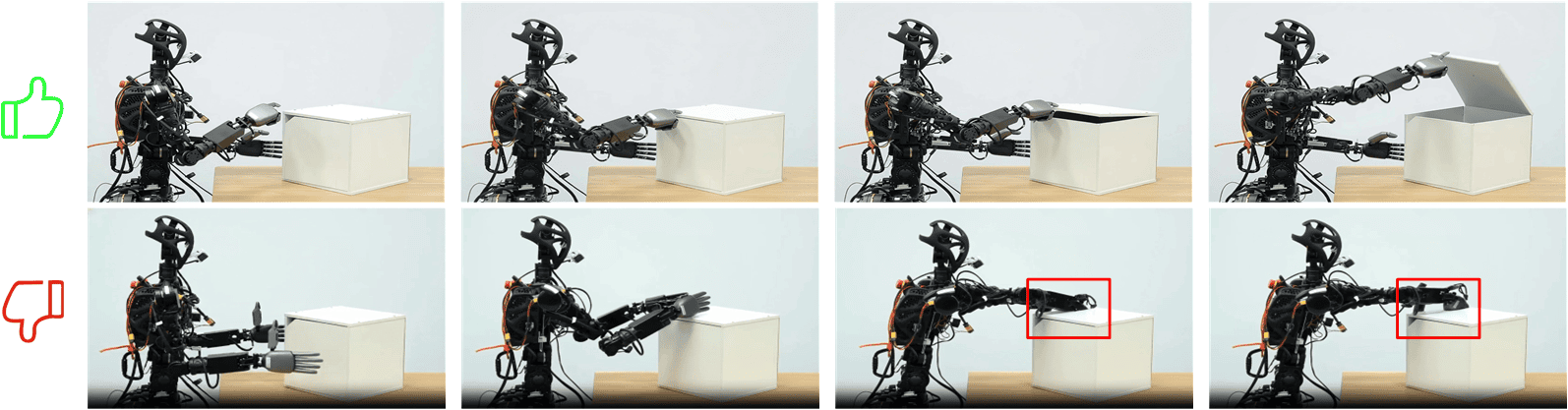}
    \caption{\textbf{Visualization of robot interpenetration.}}
    \label{fig:physical_02}
\end{figure*}

\begin{figure*}[htbp]
    \centering
    \includegraphics[width=1.0\textwidth]{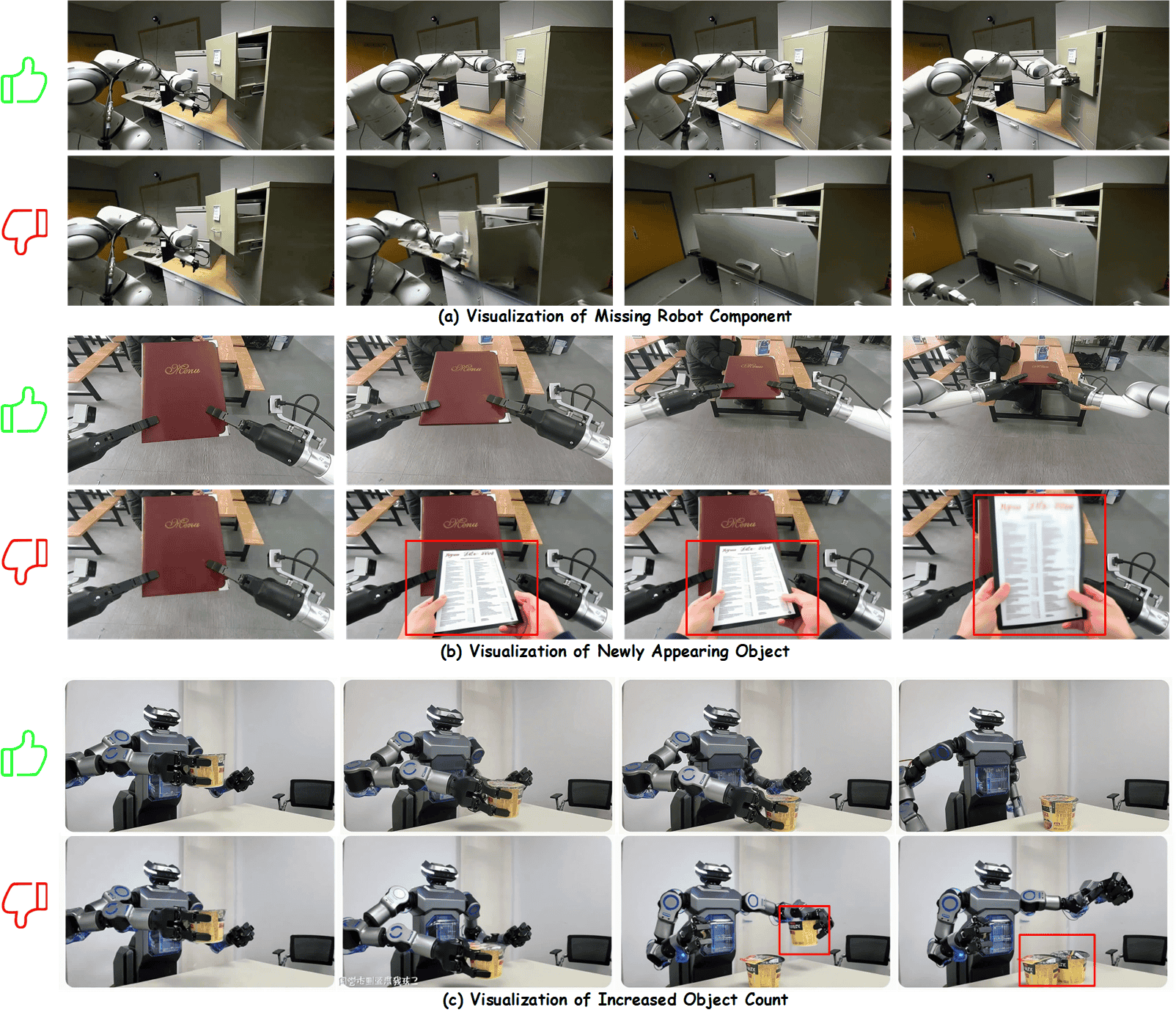}
    \caption{\textbf{Visualization of robot/subject sudden appearance, disappearance, or duplication.}}
    \label{fig:physical_03}
\end{figure*}

To capture these issues, we introduce the \emph{Physical-Semantic Plausibility} metric, implemented via a VQA-style evaluation pipeline. The MLLM receives a grid image composed of key frames from the generated video and is prompted to detect the following types of violations:
\begin{itemize}
    \item \textbf{Floating and unsupported entities.} As illustrated in Figure~\ref{fig:physical_01}, the metallic spoon and the single-arm robot's manipulator are suspended in mid-air without any physically plausible support.
    \item \textbf{Interpenetration.} As shown in Figure~\ref{fig:physical_02}, the humanoid robot hand unrealistically penetrates the box, indicating a severe violation of rigid-body constraints.
    \item \textbf{Sudden appearance, disappearance, or duplication.} As demonstrated in Figure~\ref{fig:physical_03}, (a) the robotic arm suddenly disappears in later frames, (b) human hands and a new notebook suddenly appear, and (c) the number of instant noodle packs is spuriously duplicated.
    \item \textbf{Non-contact attachment and incorrect grasping.} For example, in Figure~\ref{fig:physical_01}(a), the metallic spoon moves rigidly with the gripper even though there is no clear contact or gripper closure, resulting in an unrealistic ``sticking'' effect.
\end{itemize}

These anomalies are treated as severe physical violations that significantly reduce the credibility of the generated video. Beyond local error detection, the evaluator is also required to assess whether the overall action sequence and causal progression are reasonable, thereby characterizing the extent to which the model produces videos that are consistent with basic physical laws and human common sense.

\subsection{Task-Adherence Consistency}
\label{subsubsec:task_adherence}
\begin{figure*}[htbp]
    \centering
    \includegraphics[width=1.0\textwidth]{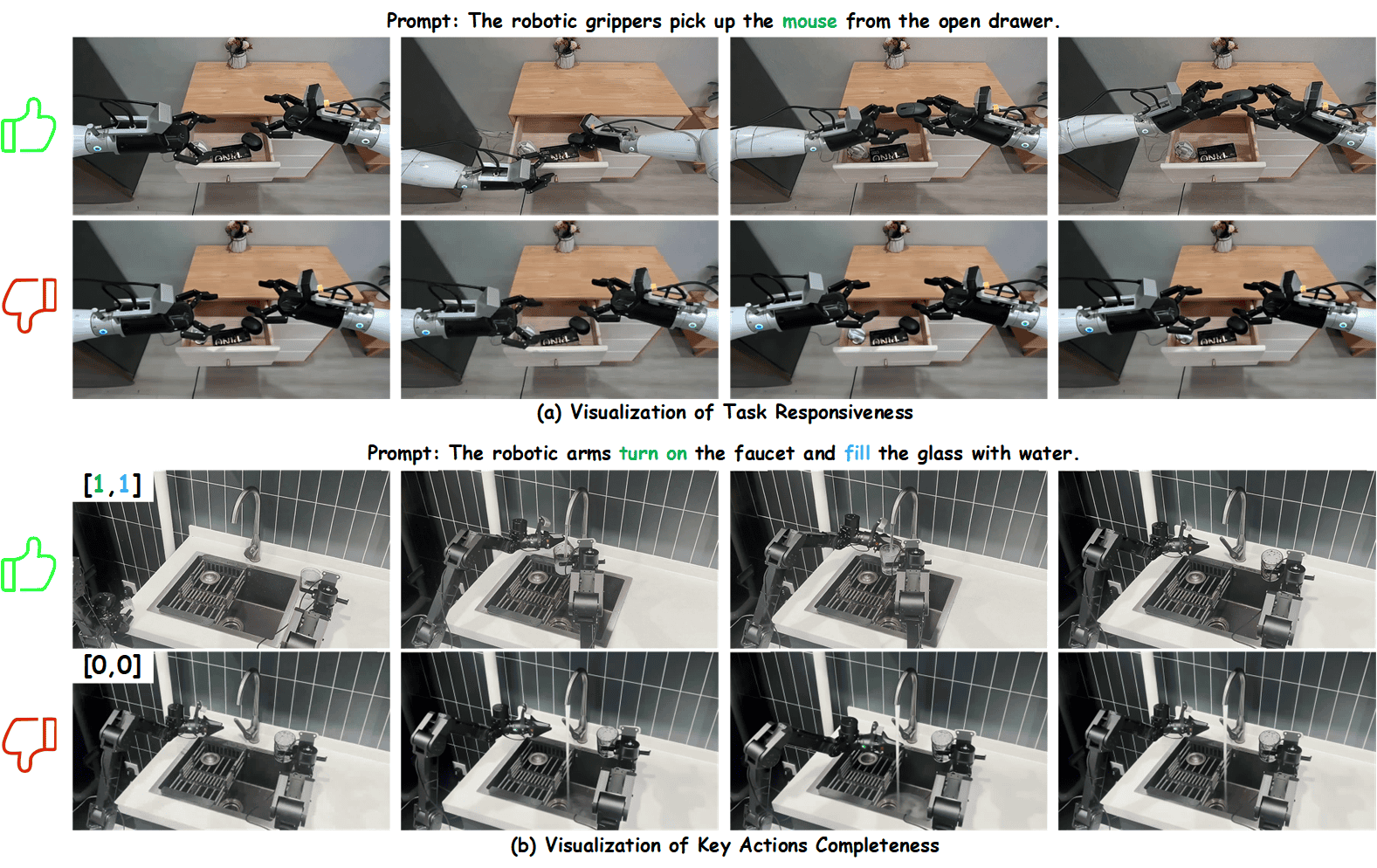}
    \caption{\textbf{Visualization of task responsiveness and key actions completeness.}}
    \label{fig:task}
\end{figure*}

Robotics video generation models often exhibit task-level deviations, such as ignoring the specified objective or omitting critical action stages. To measure this behavior, we design the \emph{Task-Adherence Consistency} metric, using a VQA-style evaluation protocol. The MLLM inspects the grid of key frames and assesses the following:

\begin{itemize}
    \item \textbf{Task responsiveness.} As shown in Figure~\ref{fig:task}(a), the failure case illustrates a robot gripper that does not respond to the instruction to grasp the mouse; the gripper remains static, and the intended task is never initiated or completed.
    \item \textbf{Key action completeness.} As illustrated in Figure~\ref{fig:task}(b), the failure case omits crucial actions such as \emph{turn on} and \emph{fill}: the faucet is never visibly operated, yet water still flows from the tap, disrupting the causal chain between actions and outcomes.
\end{itemize}

These phenomena reflect shortcomings in semantic understanding, action planning, and execution consistency with respect to the prompt. Importantly, they are also difficult to capture with conventional low-level perception metrics, highlighting the necessity of explicit task-adherence evaluation in robotics contexts.

Concretely, Task-Adherence Consistency is instantiated with task-specific criteria for the five task families introduced in Section~\ref{subsec:task_oriented_dataset}, with a focus on the following:

\begin{itemize}
    \item \textbf{Common Manipulation.} Task adherence is primarily assessed through: (i) \emph{Task Completion}, which checks whether the robot successfully accomplishes the manipulation objective described in the prompt while exhibiting reasonable intermediate phases (e.g., approach $\rightarrow$ grasp $\rightarrow$ move $\rightarrow$ place); and (ii) \emph{Action Effectiveness}, which evaluates the physical plausibility and dynamic coherence of the manipulation, including natural gripper closure, appropriate contact locations, and smooth trajectories. Attempts with obviously discontinuous, incomplete, or physically implausible actions are regarded as failures.
    \item \textbf{Multi-Entity Collaboration.} For collaborative scenes involving a Primary and a Secondary Entity, task adherence is assessed through two aspects: (i) \emph{Task Completion}, requiring that both entities execute their respective roles and complete all required interaction steps in a temporally coherent and logically consistent manner; and (ii) \emph{Action Effectiveness}, which measures the completeness and coordination of interaction behaviors. For contact-based interactions, a full sequence of ``approach $\rightarrow$ contact $\rightarrow$ release/transfer'' is expected; for non-contact interactions (e.g., following, joint motion), a coherent process of ``initiation $\rightarrow$ alignment $\rightarrow$ sustained coordination'' is required. Missing stages, asynchronous responses, or logically inconsistent behaviors are treated as unsuccessful.
    \item \textbf{Spatial Relationship.} In spatial reasoning scenarios, task adherence is assessed through: (i) \emph{Spatial Relation Accuracy}, which checks whether the spatial relations between entities (e.g., above/below, left/right, front/behind) match the textual description with consistent orientation, scale, and viewpoint; and (ii) \emph{Manipulation Feasibility}, which examines whether the direction, trajectory, and intent of the robot's motion are compatible with the described spatial relations (e.g., moving leftward when instructed to move ``to the left of''). Trajectories that contradict the described direction or result in physically unreasonable motions are considered incorrect.
    \item \textbf{Visual Reasoning.} In visually and semantically complex scenes, task adherence is assessed through two aspects: (i) \emph{Visual Reasoning Accuracy}, evaluated via an automatic Question Chain mechanism: given the prompt, MLLM first generates a set of stepwise verification questions covering the trigger-feedback-outcome logic. The same MLLM then answers these questions based on the generated video, and a score is computed as follows:

    \begin{equation}
        \text{Score} = 5 \times \frac{\text{completed questions}}{\text{total questions}},
    \end{equation}

    where missing or incorrect events are treated as unfulfilled steps.
 This encourages the video to satisfy both the visual and logical requirements of the task. Additionally, (ii) \emph{Action Effectiveness} measures the physical plausibility and dynamic coherence of the robot's motions, penalizing clearly discontinuous, incomplete, or physically implausible actions even if some high-level reasoning appears correct.
    \item \textbf{Long-Horizon Planning.} For long-horizon tasks composed of multiple ordered sub-events, task adherence is assessed through: (i) \emph{Event Completion Rate}. For each sample, an event list is defined as an ordered set of events. This list is reformulated into a numbered sequence, e.g., \textit{1. open the refrigerator door; 2. take out the green box; 3. close the door}, and the final score is computed as follows:

    \begin{equation}
        \text{Score} = 5 \times \frac{\text{completed events}}{\text{total events}},
    \end{equation}

    this measures how completely the required event sequence is executed. Complementarily, (ii) \emph{Action Effectiveness} again assesses the physical plausibility and temporal coherence of the underlying motions (e.g., natural body coordination, stable landing, and smooth transitions between stages), ensuring that partially correct high-level event ordering without valid execution is not over-rewarded.
\end{itemize}

\subsection{Robot-Subject Stability}
\label{subsubsec:robot_subject_stability}
\begin{figure*}[htbp]
    \centering
    \includegraphics[width=1.0\textwidth]{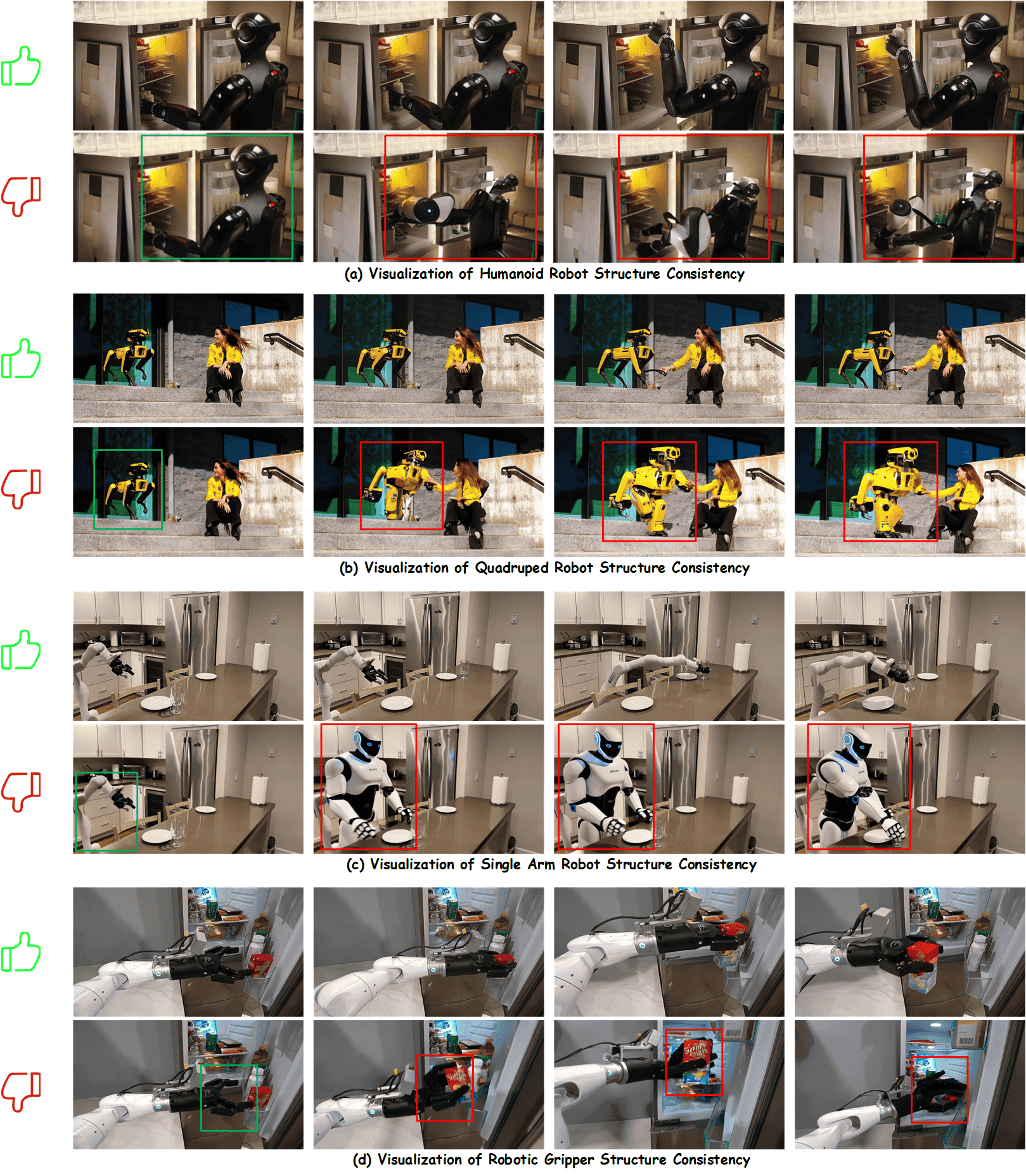}
    \caption{\textbf{Visualization of robot structural stability.}}
    \label{fig:robot_consistency}
\end{figure*}
\begin{figure*}[htbp]
    \centering
    \includegraphics[width=1.0\textwidth]{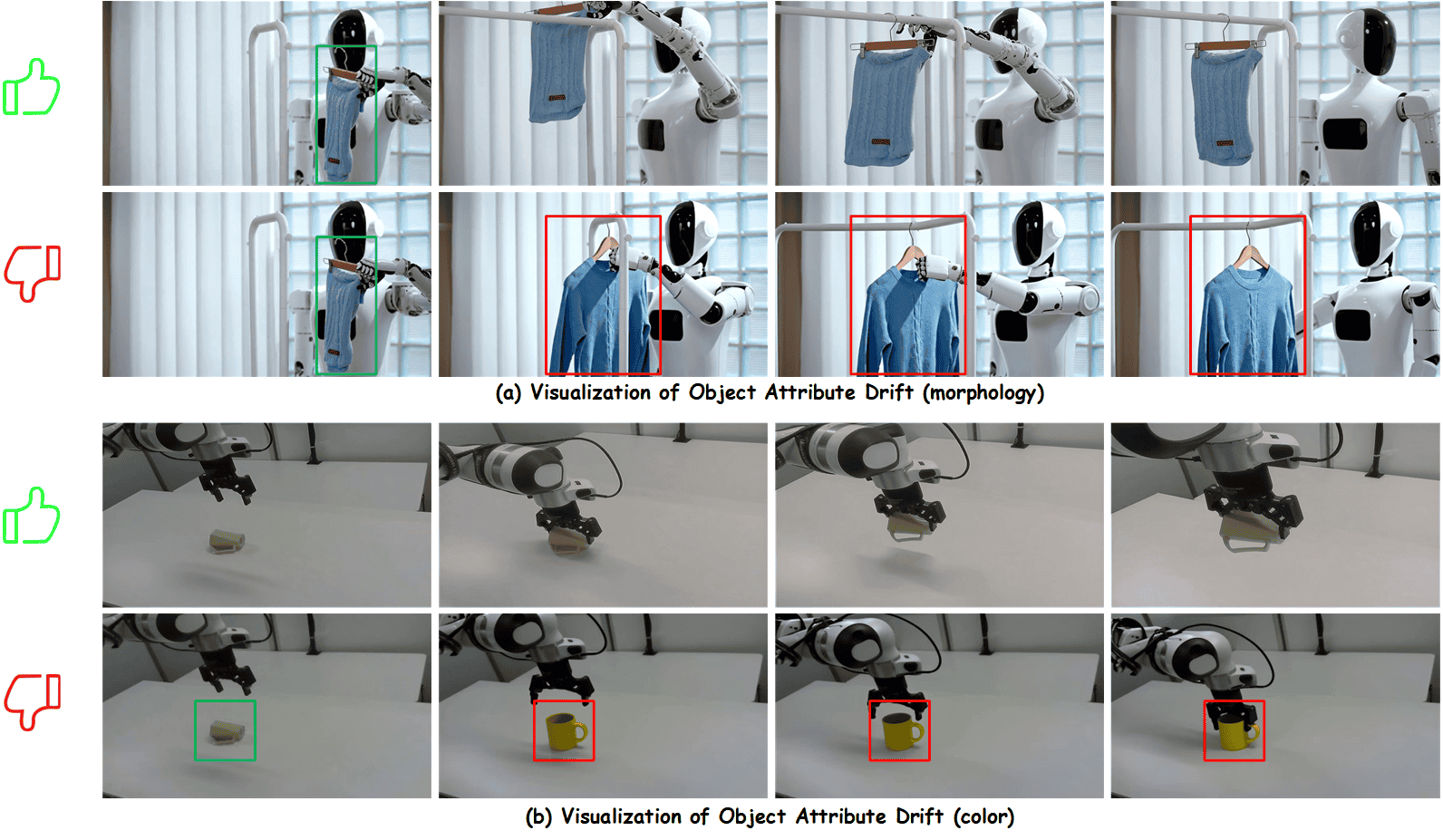}
    \caption{\textbf{Visualization of subject appearance stability.}}
    \label{fig:subject_consistency}
\end{figure*}

In robotics video generation, maintaining stable structure and appearance for both the robot and the manipulated objects is essential for assessing generation quality. In practice, models often exhibit abnormal changes in robot morphology or severe distortions of subject attributes. To systematically evaluate these issues, we propose the \emph{Robot-Subject Stability} metric, which separately measures the visual and semantic consistency of the robot and the target subject throughout the generation process.

We adopt a comparative VQA mechanism: the system simultaneously observes two frames, with the left frame as a reference image and the right frame as a generated frame, and focuses on a specified entity (e.g., the \emph{robotic gripper} or the \emph{target subject}). The MLLM is prompted to judge how well the entity's appearance, structure, and semantics are preserved between the two frames. Specifically, the evaluator identifies:

\begin{itemize}
    \item \textbf{Robot structural stability.} As shown in Figure~\ref{fig:robot_consistency}, (a) a humanoid robot degenerates into a single-arm robot, (b) a quadruped robot morphs into a small humanoid robot, (c) a single-arm robot transforms into a humanoid robot, and (d) a parallel gripper deforms into a dexterous robotic hand. These cases reveal structural drift and inconsistency in robot embodiment over time.
    \item \textbf{Subject appearance stability.} As illustrated in Figure~\ref{fig:subject_consistency}, (a) a rectangular knitted sleeve transforms into a long-sleeve sweater, and (b) a green plastic cup on the table becomes a round yellow mug, indicating a loss of identity-preserving appearance.
\end{itemize}

Beyond these explicit examples, we also observe a range of additional anomalies: changes in the number of robot links or arms during task execution, the spontaneous generation of extra manipulators, and unnatural variations in arm length, connectivity, or joint bending direction over time. Target objects may also undergo unrealistic material changes, such as a rigid object bending like a deformable one. The Robot-Subject Stability metric is designed to capture such inconsistencies, providing a focused measure of whether the model can preserve both robot morphology and object identity across the video sequence.

\subsection{Motion Amplitude}
\label{sec:motion_amplitude}
\begin{figure*}[htbp]
    \centering
    \includegraphics[width=1.0\textwidth]{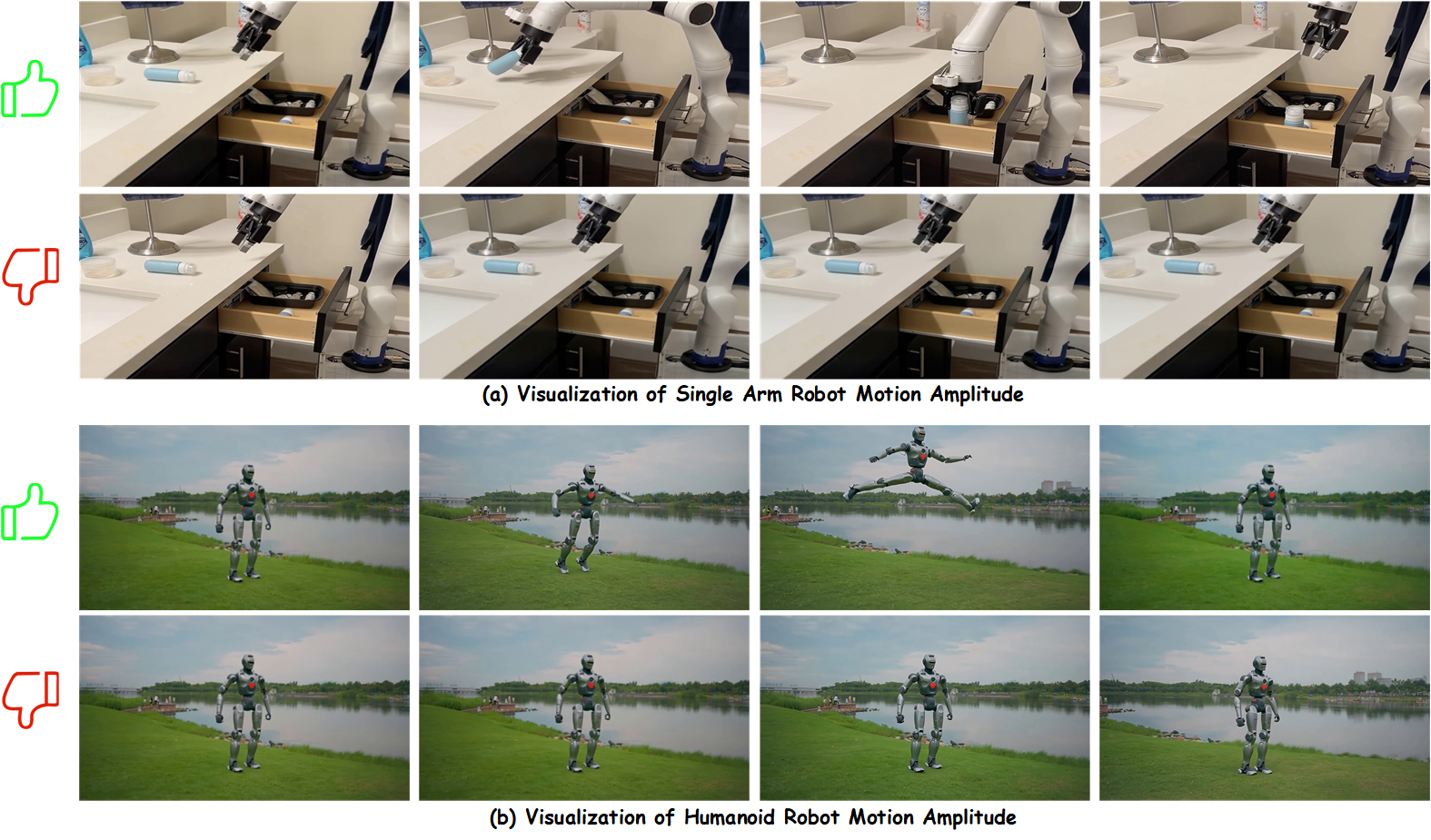}
    \caption{\textbf{Visualization of robot motion amplitude.}}
    \label{fig:motion_amplitude}
\end{figure*}
\noindent\myparagraph{Motivation.}A common failure mode in robotic video generation is that the robot remains nearly static while the generated frames appear visually smooth, as illustrated in Figure~\ref{fig:motion_amplitude}(a)(b).  
This makes pure smoothness-based metrics insufficient.  
Following the perceptual motion estimation idea introduced in VMBench~\cite{ling2025vmbench}, a \emph{Motion Amplitude Score (MAS)} is used to measure the perceptible dynamic behavior of the robot while explicitly compensating for camera motion.

\noindent\myparagraph{Robot Localization and Tracking.}The robot is first localized using GroundingDINO, and temporally stable segmentation masks are obtained via SAM2. CoTracker is then used to track a dense grid of keypoints inside the robot mask, ensuring that the estimated motion truly reflects robot articulation rather than background drift or mask leakage.

\noindent\myparagraph{Frame-Level Motion.}Let $\mathbf{p}_{t,k}$ denote the 2D location of the $k$-th tracked point at frame $t$.  
The raw frame-to-frame displacement is computed as:
\begin{equation}
\bar{D}_t = \frac{1}{K} \sum_{k=1}^{K} 
\bigl\lVert \mathbf{p}_{t,k} - \mathbf{p}_{t-1,k} \bigr\rVert_2 .
\label{eq:raw_motion}
\end{equation}
To ensure consistency across resolutions, the motion is normalized by the video diagonal:
\begin{equation}
\tilde{D}_t = 
\frac{\bar{D}_t}{\sqrt{W^{2} + H^{2}}} .
\label{eq:normalized_motion}
\end{equation}

\noindent\myparagraph{Camera-Motion Compensation.}To estimate camera-induced movement, the robot mask is inverted and the same tracking procedure is applied to the background region.
Let $\tilde{D}^{\mathrm{bg}}_t$ denote the normalized background motion.  

A \emph{soft-zero} strategy is adopted: if the robot motion does not exceed the background motion, the small residual value is retained:
\begin{equation}
\hat{D}_t =
\begin{cases}
\tilde{D}_t - \tilde{D}^{\mathrm{bg}}_t, & 
\tilde{D}_t > \tilde{D}^{\mathrm{bg}}_t, \\[4pt]
\tilde{D}_t, & 
\tilde{D}_t \le \tilde{D}^{\mathrm{bg}}_t .
\end{cases}
\label{eq:soft_zero}
\end{equation}
This behavior matches our implementation and improves robustness against tracking noise or partial occlusion, while effectively treating the robot as ``static''.

\noindent\myparagraph{Final Score.}Finally, following VMBench, the compensated displacement is clipped to stabilize extreme values:
\begin{equation}
\mathrm{MAS} =
\frac{1}{T}\sum_{t=1}^{T}
\min\bigl(\hat{D}_t,\,1\bigr).
\label{eq:mas_final}
\end{equation}

\noindent\myparagraph{Discussion.}MAS captures whether the robot exhibits meaningful articulation rather than merely inheriting background or camera movement.  
By incorporating localization, mask-based tracking, background compensation, and a soft-zero strategy, MAS remains stable across scenes, tracking configurations, and robotic embodiments.

\subsection{Motion Smoothness}
\label{sec:motion_smoothness}
\begin{figure*}[htbp]
    \centering
    \includegraphics[width=1.0\textwidth]{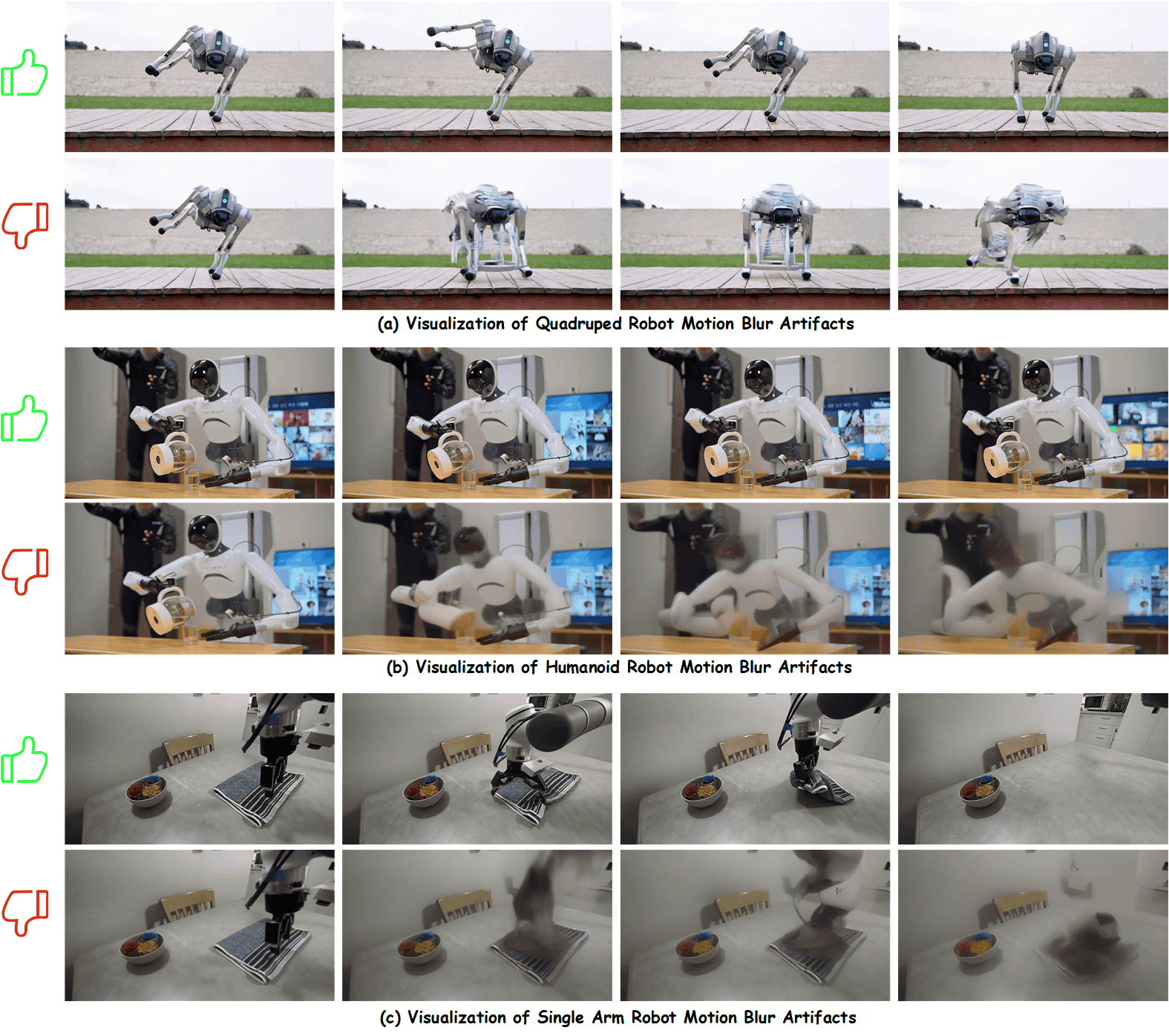}
    \caption{\textbf{Visualization of robot motion smoothness.}}
    \label{fig:motion_smoothness}
\end{figure*}
This metric evaluates the temporal continuity and naturalness of motion, aiming to detect frame-level discontinuities such as low-level temporal artifacts and high-level motion blur. As illustrated in Figure~\ref{fig:motion_smoothness}, various robot embodiments, including quadruped robots, humanoids, and single-arm manipulators, exhibit different degrees of motion-induced distortion that substantially degrade perceived video quality.

The assessment is based on the motion-smoothness principle introduced in VMBench~\cite{ling2025vmbench}, with temporal consistency estimated using Q-Align aesthetic quality scores. For each video, frames are processed with a sliding window of size $w$ (default $w = 3$). Each window is fed into Q-Align to obtain a per-frame quality score sequence $\{Q_t\}_{t=1}^{T}$. Temporal quality fluctuation is then measured by the magnitude of adjacent-frame differences:
\begin{equation}
\Delta Q_t = \lvert Q_t - Q_{t-1} \rvert, \qquad t = 2,\dots,T.
\end{equation}
To ensure comparability across videos with different motion intensities, the threshold for detecting abnormal temporal variations is determined by the \emph{Motion Amplitude} value $m$ defined in Section~\ref{sec:motion_amplitude}. A piecewise adaptive threshold function is used:
\begin{equation}
\tau_s(m) =
\begin{cases}
0.01,  & m < 0.1,\\[2pt]
0.015, & 0.1 \le m < 0.3,\\[2pt]
0.025, & 0.3 \le m < 0.5,\\[2pt]
0.03,  & m \ge 0.5.
\end{cases}
\end{equation}
Lower-motion videos therefore adopt a stricter threshold for detecting subtle temporal inconsistencies, while higher-motion videos receive a relaxed threshold to avoid penalizing naturally rapid movements. The function is determined through grid search on a validation split to ensure reproducibility.

A frame $t$ is marked as temporally abnormal if its score fluctuation exceeds the adaptive threshold:
\begin{equation}
I_t = \mathbb{I}[\Delta Q_t > \tau_s(m)],
\end{equation}
where $\mathbb{I}[\cdot]$ denotes the indicator function. To robustly capture abrupt artifacts such as frame drops or transient distortions, adjacent frames of each abnormal index are also flagged.

The final Motion Smoothness Score (MSS) is computed as the proportion of ``normal'' frames in the entire sequence:
\begin{equation}
\mathrm{MSS} = 1 - \frac{1}{T}\sum_{t=2}^{T} I_t.
\end{equation}
A higher MSS indicates smoother and more temporally coherent motion, whereas videos with frequent jitter, abrupt discontinuities, or artifact-heavy transitions yield lower MSS values.

\subsection{Score Aggregation}
\label{subsec:score_aggregation}

We consolidate five fine-grained evaluation signals into two final indicators, \emph{Task Completion} and \emph{Visual Quality}.

\noindent\myparagraph{Notation.}We denote the normalized values of the five fine-grained metrics as follows:
(1) \textbf{PSS}: Physical-Semantic Plausibility,
(2) \textbf{TAC}: Task-Adherence Consistency,
(3) \textbf{RSS}: Robot-Subject Stability,
(4) \textbf{MS}: Motion Smoothness,
(5) \textbf{MA}: Motion Amplitude.

\noindent\myparagraph{Normalization.}Given a raw metric value $s$ defined over range $[s_{\min}, s_{\max}]$, its normalized value is
\begin{equation}
    s \leftarrow \mathrm{clip}_{[0,1]}\!\left(
        \frac{s - s_{\min}}{s_{\max} - s_{\min}}
    \right).
\end{equation}

% -------------------------------------------------
\noindent\myparagraph{Penalty terms.}Two penalty terms are used to down-weight videos with insufficient subject motion or unstable visual composition.

Motion-amplitude penalty. Let MA denote the normalized motion amplitude. A soft penalty is applied when MA falls below the threshold $t$:
\begin{equation}
P_{\mathrm{MA}}(\mathrm{MA}) =
\begin{cases}
(t - \mathrm{MA}) + \delta, & \mathrm{MA} < t_{\mathrm{low}}, \\[3pt]
t - \mathrm{MA}, & t_{\mathrm{low}} \le \mathrm{MA} < t, \\[3pt]
0, & \mathrm{MA} \ge t,
\end{cases}
\label{eq:ma_penalty}
\end{equation}
with $t = 0.1$, $t_{\mathrm{low}} = 0.05$, and $\delta = 0.1$.

Stability-consistency penalty. Robot and object level stability grades are mapped to penalty magnitudes:
\begin{equation}
p(g) \in \{0.2, 0.4, 0.6, 0.8\}
\quad \text{for grades } g \in \{B, C, D, E\},
\end{equation}
while grade A incurs zero penalty.  
Let $g_r$ and $g_o$ denote robot- and object-related stability grades:
\begin{equation}
P_{\mathrm{RSS}} =
\begin{cases}
\dfrac{p(g_r) + p(g_o)}{2}, & \text{if both exist}, \\[6pt]
p(g_r), & \text{if only } g_r \text{ exists}, \\[4pt]
0, & \text{otherwise}.
\end{cases}
\label{eq:rss_penalty}
\end{equation}

% -------------------------------------------------
\noindent\myparagraph{Final indicators.}

Task Completion (TC). Task correctness is computed from Physical-Semantic Plausibility (PSS) and Task-Adherence Consistency (TAC):
\begin{equation}
    \mathrm{TC} = \frac{\mathrm{PSS} + \mathrm{TAC}}{2}.
    \label{eq:tr_final}
\end{equation}

Visual Quality (VQ). Visual realism and temporal coherence are expressed as a weighted combination of RSS and MS, penalized by low motion amplitude and visual instability:
\begin{equation}
\mathrm{VQ} =
\max\!\Bigl(
0,\;
0.8 \cdot \mathrm{RSS}
+ 0.2 \cdot \mathrm{MS}
- P_{\mathrm{MA}}(\mathrm{MA})
- P_{\mathrm{RSS}}
\Bigr).
\label{eq:vq_final}
\end{equation}

% -------------------------------------------------
\noindent\myparagraph{Model-level aggregation.}For each model, TC and VQ are computed for all evaluation samples.  
The final model score corresponds to the mean values of TC and VQ across samples,  
which are used for quantitative comparison and ranking in our benchmark.

\section{Model Descriptions and Implementation Setups}
\label{subsec: Setups}

\subsection{Commercial Models}
\noindent\myparagraph{Wan 2.6.}Wan is a comprehensive family of open-source foundational video generative models built on the Diffusion Transformer architecture. It supports multiple downstream tasks including T2V, I2V, editing, inpainting, and video-to-audio. We use the official API to generate 5-second 720P videos at 30 fps.

\noindent\myparagraph{Wan 2.5.}Wan is a comprehensive family of open-source foundational video generative models built on the Diffusion Transformer architecture. It supports multiple downstream tasks including T2V, I2V, editing, inpainting, and video-to-audio. We use the official API to generate 5-second 720P videos at 24 fps.

\noindent\myparagraph{Hailuo.}Hailuo provides multimodal models for T2V, I2V, and T2A tasks, supporting resolutions up to 1080p and long-duration outputs with high temporal coherence. We generate 6 second videos at 1364 × 768 and 24 fps using the official Hailuo API.

\noindent\myparagraph{Veo3.}Veo 3 is Google's latest foundational video generation model supporting high-resolution (1080p), long-duration (up to 60 seconds), and audio-integrated video synthesis using a large-scale Diffusion Transformer. We use the official Veo 3 API to generate videos up to 720p, 8 seconds, and 24 fps.

\noindent\myparagraph{Kling 2.6 pro.}  We use the official Kling 2.6 pro model with default parameters to generate a 5-second video at 1920 × 1080 resolution and 24 fps.

\noindent\myparagraph{Seedance 1.0.}Seedance 1.0 is a large-scale video generation model from ByteDance, supporting text-to-video and image-to-video generation with high aesthetic quality and temporal stability. It integrates a 3D causal VAE with a 4×16×16 compression ratio. We use the Seedance 1.0 model to generate 5-second videos at 1280 × 720 resolution and 24 fps.

\noindent\myparagraph{Seedance 1.5.}Seedance 1.5 is a large-scale video generation model from ByteDance, supporting text-to-video and image-to-video generation with high aesthetic quality and temporal stability. We use the Seedance 1.5 model to generate 5-second videos at 1280 × 720 resolution and 24 fps.

\noindent\myparagraph{Sora v1.}  We use the official Sora v1 model with default parameters to generate a 5-second video at 1280 × 720 resolution and 30 fps.

\noindent\myparagraph{Sora v2 Pro.}  We use the official Sora v2 Pro model with default parameters to generate a 4-second video at 1280 × 720 resolution and 30 fps.

\subsection{Open-source Models}
\noindent\myparagraph{Wan2.2\_A14B.}Wan2.2\_A14B is an open-source large-scale video generation model that incorporates a Mixture-of-Experts (MoE) architecture. This architecture dynamically allocates specialized expert networks to enhance the model's capacity and temporal understanding. It supports multimodal inputs (text and images) for generating open-domain videos at 1280 × 720 resolution (720P), with a typical duration of 5 seconds (120 frames) at 24 fps. Compared to its predecessors, Wan2.2\_A14B demonstrates superior performance in generalizing across diverse scenes, modeling complex motions, and achieving fine-grained aesthetic control. The technical report and model weights are publicly available. We utilize the official Wan2.2\_A14B model with its default parameters to generate a 5-second video (81 frames) at a spatial resolution of 1280 × 720 and a frame rate of 16 fps. The model supports both text and image inputs.

\noindent\myparagraph{LongCat-Video.}LongCat-Video is an open-source foundational video generation model with 13.6B parameters, developed by the Meituan LongCat Team. It unifies text-to-video (T2V), image-to-video (I2V), and video continuation (VC) tasks within a single Diffusion Transformer (DiT) architecture, supporting efficient minute-long video generation without quality degradation. The model employs multi-reward RLHF optimization (Group Relative Policy Optimization) to enhance visual quality, motion coherence, and text alignment. We employ the official LongCat-Video model with default configurations, generating 1280 × 704 resolution videos at 15 fps using a coarse-to-fine generation strategy.

\noindent\myparagraph{Wan2.2\_5B.}Wan2.2\_5B is a medium-scale model in the Wan2.2 series with 5 billion parameters, utilizing a Transformer backbone architecture. It supports multi-modal video generation from text and images, generating 5-second (121 frames) open-domain videos at 1280 × 720 and 24 fps. Model weights and documentation are fully open-sourced. We use the official Wan2.2\_5B model with default settings to generate a 5-second video (120 frames) at a resolution of 1248 × 704 and 24 fps.

\noindent\myparagraph{Wan2.1\_14B.}Wan2.1\_14B is an early large-parameter video generation model in the Wan series with 14 billion parameters, based on a multi-modal diffusion architecture. It generates 5-second (120-frame) open-domain videos at 1280 × 720 and 24 fps, emphasizing complex scene modeling and object motion understanding. We use the official Wan2.1\_14B model with default parameters to generate a 5-second video (81 frames) at 832 × 480 resolution and 16 fps.

\noindent\myparagraph{HunyuanVideo.}  We use the official HunyuanVideo model with default parameters to generate a 5-second video at 1248 × 704 resolution and 24 fps.

\noindent\myparagraph{HunyuanVideo 1.5.}  We use the official HunyuanVideo 1.5 model with default parameters to generate a 5-second video at 848 × 480 resolution and 24 fps.

\noindent\myparagraph{SkyReels-V2.}SkyReels-V2 is an open-source infinite-length film generative model developed by Skywork AI. It supports text-to-video, image-to-video, and video continuation tasks, with modules including SkyCaptioner-V1, multi-stage pretraining, RL for motion quality, and a diffusion forcing framework for long video synthesis. We use the official SkyReels-V2 model with default settings to generate 960 × 544 resolution videos at 24 fps for approximately 4 seconds.

\noindent\myparagraph{LTX-Video.}LTX-Video is an open-source transformer-based latent diffusion model developed by Lightricks. It integrates a Video-VAE and denoising transformer with a 1:192 compression ratio using spatiotemporal downscaling, enabling efficient latent-space processing. It supports both text-to-video and image-to-video generation. We use the official LTX-Video model with default parameters to generate a 5-second video at 832 × 480 resolution and 16 fps.

\noindent\myparagraph{LTX-2.}LTX-2 is an open-source transformer-based latent diffusion model. We use the official LTX-2 model with default parameters to generate a 5-second video at 1536 × 1024 resolution and 24 fps.

\noindent\myparagraph{FramePack.}FramePack is a neural structure for next-frame prediction designed to avoid forgetting and drifting in video generation. It compresses input frames by importance to maintain a fixed transformer context, enabling long video synthesis with low computational cost. It supports T2V and I2V and can be integrated with models such as HunyuanVideo or Wan. We employ FramePack with a base model to generate 5-second videos at 832 × 480 resolution.

\noindent\myparagraph{CogVideoX.}CogVideoX is an open-source text-to-video generation model employing a diffusion transformer and a 3D VAE. It introduces expert-adaptive LayerNorm and multi-resolution frame packing to enhance motion consistency. It produces up to 10-second videos at 720 × 1280 and 16 fps.
We use the official CogVideoX weights and inference scripts with default settings to generate 6-second videos (160 frames) at 720 × 480 and 8 fps.

\subsection{Robotics-specific Models}
\noindent\myparagraph{Cosmos 2.5.}Cosmos-Predict2.5 and Cosmos-Transfer2.5 are NVIDIA’s world simulation foundation models for physical AI, supporting Text2World, Image2World, and Video2World simulations with flow-based architectures.
We generate videos at 1280 × 720 resolution and 16 fps using official NVIDIA codebase.

\noindent\myparagraph{DreamGen (GR).}GRL is DreamGen’s generalizable robot learner model integrating synthetic neural trajectories and real teleoperation data for policy learning. We generate all trajectory videos at 768 × 432 resolution and 16 fps.

\noindent\myparagraph{DreamGen (DROID).}DROID is a dynamics-aware imitation learning model using video diffusion with inverse dynamics modeling for realistic physical trajectories.
We generate all videos at up to 768 × 432 resolution and 16 fps.

\noindent\myparagraph{UnifoLM-WMA-0.}UnifoLM-WMA-0 integrates video world modeling and action policy learning for generalizable robotic reasoning and control.
We generate videos up to 320 × 512 resolution, 30 fps.

\noindent\myparagraph{Vidar.}Vidar is an embodied video diffusion model designed for robotic manipulation, integrating a strong video diffusion prior and a Masked Inverse Dynamics Model (MIDM).
Vidar generates 4-second multi-view interactive videos at 704 × 480 resolution and 15 fps with automatically annotated action trajectories.

\section{Human Preference Study Details}
\label{sec:human_preference}

\begin{figure}[t]
  \centering
  \includegraphics[width=0.7\linewidth]{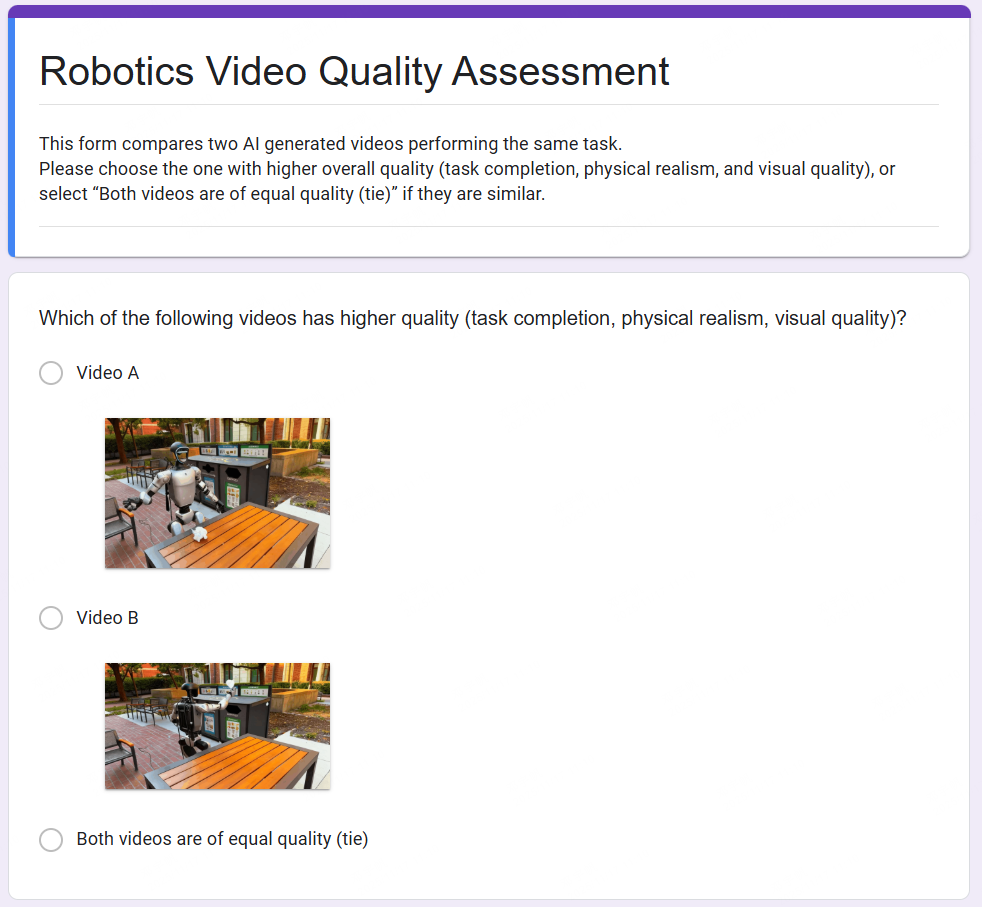}
  \caption{\textbf{Visualization of the Questionnaire for User Study}}
  \label{fig:user_study}
\end{figure}

To complement the main paper's human preference analysis, this section provides additional statistical examinations that further characterize the agreement between human judgments and our automatic benchmark. The questionnaire used in our human preference study is shown in Figure~\ref{fig:user_study}. As described in the main paper, thirty participants compared pairs of generated videos for the same prompt and selected the better one (or "Tie"). These votes were converted into per-model human scores using the \(5/3/1\) win-tie-loss scheme, and the resulting model-level ranking exhibited a strong correlation with RBench (\(\rho = 0.96\)).

While rank correlation quantifies the consistency of \emph{ordering}, it does not measure whether the two scoring methods agree in an \emph{absolute} sense (i.e., whether they assign comparable magnitudes). To assess this complementary notion of agreement, we conduct a Bland-Altman analysis between human scores and benchmark scores. Because the two score scales may differ by a systematic offset or slope, we additionally apply a leave-one-out (LOO) linear calibration to correct for potential scale mismatch without overfitting. The following subsections detail the calibration procedure, Bland-Altman computation, accompanying figures, and per-model statistics.

% ---------------- Bland–Altman after LOO calibration ----------------
\begin{figure}[t]
  \centering
  \includegraphics[width=0.75\linewidth]{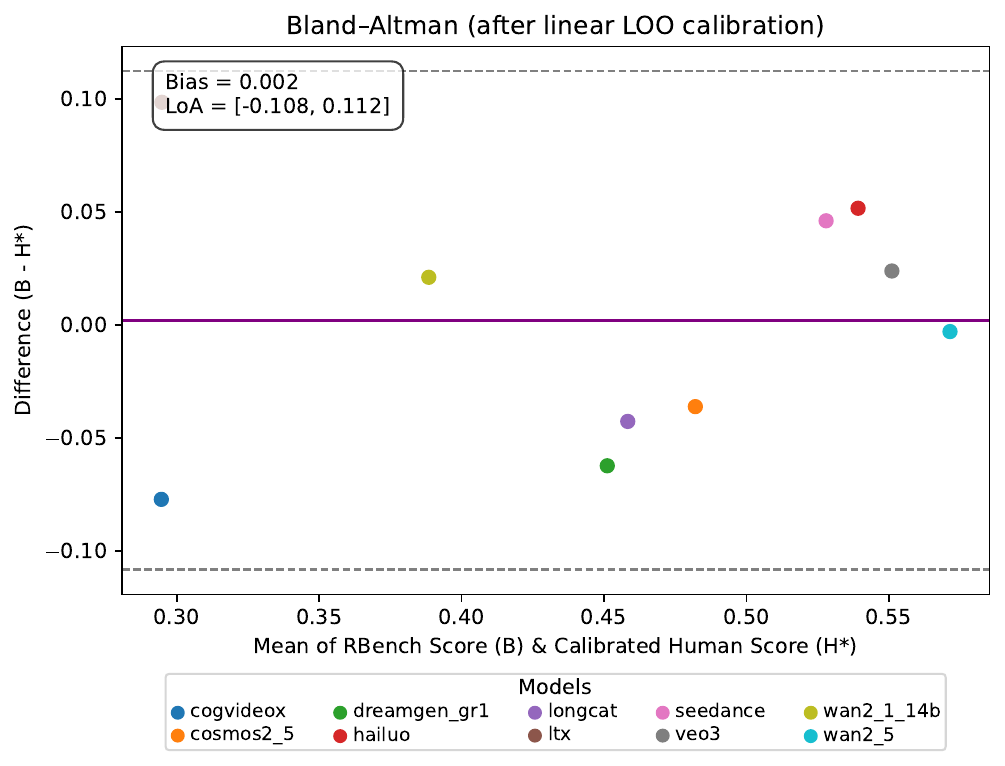}
  \caption{Bland--Altman plot after linear leave-one-out calibration (\(H^{\ast}\)).
  Points are models; x-axis \(m_i=\tfrac{B_i + H_i^{\ast}}{2}\), y-axis \(d_i=B_i-H_i^{\ast}\).
  The solid line indicates the bias (\(\bar d\)); dashed lines show the 95\% limits of agreement (LoA).
  In our study the legend reports \(\text{Bias}=0.002\) and \(\text{LoA}=[-0.108,\,0.112]\).}
  \label{fig:ba-cal}
\end{figure}

\paragraph{Bland--Altman basics.}
Bland--Altman analysis assesses \emph{agreement} between two measurement methods by plotting, for each item \(i\), the difference \(d_i = B_i - H_i\) against the mean \(m_i = \tfrac{B_i + H_i}{2}\), where \(H\) and \(B\) denote human and automatic scores, respectively.
As shown in Fig.~\ref{fig:ba-cal}, the horizontal solid line represents the \emph{bias} \(\bar d = \tfrac{1}{n}\sum_i d_i\) (average difference).
The dashed lines denote the \emph{95\% limits of agreement (LoA)}: \(\bar d \pm 1.96\, s_d\), where \(s_d\) is the standard deviation of the differences.
Narrower LoA and small bias indicate stronger agreement beyond mere correlation.

\paragraph{Linear LOO calibration.}
Human and automatic scores can exhibit \emph{scale} mismatch. 
To mitigate such systematic differences, we apply a \emph{leave-one-out (LOO) linear calibration} to the benchmark.

\noindent\textbf{Mathematical formulation.}
Let \(\{(B_i, H_i)\}_{i=1}^{n}\) be the automatic benchmark and human scores for \(n\) models.
For each \(i \in \{1,\dots,n\}\), define the training index set \(S_{-i}=\{j:\, j\neq i\}\) and estimate the OLS (Ordinary Least Squares) calibration parameters by
\begin{equation}
(\hat{\alpha}_{-i}, \hat{\beta}_{-i})
~=~
\arg\min_{\alpha,\beta}
\sum_{j\in S_{-i}} \bigl(B_j - \alpha - \beta H_j\bigr)^2 .
\label{eq:loo-fit}
\end{equation}
The calibrated (benchmark-aligned) score for model \(i\) is
\begin{equation}
H_i^{\ast} \;=\; \hat{\alpha}_{-i} + \hat{\beta}_{-i}\, B_i .
\label{eq:loo-cal}
\end{equation}

\noindent\textbf{Bland--Altman with calibrated scores.}
We then perform Bland--Altman analysis on \((B, H^{\ast})\) by computing, for each \(i\),
\begin{equation}
d_i \;=\; B_i - H_i^{\ast},
\qquad
m_i \;=\; \tfrac{B_i + H_i^{\ast}}{2}.
\label{eq:ba-vars}
\end{equation}
The bias \(\bar d\) and the 95\% LoA \(\bar d \pm 1.96\, s_d\) are computed from \(\{d_i\}_{i=1}^{n}\) as in the basics above.

Figure~\ref{fig:ba-cal} shows the Bland-Altman plot after linear LOO calibration, with the bias and LoA reported in the legend.  
Table~\ref{tab:ba-cal} lists, for each model, the mean \(m_i\), the difference \(d_i\), and the corresponding LOO calibration parameters \(\hat{\alpha}_{-i}\) and \(\hat{\beta}_{-i}\) used to obtain \(H_i^{\ast}\).  
As a complement to the correlation analysis (Spearman \(\rho = 0.96\) between human scores and the automatic benchmark), the Bland-Altman view quantifies absolute agreement: the small bias and tight LoA indicate that the calibrated benchmark scores are in close agreement with human judgments.

% ---------------- Table: Bland–Altman statistics (3 d.p.) ----------------
\begin{table}[t]
  \centering
  \scriptsize
  \setlength{\tabcolsep}{5pt}
  \renewcommand{\arraystretch}{1.15}
  \begin{tabular}{lrrrr}
    \toprule
    \textbf{Model} & \(\mathrm{mean}(B, H^{\ast})\) & \( \mathrm{diff}(B - H^{\ast}) \) & \( \alpha_{\mathrm{LOO}} \) & \( \beta_{\mathrm{LOO}} \) \\
    \midrule
    LTX-Video            & 0.295 &  0.098 & 0.141 & 0.613 \\
    Hailuo         & 0.539 &  0.052 & 0.205 & 0.496 \\
    Seedance 1.0       & 0.528 &  0.046 & 0.203 & 0.500 \\
    Veo 3           & 0.551 &  0.024 & 0.204 & 0.503 \\
    Wan2.1 14b   & 0.388 &  0.021 & 0.193 & 0.524 \\
    Wan2.5        & 0.571 & -0.003 & 0.199 & 0.517 \\
    Cosmos2.5     & 0.482 & -0.036 & 0.200 & 0.523 \\
    LongCat-Video        & 0.458 & -0.043 & 0.202 & 0.520 \\
    Dreamgen  & 0.451 & -0.062 & 0.203 & 0.522 \\
    Cogvideox-5B      & 0.295 & -0.077 & 0.242 & 0.446 \\
    \bottomrule
  \end{tabular}
  \caption{Bland--Altman statistics after linear LOO calibration.
  \(m_i=\tfrac{B_i+H_i^{\ast}}{2}\), \(d_i=B_i-H_i^{\ast}\).
  \(\alpha_{\mathrm{LOO}}\) and \(\beta_{\mathrm{LOO}}\) are the intercept and slope learned from all \emph{other} models when calibrating \(H_i^{\ast}\).
  All values are rounded to three decimals.}
  \label{tab:ba-cal}
\end{table}
\section{Prompt Template}
\label{sec:prompt_template}

This section introduces the prompt design used for the Visual Reasoning task in our MLLM-based evaluation pipeline. Among all tasks, Visual Reasoning involves the most structured form of reasoning and thus provides a representative example for illustrating our prompt design. We provide the MLLM with essential contextual information recorded during dataset construction, including the video viewpoint, a high-level content description, and the identities of the robotic manipulator and manipulated object. This background knowledge serves as the foundation for the model’s subsequent reasoning and scoring process.

Unlike other tasks, Visual Reasoning requires explicit verification of logical dependencies between robot actions. Therefore, it adopts a two-part structure consisting of a question-chain constructor and a video assessment prompt. The full templates are shown below.
\begin{figure*}[t]
  \centering
  \includegraphics[width=\linewidth]{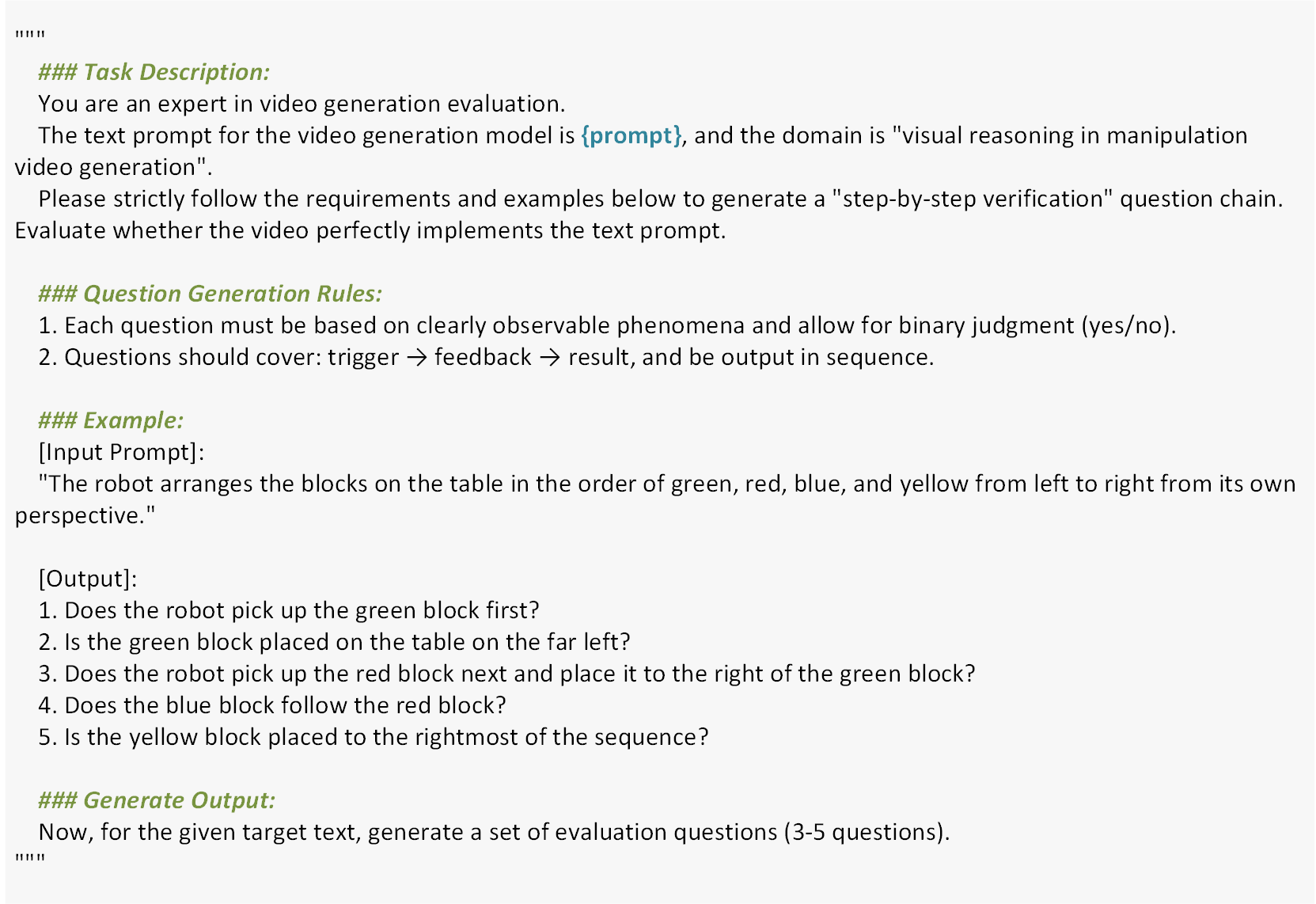}
  \caption{\textbf{Question-chain construction for Visual Reasoning.}  
  This component analyzes the original instruction and transforms it into a sequence of binary verification questions that define causal and temporal dependencies in the robot’s intended actions.}
  \label{fig:vr_question_chain}
\end{figure*}
\paragraph{Question-chain construction.}The first component converts the original text instruction into a sequence of binary verification questions (Figure~\ref{fig:vr_question_chain}).  
The model analyzes the semantics of the instruction and generates a short chain of stepwise questions that reflect the intended causal and temporal structure of the robot’s behavior.  
This question chain acts as an explicit reasoning scaffold and is subsequently fed into the main evaluation prompt.

\begin{figure*}[t]
  \centering
  \includegraphics[width=\linewidth]{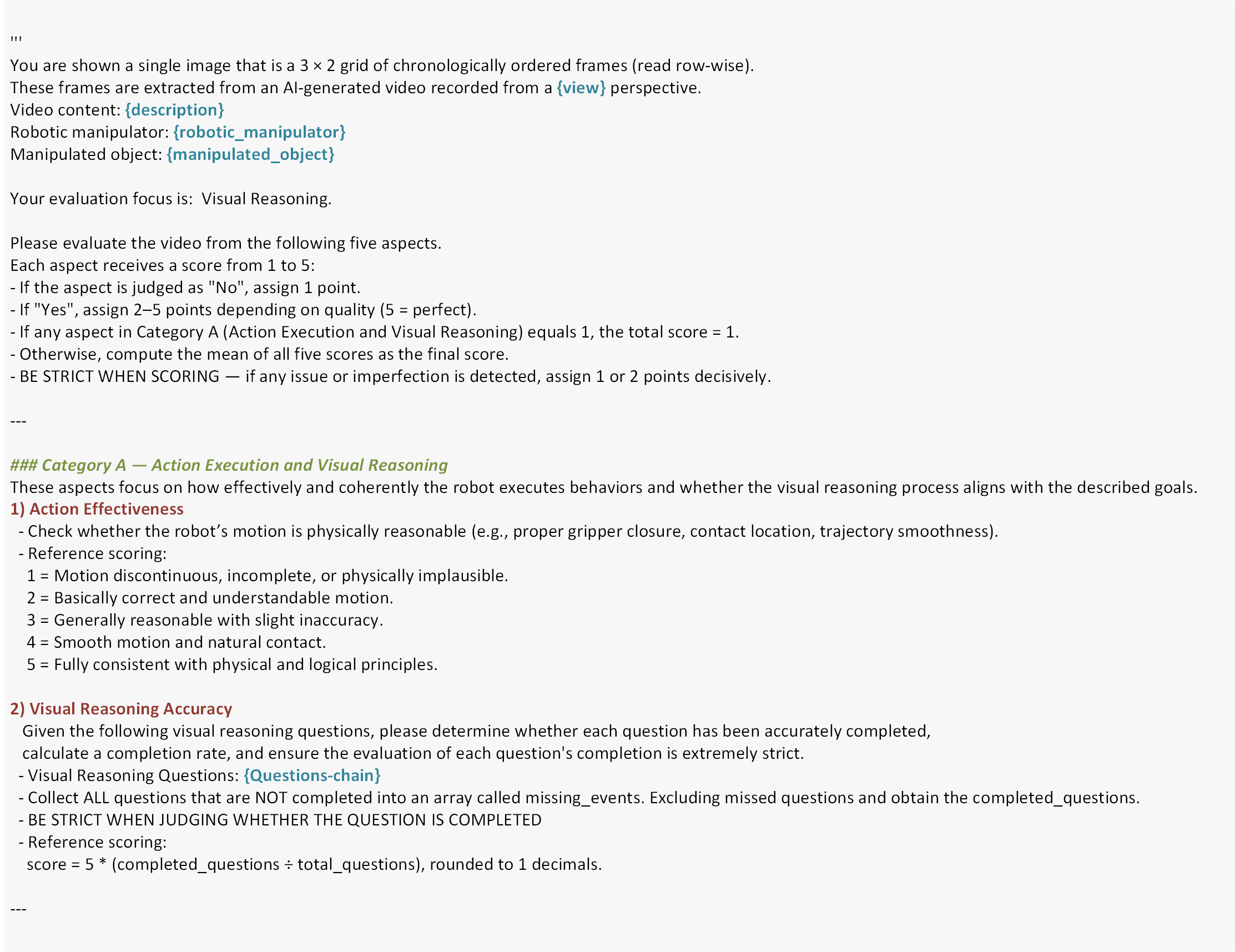}
  \caption{\textbf{Visual Reasoning evaluation prompt (Part I).}  
  The first part of the evaluation prompt integrates the structured reasoning chain with contextual information extracted during dataset construction.}
  \label{fig:vr_main_prompt_1}
\end{figure*}
\begin{figure*}[t]
  \centering
  \includegraphics[width=\linewidth]{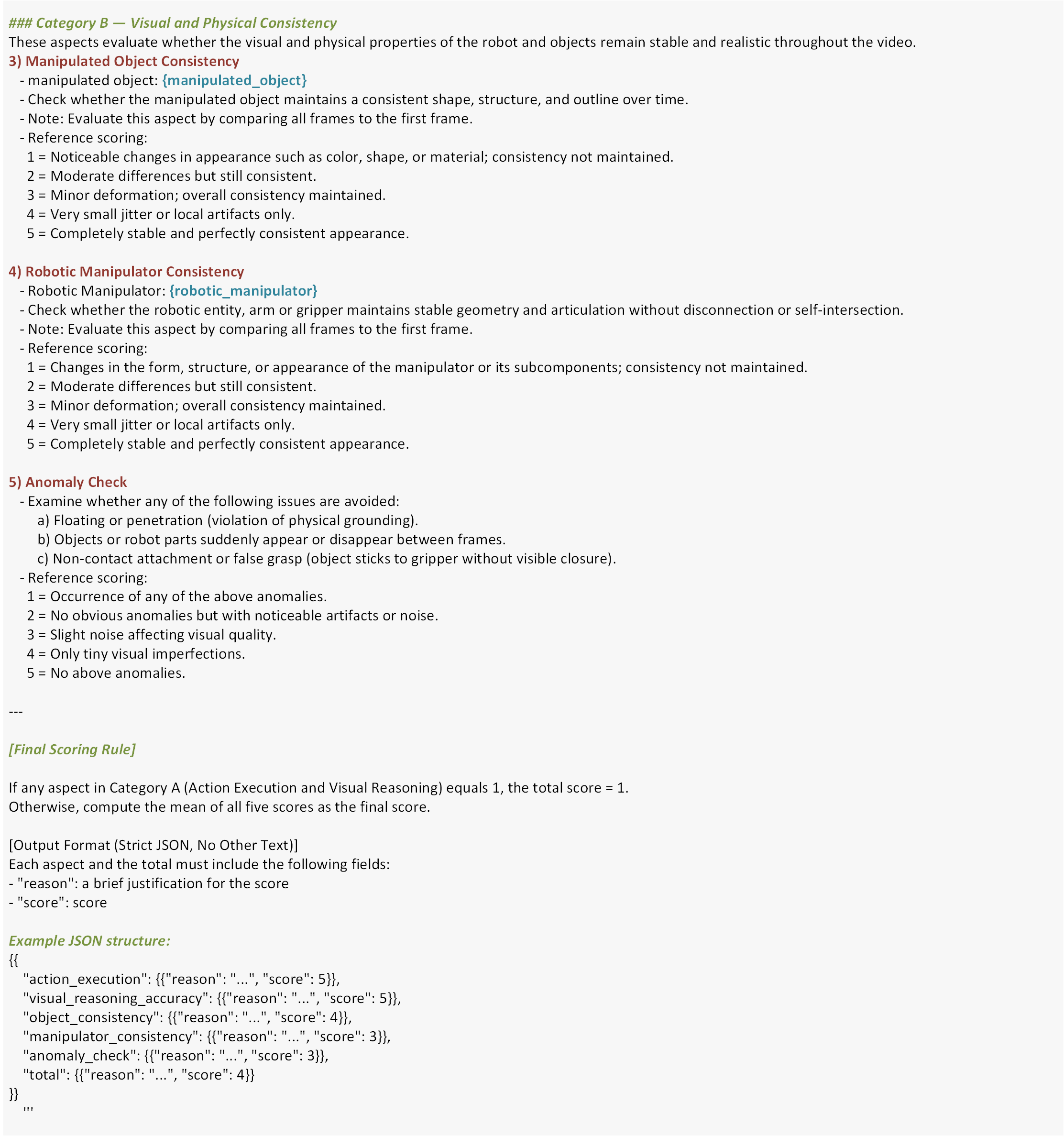}
  \caption{\textbf{Visual Reasoning evaluation prompt (Part II).}  
  The second part specifies the structured scoring protocol and the JSON output format used to ensure consistent and interpretable evaluation results.}
  \label{fig:vr_main_prompt_2}
\end{figure*}

\paragraph{Video assessment prompt.}The second component integrates the generated question chain with a 3×2 grid of chronologically ordered frames extracted from the video (Figures~\ref{fig:vr_main_prompt_1} and \ref{fig:vr_main_prompt_2}).  
It also incorporates the contextual background information provided at the beginning—namely, the view perspective, video content summary, robotic manipulator type, and manipulated object identity.  
With these inputs, the prompt instructs the model to determine whether each reasoning step has been successfully completed and to evaluate the stability, consistency, and physical plausibility of the robot and objects throughout the sequence.  
A strict scoring protocol and a structured JSON output format ensure reproducible and interpretable evaluation across models.

% ------------------------------------------  5 TASKS -------------------------------------------------
\section{Additional Qualitative Comparisons}
\label{subsec: Qualitative}

\begin{figure*}[t]
    \centering
    \includegraphics[width=\textwidth]{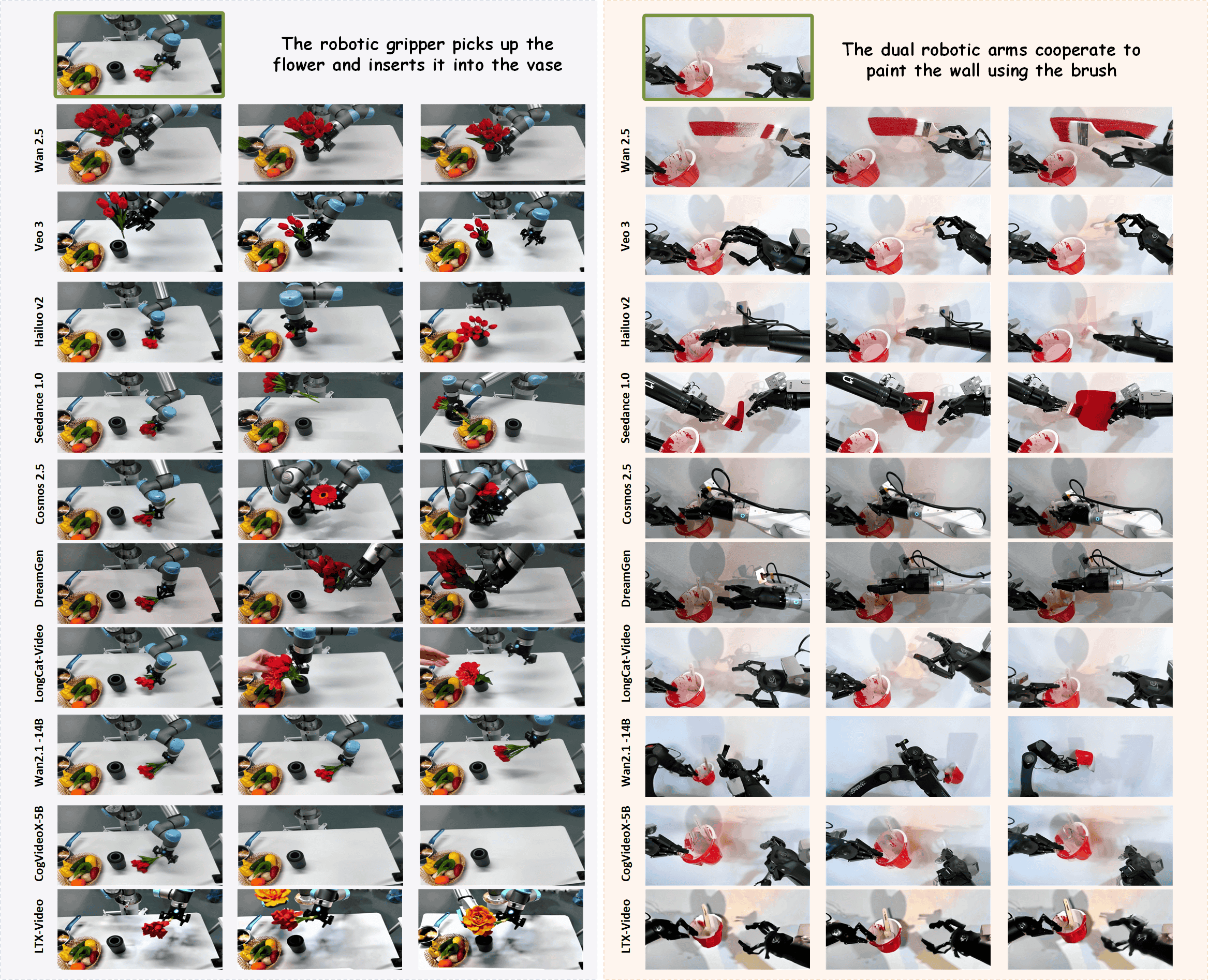}
    \caption{\textbf{Visualization of Common Manipulation.} The first row contains the reference image, and the accompanying text shows the input prompt.
}
    \label{fig:common_manipulation}
\end{figure*}

\begin{figure*}[t]
    \centering
    \includegraphics[width=\textwidth]{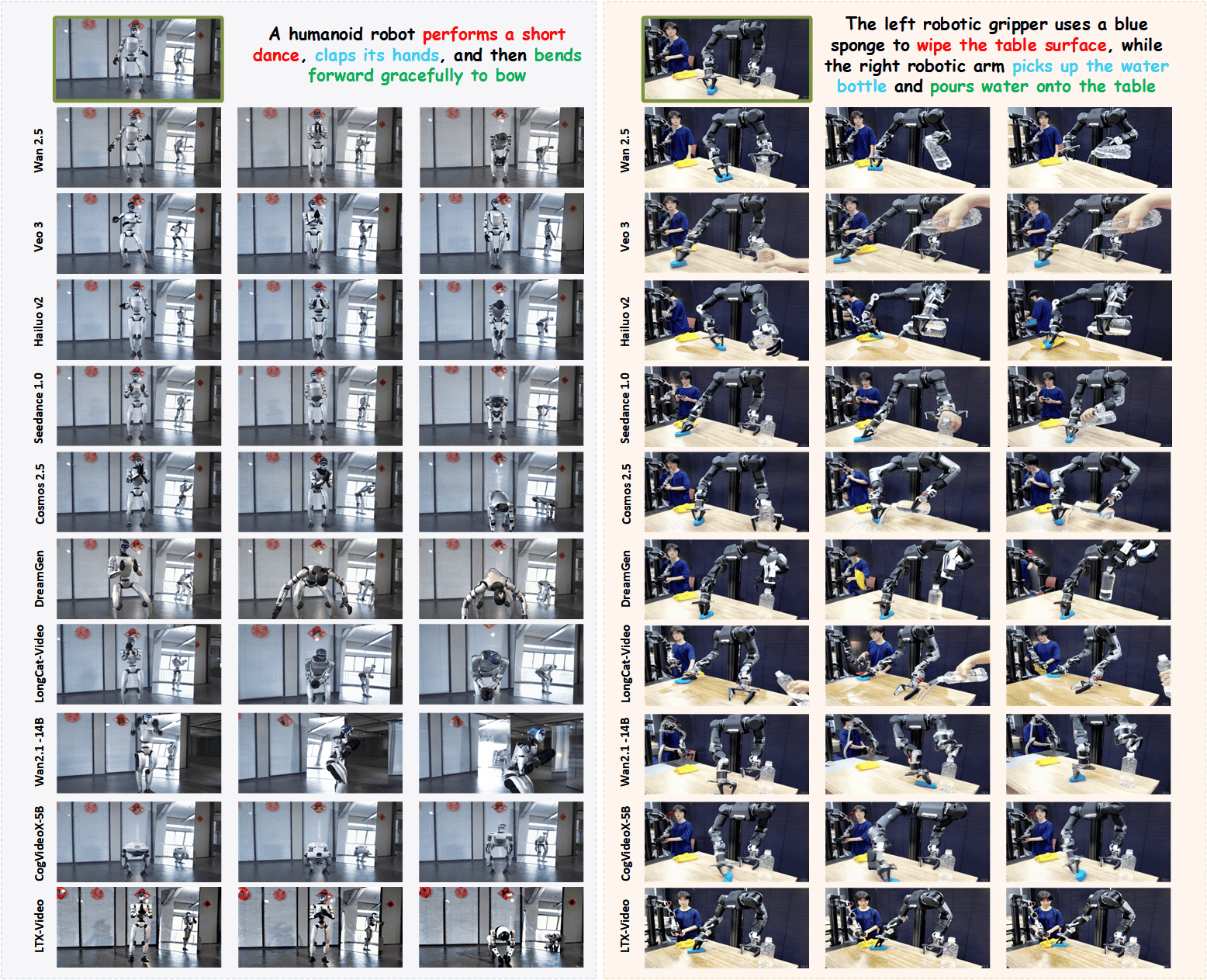}
    \caption{\textbf{Visualization of Long-Horizon Planning.} The first row contains the reference image, and the accompanying text shows the input prompt.}
    \label{fig:long_horizon_planning}
\end{figure*}

\begin{figure*}[t]
    \centering
    \includegraphics[width=\textwidth]{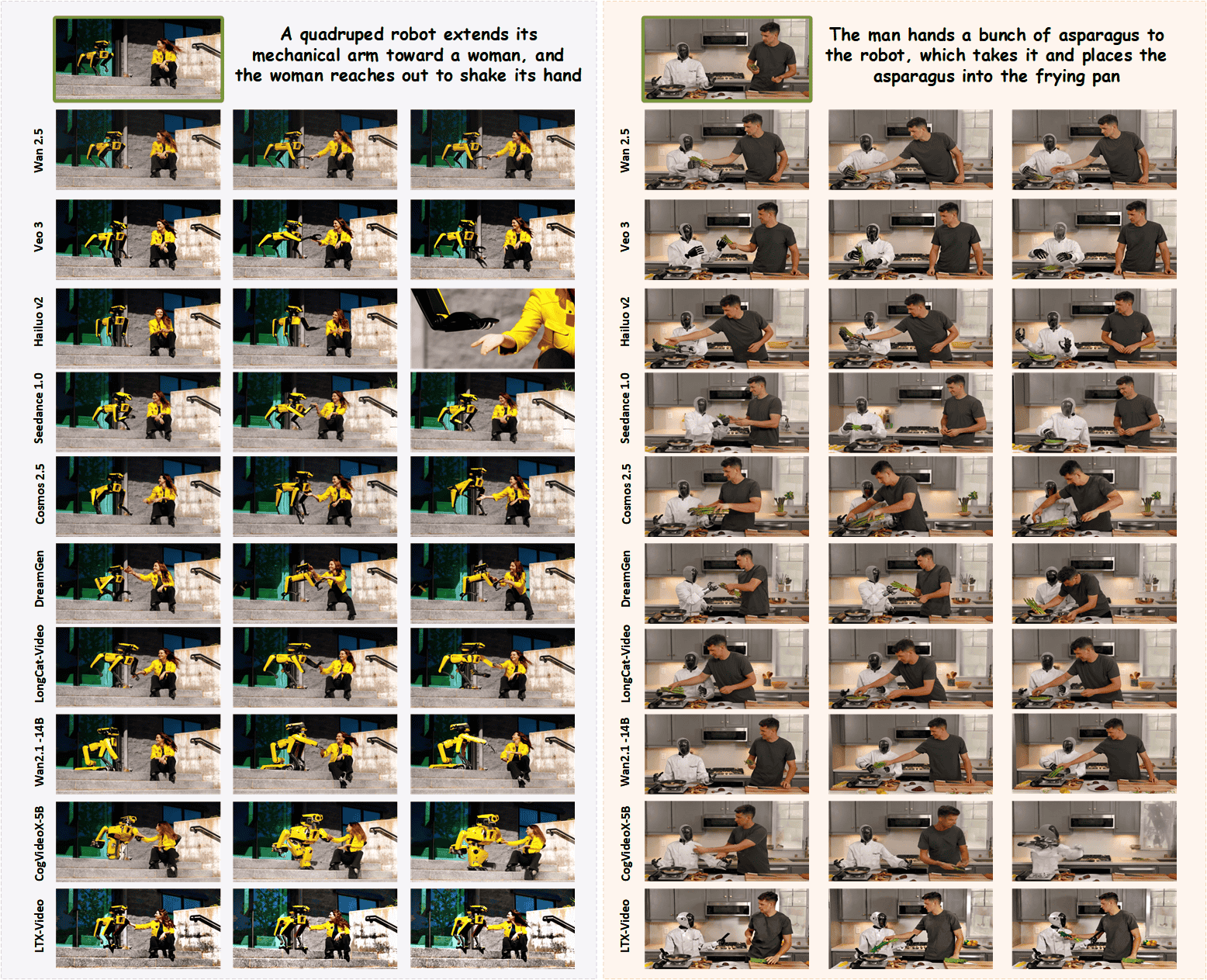}
    \caption{\textbf{Visualization of Multi-Entity Collaboration.} The first row contains the reference image, and the accompanying text shows the input prompt.}
    \label{fig:multi_entity_collaboration}
\end{figure*}

\begin{figure*}[t]
    \centering
    \includegraphics[width=\textwidth]{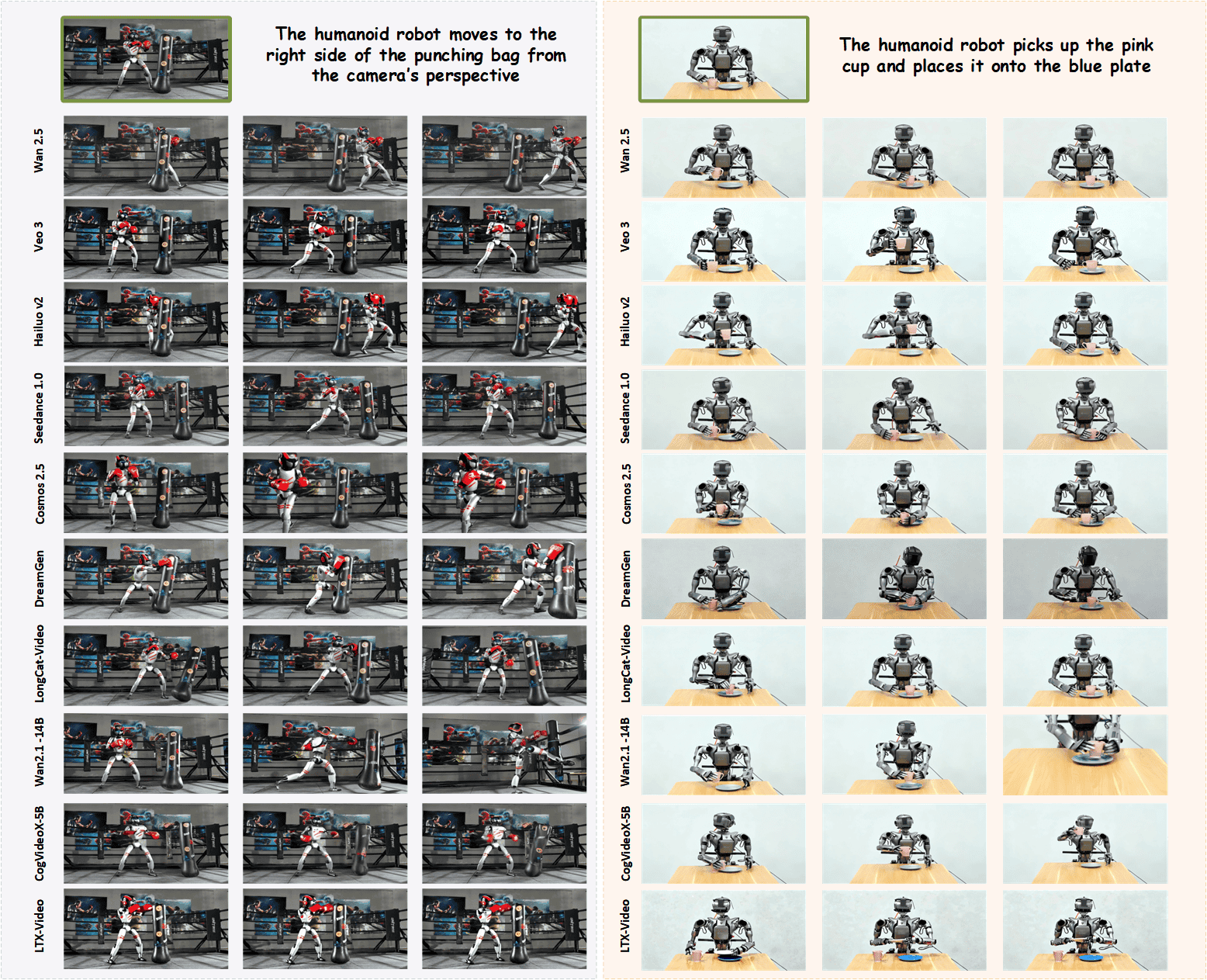}
    \caption{\textbf{Visualization of Spatial Relationship.} The first row contains the reference image, and the accompanying text shows the input prompt.}
    \label{fig:spatial_relationship}
\end{figure*}

\begin{figure*}[t]
    \centering
    \includegraphics[width=\textwidth]{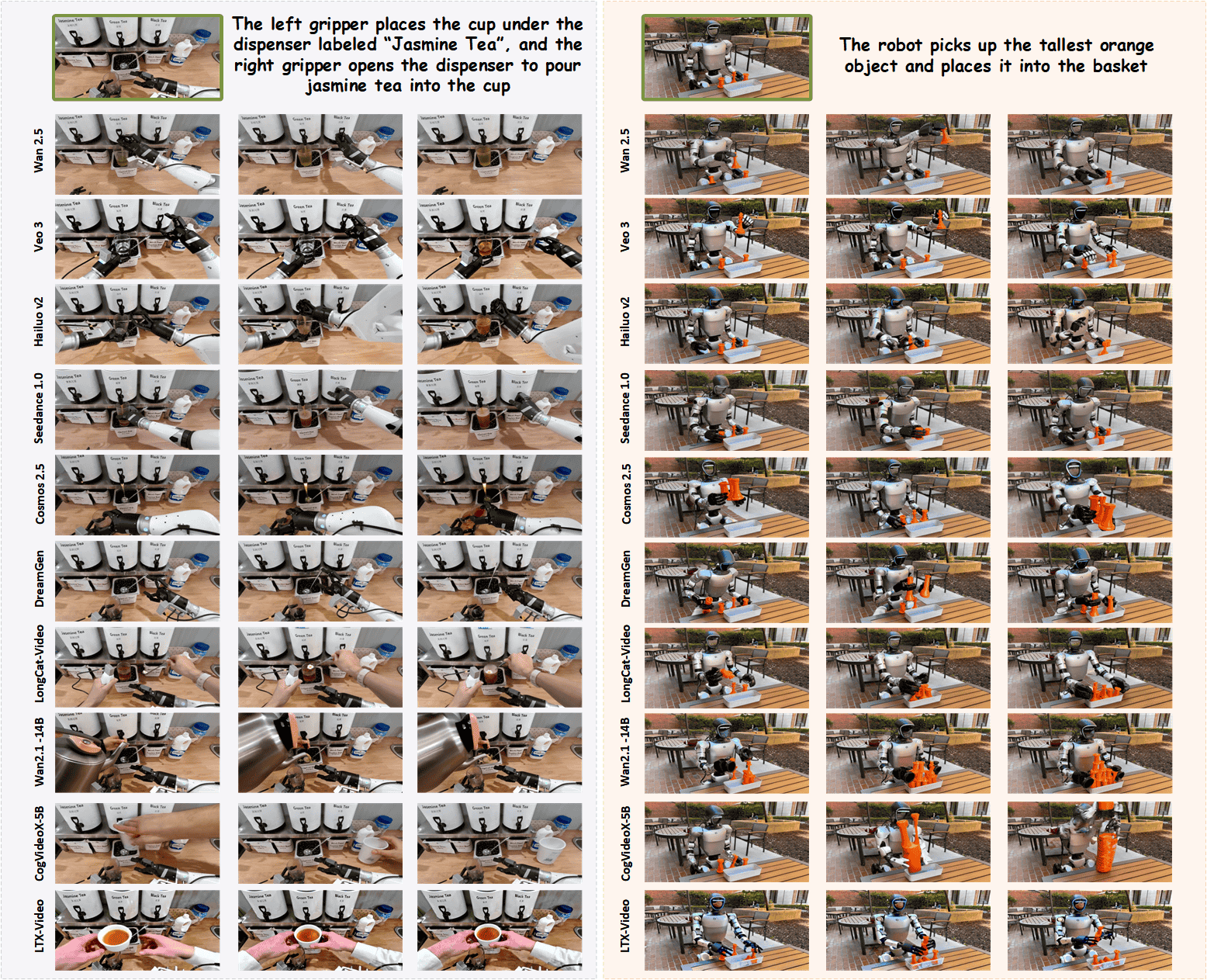}
    \caption{\textbf{Visualization of Visual Reasoning.} The first row contains the reference image, and the accompanying text shows the input prompt.}
    \label{fig:visual_reasoning}
\end{figure*}

This section provides supplementary qualitative results that extend the "Qualitative Comparison Across Representative Tasks" analysis presented in the main paper. For each of the five task categories in RBench, namely Common Manipulation, Long-Horizon Planning, Multi-Entity Collaboration, Spatial Relationship, and Visual Reasoning, we select two representative cases and visualize the generated videos from ten state-of-the-art image-to-video (I2V) models. As shown in Figures~\ref{fig:common_manipulation}, \ref{fig:long_horizon_planning}, \ref{fig:multi_entity_collaboration}, \ref{fig:spatial_relationship}, \ref{fig:visual_reasoning}, these examples offer a more detailed inspection of model behaviors under diverse embodied scenarios, complementing the quantitative results discussed in the main text.

\section{Comprehensive Quantitative Results}
\label{subsec: Quantitative}

This section presents the complete quantitative evaluation results obtained from both GPT-based and Qwen-based evaluators. We report detailed per-model scores across all five tasks (Common Manipulation, Long-Horizon Planning, Multi-Entity Collaboration, Spatial Relationship, and Visual Reasoning) and four robot embodiments (Dual Arm, Humanoid, Single Arm, and Quadruped).  
To ensure clarity and readability, we summarize below the abbreviations used in all tables. For task-level metrics, each abbreviation corresponds to a specific VQA-style prompt used in our MLLM-based evaluation; the construction of these prompts and the computation of the total score (\textbf{TS}) are illustrated with representative example in Section~\ref{sec:prompt_template}. The robot-level metrics are defined in Section~\ref{subsec:evaluation_metrics}.

\vspace{2mm}
\noindent\textbf{Task-level Metrics (per-task evaluation):}
\begin{itemize}[leftmargin=10pt, itemsep=1pt]
    \item \textbf{AES}: Action Execution Score  
    \item \textbf{TCS}: Task Completion Score  
    \item \textbf{OCS}: Object Consistency Score  
    \item \textbf{RCS}: Robot Consistency Score  
    \item \textbf{PSS}: Physical Semantic Score
    \item \textbf{ECR}: Event Completion Ratio
    \item \textbf{ECS}: Entity Consistency Score
    \item \textbf{ACS}: Action Coordination Score  
    \item \textbf{SRS}: Spatial Relation Score  
    \item \textbf{MFS}: Manipulation Feasibility Score  
    \item \textbf{VRS}: Visual Reasoning Score  
    \item \textbf{TS}: Total Score
\end{itemize}

\vspace{1mm}
\noindent\textbf{Robot-level Metrics (per-robot embodiment evaluation):}
\begin{itemize}[leftmargin=10pt, itemsep=1pt]
    \item \textbf{PSS}: Physical Semantic Score
    \item \textbf{TAC}: Task-Adherence Consistency  
    \item \textbf{RSS}: Robot–Subject Stability  
    \item \textbf{MSS}: Motion Smoothness Score
    \item \textbf{MAS}: Motion Amplitude Score
    \item \textbf{TC}: Task Completion  
    \item \textbf{VQ}: Visual Quality  
    \item \textbf{TS}: Total Score
\end{itemize}

The following tables report the complete results for all models evaluated using both GPT and Qwen. 
As shown in Table~\ref{tab:common_manipulation_gpt}, Table~\ref{tab:long-horizon_planning_gpt}, 
Table~\ref{tab:multi-entity_collaboration_gpt}, Table~\ref{tab:spatial_relationship_gpt}, and 
Table~\ref{tab:visual_reasoning_gpt}, the GPT-based evaluator provides assessments across 
five tasks.

Similarly, the Qwen-based results are presented in 
Table~\ref{tab:common_manipulation_qwen}, Table~\ref{tab:long-horizon_planning_qwen}, 
Table~\ref{tab:multi-entity_collaboration_qwen}, Table~\ref{tab:spatial_relationship_qwen}, and 
Table~\ref{tab:visual_reasoning_qwen}.

For completeness, we also provide per-embodiment results. The detailed evaluation tables for 
dual-arm, humanoid, single-arm, and quadruped robots under GPT are shown in 
Table~\ref{tab:dual_arm_gpt}, Table~\ref{tab:humanoid_gpt}, 
Table~\ref{tab:single_arm_gpt}, and Table~\ref{tab:quad_gpt}. 
The corresponding Qwen-based tables are provided in 
Table~\ref{tab:dual_arm_qwen}, Table~\ref{tab:humanoid_qwen}, 
Table~\ref{tab:single_arm_qwen}, and Table~\ref{tab:quad_qwen}.

% ---- compact tables tweaks ----
\captionsetup[table]{font=small,skip=2pt}
\setlength{\textfloatsep}{6pt plus 2pt minus 2pt}
\setlength{\floatsep}{6pt plus 2pt minus 2pt}
\setlength{\intextsep}{6pt plus 2pt minus 2pt}
\renewcommand{\floatpagefraction}{.75}
\renewcommand{\textfraction}{.1}
\renewcommand{\topfraction}{.9}
\renewcommand{\bottomfraction}{.9}
\setcounter{topnumber}{5}
\setcounter{bottomnumber}{5}
\setcounter{totalnumber}{10}

% ============================================================ 001
\begin{table*}[t]
\centering

% ===== Left Table =====
\begin{minipage}{0.48\textwidth}
\centering
\scriptsize
\setlength{\tabcolsep}{1.5pt}
\renewcommand{\arraystretch}{1.07}

\caption{\textbf{Results on Common Manipulation with GPT}}
\label{tab:common_manipulation_gpt}
\begin{tabular}{lcccccc}
\toprule
\textbf{Model} & \textbf{AES} & \textbf{TCS} & \textbf{OCS} & \textbf{RCS} & \textbf{PSS} & \textbf{TS} \\
\midrule
\rowcolor{RowBlue}
\multicolumn{7}{l}{\textit{Open-source}} \\
Wan2.2\_A14B & 0.479 & 0.410 & 0.635 & 0.765 & 0.520 & 0.381 \\
HunyuanVideo 1.5 & 0.505 & 0.480 & 0.575 & 0.695 & 0.490 & 0.442 \\
LongCat-Video & 0.469 & 0.408 & 0.591 & 0.739 & 0.510 & 0.371 \\
Wan2.1\_14B & 0.446 & 0.375 & 0.552 & 0.692 & 0.510 & 0.344 \\
LTX-2 & 0.340 & 0.330 & 0.515 & 0.620 & 0.380 & 0.284 \\
Wan2.2\_5B & 0.416 & 0.395 & 0.540 & 0.670 & 0.495 & 0.331 \\
Skyreels & 0.348 & 0.230 & 0.545 & 0.740 & 0.465 & 0.202 \\
LTX-Video & 0.414 & 0.307 & 0.552 & 0.718 & 0.484 & 0.302 \\
FramePack & 0.346 & 0.188 & 0.637 & 0.739 & 0.556 & 0.205 \\
HunyuanVideo & 0.307 & 0.190 & 0.660 & 0.755 & 0.545 & 0.177 \\
CogVideoX-5B & 0.245 & 0.140 & 0.490 & 0.635 & 0.335 & 0.115 \\
\rowcolor{RowBlue}
\multicolumn{7}{l}{\textit{Closed-source}} \\
Wan 2.6 & 0.581 & 0.596 & 0.637 & 0.750 & 0.668 & 0.545 \\
Seedance 1.5 pro & 0.654 & 0.642 & 0.591 & 0.750 & 0.556 & 0.576 \\
Wan 2.5 & 0.565 & 0.600 & 0.635 & 0.770 & 0.605 & 0.527 \\
Hailuo v2 & 0.576 & 0.625 & 0.625 & 0.745 & 0.595 & 0.559 \\
Veo 3 & 0.572 & 0.602 & 0.607 & 0.729 & 0.540 & 0.520 \\
Seedance 1.0 & 0.591 & 0.590 & 0.640 & 0.730 & 0.620 & 0.542 \\
Kling 2.6 pro & 0.561 & 0.565 & 0.610 & 0.760 & 0.590 & 0.528 \\
Sora v2 Pro & 0.354 & 0.229 & 0.637 & 0.719 & 0.561 & 0.207 \\
Sora v1 & 0.280 & 0.170 & 0.445 & 0.605 & 0.360 & 0.151 \\
\rowcolor{RowBlue}
\multicolumn{7}{l}{\textit{Robotics-specific}} \\
Cosmos 2.5 & 0.495 & 0.370 & 0.585 & 0.705 & 0.505 & 0.358 \\
DreamGen(gr1) & 0.420 & 0.341 & 0.566 & 0.683 & 0.469 & 0.311 \\
DreamGen(droid) & 0.420 & 0.369 & 0.604 & 0.640 & 0.421 & 0.358 \\
Vidar & 0.180 & 0.090 & 0.430 & 0.470 & 0.310 & 0.073 \\
UnifoLM-WMA-0 & 0.105 & 0.045 & 0.400 & 0.300 & 0.220 & 0.036 \\
\midrule
\textbf{Mean} & \textbf{0.427} & \textbf{0.371} & \textbf{0.572} & \textbf{0.685} & \textbf{0.492} & \textbf{0.338} \\
\bottomrule
\end{tabular}
\end{minipage}
\hfill
% ===== Right Table =====
\begin{minipage}{0.48\textwidth}
\centering
\scriptsize
\setlength{\tabcolsep}{1.5pt}
\renewcommand{\arraystretch}{1.07}

\caption{\textbf{Results on Long-Horizon Planning with GPT}}
\label{tab:long-horizon_planning_gpt}
\begin{tabular}{lcccccc}
\toprule
\textbf{Model} & \textbf{AES} & \textbf{ECR} & \textbf{OCS} & \textbf{RCS} & \textbf{PSS} & \textbf{TS} \\
\midrule
\rowcolor{RowBlue}
\multicolumn{7}{l}{\textit{Open-source}} \\
Wan2.2\_A14B & 0.569 & 0.626 & 0.715 & 0.722 & 0.590 & 0.500 \\
HunyuanVideo 1.5 & 0.537 & 0.527 & 0.606 & 0.725 & 0.537 & 0.437 \\
LongCat-Video & 0.507 & 0.485 & 0.598 & 0.742 & 0.598 & 0.384 \\
Wan2.1\_14B & 0.482 & 0.404 & 0.589 & 0.687 & 0.553 & 0.335 \\
LTX-2 & 0.485 & 0.446 & 0.566 & 0.676 & 0.529 & 0.386 \\
Wan2.2\_5B & 0.444 & 0.517 & 0.601 & 0.666 & 0.425 & 0.317 \\
Skyreels & 0.394 & 0.311 & 0.673 & 0.740 & 0.500 & 0.253 \\
LTX-Video & 0.433 & 0.347 & 0.566 & 0.633 & 0.458 & 0.279 \\
FramePack & 0.301 & 0.145 & 0.655 & 0.732 & 0.560 & 0.168 \\
HunyuanVideo & 0.241 & 0.101 & 0.651 & 0.714 & 0.553 & 0.147 \\
CogVideoX-5B & 0.301 & 0.223 & 0.543 & 0.629 & 0.448 & 0.212 \\
\rowcolor{RowBlue}
\multicolumn{7}{l}{\textit{Closed-source}} \\
Wan 2.6 & 0.640 & 0.545 & 0.701 & 0.743 & 0.634 & 0.514 \\
Seedance 1.5 pro & 0.638 & 0.710 & 0.677 & 0.763 & 0.625 & 0.569 \\
Wan 2.5 & 0.603 & 0.519 & 0.743 & 0.737 & 0.615 & 0.495 \\
Hailuo v2 & 0.600 & 0.677 & 0.725 & 0.706 & 0.637 & 0.544 \\
Veo 3 & 0.608 & 0.681 & 0.709 & 0.729 & 0.641 & 0.530 \\
Seedance 1.0 & 0.606 & 0.603 & 0.712 & 0.727 & 0.628 & 0.454 \\
Kling 2.6 pro & 0.618 & 0.685 & 0.710 & 0.750 & 0.697 & 0.530 \\
Sora v2 Pro & 0.422 & 0.296 & 0.646 & 0.715 & 0.603 & 0.255 \\
Sora v1 & 0.250 & 0.133 & 0.629 & 0.689 & 0.543 & 0.166 \\
\rowcolor{RowBlue}
\multicolumn{7}{l}{\textit{Robotics-specific}} \\
Cosmos 2.5 & 0.568 & 0.556 & 0.706 & 0.656 & 0.593 & 0.495 \\
DreamGen(gr1) & 0.444 & 0.284 & 0.750 & 0.731 & 0.611 & 0.333 \\
DreamGen(droid) & 0.491 & 0.505 & 0.655 & 0.637 & 0.517 & 0.316 \\
Vidar & 0.108 & -0.023 & 0.550 & 0.491 & 0.375 & 0.054 \\
UnifoLM-WMA-0 & 0.037 & 0.026 & 0.398 & 0.175 & 0.120 & 0.061 \\
\midrule
\textbf{Mean} & \textbf{0.453} & \textbf{0.413} & \textbf{0.643} & \textbf{0.677} & \textbf{0.544} & \textbf{0.349} \\
\bottomrule
\end{tabular}
\end{minipage}

\end{table*}
\vspace{-2mm}

% ============================================================ 002
\begin{table*}[t]
\centering

% ===== Left Table =====
\begin{minipage}{0.48\textwidth}
\centering
\scriptsize
\setlength{\tabcolsep}{1.5pt}
\renewcommand{\arraystretch}{1.07}

\caption{\textbf{Results on Multi-Entity Collaboration with GPT}}
\label{tab:multi-entity_collaboration_gpt}
\begin{tabular}{lcccccc}
\toprule
\textbf{Model} & \textbf{ACS} & \textbf{TCS} & \textbf{ECS} & \textbf{OCS} & \textbf{PSS} & \textbf{TS} \\
\midrule
\rowcolor{RowBlue}
\multicolumn{7}{l}{\textit{Open-source}} \\
Wan2.2\_A14B & 0.351 & 0.441 & 0.712 & 0.632 & 0.654 & 0.373 \\
HunyuanVideo 1.5 & 0.301 & 0.375 & 0.687 & 0.625 & 0.585 & 0.311 \\
LongCat-Video & 0.190 & 0.244 & 0.696 & 0.642 & 0.654 & 0.220 \\
Wan2.1\_14B & 0.261 & 0.318 & 0.693 & 0.545 & 0.551 & 0.282 \\
LTX-2 & 0.243 & 0.289 & 0.651 & 0.572 & 0.578 & 0.233 \\
Wan2.2\_5B & 0.125 & 0.181 & 0.687 & 0.568 & 0.537 & 0.141 \\
Skyreels & 0.189 & 0.195 & 0.695 & 0.664 & 0.621 & 0.203 \\
LTX-Video & 0.207 & 0.255 & 0.686 & 0.574 & 0.563 & 0.209 \\
FramePack & 0.186 & 0.186 & 0.709 & 0.598 & 0.529 & 0.173 \\
HunyuanVideo & 0.100 & 0.100 & 0.672 & 0.683 & 0.650 & 0.107 \\
CogVideoX-5B & 0.128 & 0.113 & 0.630 & 0.482 & 0.440 & 0.098 \\
\rowcolor{RowBlue}
\multicolumn{7}{l}{\textit{Closed-source}} \\
Wan 2.6 & 0.443 & 0.541 & 0.738 & 0.708 & 0.654 & 0.478 \\
Seedance 1.5 pro & 0.456 & 0.591 & 0.743 & 0.689 & 0.689 & 0.483 \\
Wan 2.5 & 0.392 & 0.453 & 0.750 & 0.633 & 0.662 & 0.401 \\
Hailuo v2 & 0.378 & 0.422 & 0.744 & 0.678 & 0.619 & 0.385 \\
Veo 3 & 0.450 & 0.432 & 0.731 & 0.737 & 0.621 & 0.430 \\
Seedance 1.0 & 0.422 & 0.488 & 0.750 & 0.702 & 0.696 & 0.447 \\
Kling 2.6 pro & 0.357 & 0.392 & 0.738 & 0.698 & 0.676 & 0.364 \\
Sora v2 Pro & 0.155 & 0.191 & 0.733 & 0.558 & 0.591 & 0.186 \\
Sora v1 & 0.128 & 0.107 & 0.595 & 0.494 & 0.523 & 0.111 \\
\rowcolor{RowBlue}
\multicolumn{7}{l}{\textit{Robotics-specific}} \\
Cosmos 2.5 & 0.203 & 0.244 & 0.727 & 0.659 & 0.670 & 0.201 \\
DreamGen(gr1) & 0.262 & 0.347 & 0.707 & 0.664 & 0.628 & 0.296 \\
DreamGen(droid) & 0.211 & 0.260 & 0.657 & 0.548 & 0.548 & 0.214 \\
Vidar & 0.056 & 0.062 & 0.454 & 0.431 & 0.329 & 0.049 \\
UnifoLM-WMA-0 & 0.017 & 0.034 & 0.392 & 0.301 & 0.295 & 0.018 \\
\midrule
\textbf{Mean} & \textbf{0.248} & \textbf{0.290} & \textbf{0.679} & \textbf{0.603} & \textbf{0.583} & \textbf{0.256} \\
\bottomrule
\end{tabular}
\end{minipage}
\hfill
% ===== Right Table =====
\begin{minipage}{0.48\textwidth}
\centering
\scriptsize
\setlength{\tabcolsep}{1.5pt}
\renewcommand{\arraystretch}{1.07}

\caption{\textbf{Results on Spatial Relationship with GPT}}
\label{tab:spatial_relationship_gpt}
\begin{tabular}{lcccccc}
\toprule
\textbf{Model} & \textbf{SRS} & \textbf{MFS} & \textbf{OCS} & \textbf{RCS} & \textbf{PSS} & \textbf{TS} \\
\midrule
\rowcolor{RowBlue}
\multicolumn{7}{l}{\textit{Open-source}} \\
Wan2.2\_A14B & 0.604 & 0.596 & 0.709 & 0.758 & 0.669 & 0.454 \\
HunyuanVideo 1.5 & 0.370 & 0.344 & 0.629 & 0.750 & 0.534 & 0.315 \\
LongCat-Video & 0.390 & 0.375 & 0.679 & 0.734 & 0.648 & 0.310 \\
Wan2.1\_14B & 0.316 & 0.338 & 0.669 & 0.713 & 0.610 & 0.267 \\
LTX-2 & 0.401 & 0.383 & 0.616 & 0.687 & 0.562 & 0.304 \\
Wan2.2\_5B & 0.441 & 0.389 & 0.602 & 0.705 & 0.485 & 0.312 \\
Skyreels & 0.388 & 0.416 & 0.601 & 0.685 & 0.574 & 0.276 \\
LTX-Video & 0.224 & 0.215 & 0.612 & 0.681 & 0.491 & 0.176 \\
FramePack & 0.364 & 0.321 & 0.628 & 0.742 & 0.557 & 0.257 \\
HunyuanVideo & 0.163 & 0.192 & 0.605 & 0.740 & 0.586 & 0.179 \\
CogVideoX-5B & 0.193 & 0.241 & 0.451 & 0.653 & 0.370 & 0.111 \\
\rowcolor{RowBlue}
\multicolumn{7}{l}{\textit{Closed-source}} \\
Wan 2.6 & 0.787 & 0.704 & 0.719 & 0.734 & 0.734 & 0.655 \\
Seedance 1.5 pro & 0.675 & 0.608 & 0.700 & 0.716 & 0.666 & 0.494 \\
Wan 2.5 & 0.750 & 0.598 & 0.757 & 0.750 & 0.674 & 0.576 \\
Hailuo v2 & 0.764 & 0.654 & 0.742 & 0.750 & 0.720 & 0.636 \\
Veo 3 & 0.601 & 0.553 & 0.726 & 0.767 & 0.684 & 0.508 \\
Seedance 1.0 & 0.484 & 0.445 & 0.710 & 0.742 & 0.679 & 0.425 \\
Kling 2.6 pro & 0.757 & 0.681 & 0.727 & 0.757 & 0.696 & 0.598 \\
Sora v2 Pro & 0.392 & 0.392 & 0.654 & 0.714 & 0.571 & 0.267 \\
Sora v1 & 0.351 & 0.305 & 0.546 & 0.583 & 0.481 & 0.223 \\
\rowcolor{RowBlue}
\multicolumn{7}{l}{\textit{Robotics-specific}} \\
Cosmos 2.5 & 0.419 & 0.395 & 0.661 & 0.693 & 0.620 & 0.338 \\
DreamGen(gr1) & 0.467 & 0.411 & 0.677 & 0.725 & 0.677 & 0.371 \\
DreamGen(droid) & 0.400 & 0.433 & 0.625 & 0.691 & 0.591 & 0.348 \\
Vidar & 0.163 & 0.250 & 0.517 & 0.500 & 0.405 & 0.105 \\
UnifoLM-WMA-0 & 0.065 & 0.141 & 0.445 & 0.532 & 0.315 & 0.040 \\
\midrule
\textbf{Mean} & \textbf{0.437} & \textbf{0.415} & \textbf{0.640} & \textbf{0.700} & \textbf{0.584} & \textbf{0.342} \\
\bottomrule
\end{tabular}
\end{minipage}

\end{table*}
\vspace{-2mm}

% ============================================================ 003
\begin{table*}[t]
\centering

% ===== Left Table =====
\begin{minipage}{0.48\textwidth}
\centering
\scriptsize
\setlength{\tabcolsep}{1.5pt}
\renewcommand{\arraystretch}{1.07}

\caption{\textbf{Results on Visual Reasoning with GPT}}
\label{tab:visual_reasoning_gpt}
\begin{tabular}{lcccccc}
\toprule
\textbf{Model} & \textbf{AES} & \textbf{VRS} & \textbf{OCS} & \textbf{RCS} & \textbf{PSS} & \textbf{TS} \\
\midrule
\rowcolor{RowBlue}
\multicolumn{7}{l}{\textit{Open-source}} \\
Wan2.2\_A14B & 0.424 & 0.401 & 0.651 & 0.709 & 0.552 & 0.330 \\
HunyuanVideo 1.5 & 0.453 & 0.456 & 0.604 & 0.709 & 0.447 & 0.364 \\
LongCat-Video & 0.271 & 0.211 & 0.559 & 0.722 & 0.516 & 0.186 \\
Wan2.1\_14B & 0.250 & 0.263 & 0.565 & 0.646 & 0.429 & 0.204 \\
LTX-2 & 0.267 & 0.224 & 0.500 & 0.610 & 0.360 & 0.163 \\
Wan2.2\_5B & 0.283 & 0.304 & 0.583 & 0.700 & 0.438 & 0.233 \\
Skyreels & 0.290 & 0.267 & 0.587 & 0.668 & 0.447 & 0.233 \\
LTX-Video & 0.283 & 0.287 & 0.644 & 0.688 & 0.516 & 0.241 \\
FramePack & 0.243 & 0.203 & 0.570 & 0.743 & 0.397 & 0.169 \\
HunyuanVideo & 0.096 & 0.058 & 0.647 & 0.744 & 0.522 & 0.035 \\
CogVideoX-5B & 0.136 & 0.120 & 0.428 & 0.577 & 0.261 & 0.079 \\
\rowcolor{RowBlue}
\multicolumn{7}{l}{\textit{Closed-source}} \\
Wan 2.6 & 0.544 & 0.622 & 0.700 & 0.733 & 0.605 & 0.530 \\
Seedance 1.5 pro & 0.494 & 0.570 & 0.635 & 0.701 & 0.570 & 0.470 \\
Wan 2.5 & 0.482 & 0.488 & 0.693 & 0.732 & 0.539 & 0.437 \\
Hailuo v2 & 0.511 & 0.541 & 0.666 & 0.727 & 0.566 & 0.473 \\
Veo 3 & 0.511 & 0.610 & 0.633 & 0.711 & 0.577 & 0.504 \\
Seedance 1.0 & 0.505 & 0.505 & 0.705 & 0.733 & 0.644 & 0.441 \\
Kling 2.6 pro & 0.477 & 0.410 & 0.627 & 0.733 & 0.555 & 0.357 \\
Sora v2 Pro & 0.193 & 0.159 & 0.642 & 0.738 & 0.556 & 0.115 \\
Sora v1 & 0.219 & 0.158 & 0.475 & 0.646 & 0.250 & 0.139 \\
\rowcolor{RowBlue}
\multicolumn{7}{l}{\textit{Robotics-specific}} \\
Cosmos 2.5 & 0.482 & 0.493 & 0.664 & 0.744 & 0.590 & 0.399 \\
DreamGen(gr1) & 0.317 & 0.262 & 0.652 & 0.743 & 0.585 & 0.215 \\
DreamGen(droid) & 0.412 & 0.401 & 0.616 & 0.738 & 0.529 & 0.338 \\
Vidar & 0.090 & 0.050 & 0.556 & 0.659 & 0.431 & 0.050 \\
UnifoLM-WMA-0 & 0.017 & -0.056 & 0.062 & 0.306 & 0.176 & 0.000 \\
\midrule
\textbf{Mean} & \textbf{0.330} & \textbf{0.320} & \textbf{0.587} & \textbf{0.686} & \textbf{0.482} & \textbf{0.268} \\
\bottomrule
\end{tabular}
\end{minipage}
\hfill
% ===== Right Table =====
\begin{minipage}{0.48\textwidth}
\centering
\scriptsize
\setlength{\tabcolsep}{1pt}
\renewcommand{\arraystretch}{1.07}

\caption{\textbf{Results on Dual Arm with GPT}}
\label{tab:dual_arm_gpt}
\begin{tabular}{lcccccccc}
\toprule
\textbf{Model} & \textbf{PSS} & \textbf{TAC} & \textbf{RSS} & \textbf{MS} & \textbf{MA} & \textbf{TC} & \textbf{VQ} & \textbf{TS} \\
\midrule
\rowcolor{RowBlue}
\multicolumn{9}{l}{\textit{Open-source}} \\
Wan2.2\_A14B & 0.638 & 0.570 & 0.764 & 0.915 & 0.204 & 0.604 & 0.561 & 0.582 \\
HunyuanVideo 1.5 & 0.612 & 0.622 & 0.649 & 0.951 & 0.370 & 0.618 & 0.434 & 0.526 \\
LongCat-Video & 0.620 & 0.540 & 0.771 & 0.937 & 0.244 & 0.580 & 0.572 & 0.576 \\
Wan2.1\_14B & 0.600 & 0.540 & 0.650 & 0.850 & 0.261 & 0.570 & 0.424 & 0.497 \\
LTX-2 & 0.488 & 0.415 & 0.637 & 0.848 & 0.378 & 0.451 & 0.396 & 0.423 \\
Wan2.2\_5B & 0.575 & 0.498 & 0.606 & 0.940 & 0.269 & 0.536 & 0.360 & 0.448 \\
Skyreels & 0.598 & 0.498 & 0.658 & 0.884 & 0.252 & 0.548 & 0.406 & 0.477 \\
LTX-Video & 0.530 & 0.442 & 0.698 & 0.812 & 0.143 & 0.486 & 0.425 & 0.455 \\
FramePack & 0.550 & 0.395 & 0.712 & 0.885 & 0.103 & 0.472 & 0.457 & 0.464 \\
HunyuanVideo & 0.510 & 0.280 & 0.794 & 0.959 & 0.107 & 0.395 & 0.564 & 0.479 \\
CogVideoX-5B & 0.480 & 0.358 & 0.638 & 0.752 & 0.143 & 0.419 & 0.352 & 0.385 \\
\rowcolor{RowBlue}
\multicolumn{9}{l}{\textit{Closed-source}} \\
Wan 2.6 & 0.655 & 0.708 & 0.819 & 0.984 & 0.333 & 0.681 & 0.680 & 0.680 \\
Seedance 1.5 pro & 0.668 & 0.800 & 0.721 & 0.960 & 0.399 & 0.734 & 0.547 & 0.640 \\
Wan 2.5 & 0.670 & 0.740 & 0.758 & 0.970 & 0.347 & 0.705 & 0.563 & 0.633 \\
Hailuo v2 & 0.658 & 0.720 & 0.751 & 0.983 & 0.312 & 0.689 & 0.534 & 0.611 \\
Veo 3 & 0.665 & 0.682 & 0.711 & 0.973 & 0.262 & 0.674 & 0.546 & 0.610 \\
Seedance 1.0 & 0.668 & 0.648 & 0.801 & 0.972 & 0.294 & 0.658 & 0.623 & 0.640 \\
Kling 2.6 pro & 0.672 & 0.615 & 0.770 & 0.965 & 0.210 & 0.644 & 0.567 & 0.605 \\
Sora v2 Pro & 0.565 & 0.364 & 0.776 & 0.950 & 0.272 & 0.465 & 0.560 & 0.512 \\
Sora v1 & 0.422 & 0.312 & 0.531 & 0.880 & 0.210 & 0.368 & 0.279 & 0.323 \\
\rowcolor{RowBlue}
\multicolumn{9}{l}{\textit{Robotics-specific}} \\
Cosmos 2.5 & 0.665 & 0.575 & 0.746 & 0.930 & 0.127 & 0.620 & 0.500 & 0.560 \\
DreamGen(gr1) & 0.630 & 0.492 & 0.779 & 0.939 & 0.123 & 0.561 & 0.503 & 0.532 \\
DreamGen(droid) & 0.610 & 0.542 & 0.668 & 0.863 & 0.201 & 0.576 & 0.375 & 0.475 \\
Vidar & 0.450 & 0.238 & 0.781 & 0.933 & 0.025 & 0.344 & 0.475 & 0.409 \\
UnifoLM-WMA-0 & 0.348 & 0.252 & 0.348 & 0.497 & 0.120 & 0.300 & 0.089 & 0.194 \\
\midrule
\textbf{Mean} & \textbf{0.581} & \textbf{0.513} & \textbf{0.701} & \textbf{0.901} & \textbf{0.228} & \textbf{0.547} & \textbf{0.471} & \textbf{0.509} \\
\bottomrule
\end{tabular}
\end{minipage}

\end{table*}
\vspace{-2mm}

% ============================================================ 004
\begin{table*}[t]
\centering

% ===== Left Table =====
\begin{minipage}{0.48\textwidth}
\centering
\scriptsize
\setlength{\tabcolsep}{0.5pt}
\renewcommand{\arraystretch}{1.07}

\caption{\textbf{Results on Humanoid with GPT}}
\label{tab:humanoid_gpt}
\begin{tabular}{lcccccccc}
\toprule
\textbf{Model} & \textbf{PSS} & \textbf{TAC} & \textbf{RSS} & \textbf{MS} & \textbf{MA} & \textbf{TC} & \textbf{VQ} & \textbf{TS} \\
\midrule
\rowcolor{RowBlue}
\multicolumn{9}{l}{\textit{Open-source}} \\
Wan2.2\_A14B & 0.678 & 0.748 & 0.787 & 0.966 & 0.105 & 0.712 & 0.584 & 0.647 \\
HunyuanVideo 1.5 & 0.660 & 0.800 & 0.703 & 0.922 & 0.206 & 0.730 & 0.460 & 0.595 \\
LongCat-Video & 0.652 & 0.668 & 0.801 & 0.969 & 0.089 & 0.660 & 0.583 & 0.621 \\
Wan2.1\_14B & 0.635 & 0.660 & 0.765 & 0.938 & 0.167 & 0.648 & 0.550 & 0.599 \\
LTX-2 & 0.622 & 0.608 & 0.736 & 0.880 & 0.231 & 0.615 & 0.495 & 0.554 \\
Wan2.2\_5B & 0.668 & 0.695 & 0.755 & 0.963 & 0.152 & 0.681 & 0.533 & 0.607 \\
Skyreels & 0.610 & 0.598 & 0.725 & 0.808 & 0.086 & 0.604 & 0.414 & 0.509 \\
LTX-Video & 0.602 & 0.630 & 0.671 & 0.628 & 0.050 & 0.616 & 0.312 & 0.463 \\
FramePack & 0.598 & 0.572 & 0.776 & 0.864 & 0.069 & 0.585 & 0.511 & 0.548 \\
HunyuanVideo & 0.595 & 0.465 & 0.763 & 0.933 & 0.096 & 0.530 & 0.517 & 0.523 \\
CogVideoX-5B & 0.570 & 0.565 & 0.698 & 0.647 & 0.127 & 0.568 & 0.424 & 0.496 \\
\rowcolor{RowBlue}
\multicolumn{9}{l}{\textit{Closed-source}} \\
Wan 2.6 & 0.665 & 0.818 & 0.794 & 0.980 & 0.132 & 0.741 & 0.593 & 0.667 \\
Seedance 1.5 pro & 0.682 & 0.838 & 0.793 & 0.955 & 0.158 & 0.760 & 0.623 & 0.691 \\
Wan 2.5 & 0.698 & 0.782 & 0.779 & 0.981 & 0.122 & 0.740 & 0.568 & 0.653 \\
Hailuo v2 & 0.682 & 0.815 & 0.741 & 0.970 & 0.133 & 0.749 & 0.521 & 0.635 \\
Veo 3 & 0.655 & 0.785 & 0.776 & 0.968 & 0.132 & 0.720 & 0.554 & 0.637 \\
Seedance 1.0 & 0.675 & 0.752 & 0.833 & 0.964 & 0.166 & 0.714 & 0.658 & 0.686 \\
Kling 2.6 pro & 0.672 & 0.762 & 0.750 & 0.959 & 0.113 & 0.718 & 0.508 & 0.613 \\
Sora v2 Pro & 0.638 & 0.565 & 0.774 & 0.936 & 0.085 & 0.602 & 0.520 & 0.561 \\
Sora v1 & 0.542 & 0.450 & 0.597 & 0.900 & 0.216 & 0.496 & 0.342 & 0.419 \\
\rowcolor{RowBlue}
\multicolumn{9}{l}{\textit{Robotics-specific}} \\
Cosmos 2.5 & 0.650 & 0.720 & 0.797 & 0.925 & 0.071 & 0.685 & 0.566 & 0.625 \\
DreamGen(gr1) & 0.652 & 0.595 & 0.781 & 0.885 & 0.079 & 0.624 & 0.526 & 0.575 \\
DreamGen(droid) & 0.620 & 0.700 & 0.704 & 0.843 & 0.137 & 0.660 & 0.453 & 0.556 \\
Vidar & 0.445 & 0.298 & 0.694 & 0.855 & 0.025 & 0.371 & 0.343 & 0.357 \\
UnifoLM-WMA-0 & 0.270 & 0.282 & 0.386 & 0.512 & 0.112 & 0.276 & 0.125 & 0.200 \\
\midrule
\textbf{Mean} & \textbf{0.617} & \textbf{0.646} & \textbf{0.735} & \textbf{0.886} & \textbf{0.122} & \textbf{0.632} & \textbf{0.491} & \textbf{0.561} \\
\bottomrule
\end{tabular}
\end{minipage}
\hfill
% ===== Right Table =====
\begin{minipage}{0.48\textwidth}
\centering
\scriptsize
\setlength{\tabcolsep}{0.5pt}
\renewcommand{\arraystretch}{1.07}

\caption{\textbf{Results on Single Arm with GPT}}
\label{tab:single_arm_gpt}
\begin{tabular}{lcccccccc}
\toprule
\textbf{Model} & \textbf{PSS} & \textbf{TAC} & \textbf{RSS} & \textbf{MS} & \textbf{MA} & \textbf{TC} & \textbf{VQ} & \textbf{TS} \\
\midrule
\rowcolor{RowBlue}
\multicolumn{9}{l}{\textit{Open-source}} \\
Wan2.2\_A14B & 0.638 & 0.582 & 0.783 & 0.942 & 0.263 & 0.610 & 0.607 & 0.608 \\
HunyuanVideo 1.5 & 0.510 & 0.622 & 0.651 & 0.949 & 0.355 & 0.566 & 0.460 & 0.513 \\
LongCat-Video & 0.562 & 0.530 & 0.807 & 0.945 & 0.298 & 0.546 & 0.625 & 0.585 \\
Wan2.1\_14B & 0.542 & 0.472 & 0.677 & 0.849 & 0.282 & 0.507 & 0.422 & 0.464 \\
LTX-2 & 0.490 & 0.410 & 0.681 & 0.919 & 0.464 & 0.450 & 0.456 & 0.453 \\
Wan2.2\_5B & 0.518 & 0.480 & 0.619 & 0.943 & 0.313 & 0.499 & 0.372 & 0.435 \\
Skyreels & 0.525 & 0.495 & 0.712 & 0.911 & 0.286 & 0.510 & 0.504 & 0.507 \\
LTX-Video & 0.492 & 0.408 & 0.686 & 0.804 & 0.158 & 0.450 & 0.431 & 0.440 \\
FramePack & 0.445 & 0.318 & 0.760 & 0.888 & 0.104 & 0.381 & 0.498 & 0.439 \\
HunyuanVideo & 0.445 & 0.265 & 0.809 & 0.963 & 0.118 & 0.355 & 0.552 & 0.453 \\
CogVideoX-5B & 0.405 & 0.335 & 0.582 & 0.815 & 0.256 & 0.370 & 0.307 & 0.338 \\
\rowcolor{RowBlue}
\multicolumn{9}{l}{\textit{Closed-source}} \\
Wan 2.6 & 0.652 & 0.710 & 0.796 & 0.983 & 0.392 & 0.681 & 0.651 & 0.666 \\
Seedance 1.5 pro & 0.635 & 0.832 & 0.752 & 0.960 & 0.419 & 0.734 & 0.561 & 0.647 \\
Wan 2.5 & 0.668 & 0.802 & 0.787 & 0.969 & 0.412 & 0.735 & 0.624 & 0.679 \\
Hailuo v2 & 0.665 & 0.752 & 0.680 & 0.989 & 0.396 & 0.709 & 0.479 & 0.594 \\
Veo 3 & 0.642 & 0.755 & 0.750 & 0.977 & 0.362 & 0.699 & 0.568 & 0.633 \\
Seedance 1.0 & 0.658 & 0.655 & 0.769 & 0.967 & 0.306 & 0.656 & 0.589 & 0.622 \\
Kling 2.6 pro & 0.622 & 0.640 & 0.714 & 0.958 & 0.305 & 0.631 & 0.508 & 0.569 \\
Sora v2 Pro & 0.490 & 0.310 & 0.784 & 0.962 & 0.324 & 0.400 & 0.552 & 0.476 \\
Sora v1 & 0.350 & 0.320 & 0.532 & 0.884 & 0.247 & 0.335 & 0.293 & 0.314 \\
\rowcolor{RowBlue}
\multicolumn{9}{l}{\textit{Robotics-specific}} \\
Cosmos 2.5 & 0.632 & 0.592 & 0.737 & 0.888 & 0.206 & 0.612 & 0.475 & 0.543 \\
DreamGen(gr1) & 0.618 & 0.620 & 0.716 & 0.932 & 0.263 & 0.619 & 0.508 & 0.563 \\
DreamGen(droid) & 0.568 & 0.568 & 0.691 & 0.930 & 0.214 & 0.568 & 0.430 & 0.499 \\
Vidar & 0.415 & 0.272 & 0.726 & 0.929 & 0.068 & 0.344 & 0.420 & 0.382 \\
UnifoLM-WMA-0 & 0.428 & 0.315 & 0.437 & 0.709 & 0.389 & 0.371 & 0.164 & 0.267 \\
\midrule
\textbf{Mean} & \textbf{0.544} & \textbf{0.522} & \textbf{0.705} & \textbf{0.918} & \textbf{0.288} & \textbf{0.533} & \textbf{0.482} & \textbf{0.507} \\
\bottomrule
\end{tabular}
\end{minipage}

\end{table*}
\vspace{-2mm}

% ============================================================ 005
\begin{table*}[t]
\centering

% ===== Left Table =====
\begin{minipage}{0.44\textwidth}
\centering
\scriptsize
\setlength{\tabcolsep}{1pt}
\renewcommand{\arraystretch}{1.07}

\caption{\textbf{Results on Quadruped Robot with GPT}}
\label{tab:quad_gpt}
\begin{tabular}{lcccccccc}
\toprule
\textbf{Model} & \textbf{PSS} & \textbf{TAC} & \textbf{RSS} & \textbf{MS} & \textbf{MA} & \textbf{TC} & \textbf{VQ} & \textbf{TS} \\
\midrule
\rowcolor{RowBlue}
\multicolumn{9}{l}{\textit{Open-source}} \\
Wan2.2\_A14B & 0.760 & 0.712 & 0.800 & 0.888 & 0.196 & 0.736 & 0.643 & 0.689 \\
HunyuanVideo 1.5 & 0.715 & 0.738 & 0.713 & 0.940 & 0.403 & 0.726 & 0.542 & 0.634 \\
LongCat-Video & 0.742 & 0.628 & 0.827 & 0.923 & 0.179 & 0.685 & 0.676 & 0.680 \\
Wan2.1\_14B & 0.722 & 0.670 & 0.706 & 0.850 & 0.303 & 0.696 & 0.495 & 0.595 \\
LTX-2 & 0.715 & 0.670 & 0.758 & 0.845 & 0.287 & 0.692 & 0.552 & 0.622 \\
Wan2.2\_5B & 0.712 & 0.678 & 0.688 & 0.921 & 0.273 & 0.695 & 0.486 & 0.590 \\
Skyreels & 0.732 & 0.605 & 0.722 & 0.853 & 0.163 & 0.669 & 0.503 & 0.586 \\
LTX-Video & 0.698 & 0.678 & 0.668 & 0.676 & 0.122 & 0.688 & 0.364 & 0.526 \\
FramePack & 0.720 & 0.575 & 0.827 & 0.876 & 0.069 & 0.648 & 0.605 & 0.626 \\
HunyuanVideo & 0.730 & 0.602 & 0.788 & 0.953 & 0.127 & 0.666 & 0.584 & 0.625 \\
CogVideoX-5B & 0.655 & 0.560 & 0.624 & 0.618 & 0.220 & 0.608 & 0.322 & 0.464 \\
\rowcolor{RowBlue}
\multicolumn{9}{l}{\textit{Closed-source}} \\
Wan 2.6 & 0.755 & 0.792 & 0.813 & 0.970 & 0.316 & 0.774 & 0.672 & 0.723 \\
Seedance 1.5 pro & 0.748 & 0.820 & 0.746 & 0.884 & 0.407 & 0.784 & 0.577 & 0.680 \\
Wan 2.5 & 0.785 & 0.792 & 0.809 & 0.948 & 0.322 & 0.789 & 0.664 & 0.726 \\
Hailuo v2 & 0.748 & 0.738 & 0.711 & 0.961 & 0.354 & 0.742 & 0.538 & 0.640 \\
Veo 3 & 0.745 & 0.695 & 0.798 & 0.961 & 0.214 & 0.720 & 0.658 & 0.689 \\
Seedance 1.0 & 0.768 & 0.735 & 0.791 & 0.945 & 0.334 & 0.751 & 0.645 & 0.698 \\
Kling 2.6 pro & 0.740 & 0.738 & 0.736 & 0.861 & 0.258 & 0.739 & 0.535 & 0.637 \\
Sora v2 Pro & 0.731 & 0.670 & 0.789 & 0.922 & 0.239 & 0.701 & 0.626 & 0.663 \\
Sora v1 & 0.700 & 0.672 & 0.620 & 0.863 & 0.282 & 0.686 & 0.401 & 0.543 \\
\rowcolor{RowBlue}
\multicolumn{9}{l}{\textit{Robotics-specific}} \\
Cosmos 2.5 & 0.752 & 0.622 & 0.808 & 0.892 & 0.137 & 0.688 & 0.629 & 0.658 \\
DreamGen(gr1) & 0.712 & 0.655 & 0.706 & 0.788 & 0.164 & 0.684 & 0.474 & 0.579 \\
DreamGen(droid) & 0.705 & 0.568 & 0.687 & 0.854 & 0.160 & 0.636 & 0.448 & 0.542 \\
Vidar & 0.528 & 0.472 & 0.552 & 0.749 & 0.074 & 0.500 & 0.247 & 0.373 \\
UnifoLM-WMA-0 & 0.475 & 0.390 & 0.410 & 0.497 & 0.132 & 0.432 & 0.154 & 0.293 \\
\midrule
\textbf{Mean} & \textbf{0.711} & \textbf{0.659} & \textbf{0.723} & \textbf{0.857} & \textbf{0.229} & \textbf{0.685} & \textbf{0.521} & \textbf{0.603} \\
\bottomrule
\end{tabular}
\end{minipage}
\hfill
% ===== Right Table =====
\begin{minipage}{0.48\textwidth}
\centering
\scriptsize
\setlength{\tabcolsep}{1pt}
\renewcommand{\arraystretch}{1.07}

\caption{\textbf{Results on Common Manipulation with Qwen}}
\label{tab:common_manipulation_qwen}
\begin{tabular}{lcccccc}
\toprule
\textbf{Model} & \textbf{AES} & \textbf{TCS} & \textbf{OCS} & \textbf{RCS} & \textbf{PSS} & \textbf{TS} \\
\midrule
\rowcolor{RowBlue}
\multicolumn{7}{l}{\textit{Open-source}} \\
Wan2.2\_A14B & 0.804 & 0.815 & 0.934 & 0.913 & 0.923 & 0.708 \\
LongCat-Video & 0.681 & 0.704 & 0.840 & 0.920 & 0.897 & 0.677 \\
Wan2.2\_5B & 0.620 & 0.630 & 0.940 & 0.890 & 0.860 & 0.597 \\
Wan2.1\_14B & 0.663 & 0.653 & 0.903 & 0.932 & 0.807 & 0.687 \\
Skyreels & 0.547 & 0.559 & 0.928 & 0.916 & 0.952 & 0.546 \\
LTX-Video & 0.630 & 0.666 & 0.845 & 0.904 & 0.880 & 0.450 \\
FramePack & 0.336 & 0.347 & 0.913 & 0.956 & 0.934 & 0.455 \\
CogVideoX-5B & 0.352 & 0.340 & 0.625 & 0.704 & 0.647 & 0.289 \\
\rowcolor{RowBlue}
\multicolumn{7}{l}{\textit{Closed-source}} \\
Wan 2.5 & 0.928 & 0.946 & 0.946 & 0.964 & 0.973 & 0.887 \\
Hailuo v2 & 0.916 & 0.952 & 0.940 & 0.940 & 0.976 & 0.843 \\
Veo 3 & 0.953 & 0.962 & 0.962 & 0.981 & 0.990 & 0.896 \\
Seedance 1.0 & 0.946 & 0.928 & 0.955 & 0.946 & 0.964 & 0.856 \\
\rowcolor{RowBlue}
\multicolumn{7}{l}{\textit{Robotics-specific}} \\
Cosmos 2.5 & 0.687 & 0.708 & 0.895 & 0.916 & 0.864 & 0.687 \\
DreamGen(gr1) & 0.625 & 0.656 & 0.906 & 0.937 & 0.906 & 0.507 \\
DreamGen(droid) & 0.630 & 0.690 & 0.785 & 0.845 & 0.773 & 0.465 \\
UnifoLM-WMA-0 & 0.043 & 0.043 & 0.532 & 0.478 & 0.369 & 0.028 \\
Vidar & 0.142 & 0.166 & 0.833 & 0.833 & 0.761 & 0.117 \\
\midrule
\textbf{Mean} & \textbf{0.597} & \textbf{0.611} & \textbf{0.867} & \textbf{0.858} & \textbf{0.835} & \textbf{0.559} \\
\bottomrule
\end{tabular}
\end{minipage}

\end{table*}
\vspace{-2mm}

% ============================================================ 006
\begin{table*}[t]
\centering

% ===== Left Table =====
\begin{minipage}{0.48\textwidth}
\centering
\scriptsize
\setlength{\tabcolsep}{1.5pt}
\renewcommand{\arraystretch}{1.07}

\caption{\textbf{Results on Long-Horizon Planning with Qwen}}
\label{tab:long-horizon_planning_qwen}
\begin{tabular}{lcccccc}
\toprule
\textbf{Model} & \textbf{AES} & \textbf{ECS} & \textbf{OCS} & \textbf{RCS} & \textbf{PSS} & \textbf{TS} \\
\midrule
\rowcolor{RowBlue}
\multicolumn{7}{l}{\textit{Open-source}} \\
Wan2.2\_A14B & 0.836 & 0.883 & 0.942 & 0.971 & 0.951 & 0.680 \\
LongCat-Video & 0.650 & 0.706 & 0.925 & 0.950 & 0.912 & 0.489 \\
Wan2.2\_5B & 0.520 & 0.594 & 0.947 & 0.937 & 0.895 & 0.449 \\
Wan2.1\_14B & 0.681 & 0.669 & 0.931 & 0.931 & 0.875 & 0.465 \\
Skyreels & 0.416 & 0.511 & 0.944 & 0.958 & 0.861 & 0.324 \\
LTX-Video & 0.352 & 0.414 & 0.779 & 0.808 & 0.705 & 0.357 \\
FramePack & 0.212 & 0.252 & 0.862 & 0.987 & 0.937 & 0.196 \\
CogVideoX-5B & 0.276 & 0.292 & 0.855 & 0.763 & 0.684 & 0.096 \\
\rowcolor{RowBlue}
\multicolumn{7}{l}{\textit{Closed-source}} \\
Wan 2.5 & 0.714 & 0.836 & 0.964 & 0.964 & 0.955 & 0.719 \\
Hailuo v2 & 0.808 & 0.903 & 0.950 & 0.991 & 0.908 & 0.705 \\
Veo 3 & 0.812 & 0.903 & 0.984 & 0.968 & 0.945 & 0.854 \\
Seedance 1.0 & 0.824 & 0.889 & 0.990 & 0.990 & 0.925 & 0.715 \\
\rowcolor{RowBlue}
\multicolumn{7}{l}{\textit{Robotics-specific}} \\
Cosmos 2.5 & 0.731 & 0.810 & 0.953 & 0.981 & 0.935 & 0.596 \\
DreamGen(gr1) & 0.475 & 0.587 & 0.950 & 0.937 & 0.887 & 0.353 \\
DreamGen(droid) & 0.602 & 0.717 & 0.882 & 0.882 & 0.852 & 0.301 \\
UnifoLM-WMA-0 & 0.000 & -0.041 & 0.645 & 0.500 & 0.500 & 0.000 \\
Vidar & 0.050 & 0.058 & 0.800 & 0.866 & 0.666 & 0.019 \\
\midrule
\textbf{Mean} & \textbf{0.505} & \textbf{0.565} & \textbf{0.893} & \textbf{0.886} & \textbf{0.833} & \textbf{0.417} \\
\bottomrule
\end{tabular}
\end{minipage}
\hfill
% ===== Right Table =====
\begin{minipage}{0.48\textwidth}
\centering
\scriptsize
\setlength{\tabcolsep}{1.5pt}
\renewcommand{\arraystretch}{1.07}

\caption{\textbf{Results on Multi-Entity Collaboration with Qwen}}
\label{tab:multi-entity_collaboration_qwen}
\begin{tabular}{lcccccc}
\toprule
\textbf{Model} & \textbf{ACS} & \textbf{TCS} & \textbf{ECS} & \textbf{OCS} & \textbf{PSS} & \textbf{TS} \\
\midrule
\rowcolor{RowBlue}
\multicolumn{7}{l}{\textit{Open-source}} \\
Wan2.2\_A14B & 0.941 & 0.941 & 0.985 & 0.992 & 1.000 & 0.920 \\
LongCat-Video & 0.887 & 0.912 & 0.968 & 0.962 & 0.968 & 0.814 \\
Wan2.2\_5B & 0.763 & 0.796 & 0.993 & 0.960 & 0.967 & 0.721 \\
Wan2.1\_14B & 0.750 & 0.786 & 0.975 & 0.945 & 0.957 & 0.702 \\
Skyreels & 0.743 & 0.736 & 1.000 & 1.000 & 0.993 & 0.686 \\
LTX-Video & 0.806 & 0.806 & 0.975 & 0.975 & 0.993 & 0.734 \\
FramePack & 0.710 & 0.703 & 0.980 & 0.993 & 0.947 & 0.630 \\
CogVideoX-5B & 0.522 & 0.536 & 0.933 & 0.882 & 0.889 & 0.426 \\
\rowcolor{RowBlue}
\multicolumn{7}{l}{\textit{Closed-source}} \\
Wan 2.5 & 0.896 & 0.908 & 0.987 & 0.975 & 1.000 & 0.915 \\
Hailuo v2 & 0.986 & 0.993 & 0.952 & 0.972 & 0.986 & 0.892 \\
Veo 3 & 0.914 & 0.914 & 0.993 & 0.987 & 1.000 & 0.924 \\
Seedance 1.0 & 0.960 & 0.967 & 0.973 & 0.960 & 1.000 & 0.898 \\
\rowcolor{RowBlue}
\multicolumn{7}{l}{\textit{Robotics-specific}} \\
Cosmos 2.5 & 0.864 & 0.878 & 0.993 & 1.000 & 0.972 & 0.768 \\
DreamGen(gr1) & 0.878 & 0.878 & 0.957 & 0.963 & 0.969 & 0.848 \\
DreamGen(droid) & 0.763 & 0.819 & 0.958 & 0.951 & 0.930 & 0.591 \\
UnifoLM-WMA-0 & 0.044 & 0.044 & 0.507 & 0.477 & 0.301 & 0.025 \\
Vidar & 0.083 & 0.097 & 0.743 & 0.736 & 0.583 & 0.082 \\
\midrule
\textbf{Mean} & \textbf{0.714} & \textbf{0.724} & \textbf{0.933} & \textbf{0.928} & \textbf{0.910} & \textbf{0.657} \\
\bottomrule
\end{tabular}
\end{minipage}

\end{table*}
\vspace{-2mm}

% ============================================================ 007
\begin{table*}[t]
\centering

% ===== Left Table =====
\begin{minipage}{0.48\textwidth}
\centering
\scriptsize
\setlength{\tabcolsep}{1.5pt}
\renewcommand{\arraystretch}{1.07}

\caption{\textbf{Results on Spatial Relationship with Qwen}}
\label{tab:spatial_relationship_qwen}
\begin{tabular}{lcccccc}
\toprule
\textbf{Model} & \textbf{SRS} & \textbf{MFS} & \textbf{OCS} & \textbf{RCS} & \textbf{PSS} & \textbf{TS} \\
\midrule
\rowcolor{RowBlue}
\multicolumn{7}{l}{\textit{Open-source}} \\
Wan2.2\_A14B & 0.833 & 0.833 & 1.000 & 1.000 & 1.000 & 0.660 \\
LongCat-Video & 0.636 & 0.636 & 1.000 & 1.000 & 1.000 & 0.465 \\
Wan2.2\_5B & 0.600 & 0.625 & 1.000 & 0.900 & 0.900 & 0.402 \\
Wan2.1\_14B & 0.636 & 0.636 & 1.000 & 1.000 & 1.000 & 0.400 \\
Skyreels & 0.750 & 0.750 & 1.000 & 0.875 & 0.875 & 0.400 \\
LTX-Video & 0.750 & 0.750 & 1.000 & 1.000 & 1.000 & 0.382 \\
FramePack & 0.400 & 0.400 & 1.000 & 1.000 & 1.000 & 0.240 \\
CogVideoX-5B & 0.500 & 0.535 & 1.000 & 0.857 & 0.714 & 0.240 \\
\rowcolor{RowBlue}
\multicolumn{7}{l}{\textit{Closed-source}} \\
Wan 2.5 & 0.916 & 0.916 & 1.000 & 0.979 & 1.000 & 0.825 \\
Hailuo v2 & 1.000 & 1.000 & 1.000 & 1.000 & 1.000 & 0.840 \\
Veo 3 & 0.933 & 0.933 & 1.000 & 1.000 & 1.000 & 0.740 \\
Seedance 1.0 & 0.666 & 0.666 & 1.000 & 1.000 & 1.000 & 0.665 \\
\rowcolor{RowBlue}
\multicolumn{7}{l}{\textit{Robotics-specific}} \\
Cosmos 2.5 & 0.875 & 0.875 & 1.000 & 1.000 & 1.000 & 0.512 \\
DreamGen(gr1) & 0.642 & 0.642 & 1.000 & 1.000 & 1.000 & 0.500 \\
DreamGen(droid) & 0.545 & 0.545 & 1.000 & 1.000 & 1.000 & 0.505 \\
UnifoLM-WMA-0 & 0.200 & 0.250 & 0.625 & 0.575 & 0.500 & 0.065 \\
Vidar & 0.367 & 0.382 & 0.691 & 0.647 & 0.573 & 0.140 \\
\midrule
\textbf{Mean} & \textbf{0.666} & \textbf{0.674} & \textbf{0.962} & \textbf{0.935} & \textbf{0.920} & \textbf{0.464} \\
\bottomrule
\end{tabular}
\end{minipage}
\hfill
% ===== Right Table =====
\begin{minipage}{0.48\textwidth}
\centering
\scriptsize
\setlength{\tabcolsep}{1.5pt}
\renewcommand{\arraystretch}{1.07}

\caption{\textbf{Results on Visual Reasoning with Qwen}}
\label{tab:visual_reasoning_qwen}
\begin{tabular}{lcccccc}
\toprule
\textbf{Model} & \textbf{AES} & \textbf{VRS} & \textbf{OCS} & \textbf{RCS} & \textbf{PSS} & \textbf{TS} \\
\midrule
\rowcolor{RowBlue}
\multicolumn{7}{l}{\textit{Open-source}} \\
Wan2.2\_A14B & 0.727 & 0.701 & 1.000 & 1.000 & 1.000 & 0.550 \\
LongCat-Video & 0.361 & 0.358 & 0.944 & 0.958 & 0.847 & 0.354 \\
Wan2.2\_5B & 0.291 & 0.218 & 0.906 & 0.906 & 0.906 & 0.420 \\
Wan2.1\_14B & 0.397 & 0.401 & 0.886 & 0.897 & 0.784 & 0.269 \\
Skyreels & 0.420 & 0.401 & 0.960 & 0.930 & 0.830 & 0.357 \\
LTX-Video & 0.390 & 0.410 & 0.859 & 0.937 & 0.859 & 0.285 \\
FramePack & 0.350 & 0.343 & 0.980 & 0.990 & 0.950 & 0.345 \\
CogVideoX-5B & 0.073 & 0.044 & 0.779 & 0.632 & 0.544 & 0.030 \\
\rowcolor{RowBlue}
\multicolumn{7}{l}{\textit{Closed-source}} \\
Wan 2.5 & 0.809 & 0.770 & 0.988 & 1.000 & 0.952 & 0.737 \\
Hailuo v2 & 0.790 & 0.882 & 1.000 & 1.000 & 0.900 & 0.820 \\
Veo 3 & 0.847 & 0.853 & 0.945 & 1.000 & 0.989 & 0.750 \\
Seedance 1.0 & 0.927 & 0.945 & 0.979 & 0.989 & 1.000 & 0.789 \\
\rowcolor{RowBlue}
\multicolumn{7}{l}{\textit{Robotics-specific}} \\
Cosmos 2.5 & 0.593 & 0.632 & 0.984 & 1.000 & 1.000 & 0.506 \\
DreamGen(gr1) & 0.437 & 0.425 & 0.937 & 0.958 & 0.822 & 0.404 \\
DreamGen(droid) & 0.600 & 0.575 & 0.800 & 0.825 & 0.787 & 0.386 \\
UnifoLM-WMA-0 & 0.000 & -0.098 & 0.177 & 0.348 & 0.289 & 0.000 \\
Vidar & 0.000 & -0.062 & 0.947 & 0.937 & 0.906 & 0.029 \\
\midrule
\textbf{Mean} & \textbf{0.452} & \textbf{0.438} & \textbf{0.890} & \textbf{0.882} & \textbf{0.825} & \textbf{0.395} \\
\bottomrule
\end{tabular}
\end{minipage}

\end{table*}
\vspace{-2mm}

% ============================================================ 008
\begin{table*}[t]
\centering

% ===== Left Table =====
\begin{minipage}{0.44\textwidth}
\centering
\scriptsize
\setlength{\tabcolsep}{1pt}
\renewcommand{\arraystretch}{1.07}

\caption{\textbf{Results on Dual Arm Robot with Qwen}}
\label{tab:dual_arm_qwen}
\begin{tabular}{lcccccccc}
\toprule
\textbf{Model} & \textbf{PSS} & \textbf{TAC} & \textbf{RSS} & \textbf{MS} & \textbf{MA} & \textbf{TC} & \textbf{VQ} & \textbf{TS} \\
\midrule
\rowcolor{RowBlue}
\multicolumn{9}{l}{\textit{Open-source}} \\
Wan2.2\_A14B & 0.852 & 0.760 & 0.767 & 0.915 & 0.204 & 0.806 & 0.550 & 0.678 \\
LongCat-Video & 0.738 & 0.638 & 0.741 & 0.937 & 0.244 & 0.688 & 0.517 & 0.602 \\
Wan2.2\_5B & 0.688 & 0.645 & 0.658 & 0.940 & 0.269 & 0.666 & 0.402 & 0.534 \\
Wan2.1\_14B & 0.790 & 0.730 & 0.639 & 0.850 & 0.261 & 0.760 & 0.364 & 0.562 \\
Skyreels & 0.755 & 0.700 & 0.675 & 0.884 & 0.252 & 0.728 & 0.419 & 0.573 \\
LTX-Video & 0.615 & 0.535 & 0.692 & 0.812 & 0.143 & 0.575 & 0.399 & 0.487 \\
FramePack & 0.612 & 0.495 & 0.719 & 0.885 & 0.103 & 0.554 & 0.445 & 0.499 \\
CogVideoX-5B & 0.505 & 0.540 & 0.613 & 0.752 & 0.143 & 0.522 & 0.323 & 0.422 \\
\rowcolor{RowBlue}
\multicolumn{9}{l}{\textit{Closed-source}} \\
Wan 2.5 & 0.920 & 0.880 & 0.761 & 0.970 & 0.347 & 0.900 & 0.588 & 0.744 \\
Hailuo v2 & 0.908 & 0.848 & 0.744 & 0.983 & 0.312 & 0.878 & 0.534 & 0.706 \\
Veo 3 & 0.870 & 0.802 & 0.777 & 0.973 & 0.262 & 0.836 & 0.581 & 0.708 \\
Seedance 1.0 & 0.895 & 0.810 & 0.801 & 0.972 & 0.294 & 0.852 & 0.608 & 0.730 \\
\rowcolor{RowBlue}
\multicolumn{9}{l}{\textit{Robotics-specific}} \\
Cosmos 2.5 & 0.792 & 0.708 & 0.791 & 0.930 & 0.127 & 0.750 & 0.543 & 0.646 \\
DreamGen(gr1) & 0.780 & 0.638 & 0.801 & 0.939 & 0.123 & 0.709 & 0.555 & 0.632 \\
DreamGen(droid) & 0.722 & 0.678 & 0.711 & 0.863 & 0.201 & 0.700 & 0.441 & 0.570 \\
UnifoLM-WMA-0 & 0.110 & 0.222 & 0.266 & 0.497 & 0.120 & 0.166 & 0.046 & 0.106 \\
Vidar & 0.295 & 0.240 & 0.804 & 0.933 & 0.025 & 0.268 & 0.511 & 0.389 \\
\midrule
\textbf{Mean} & \textbf{0.685} & \textbf{0.622} & \textbf{0.694} & \textbf{0.880} & \textbf{0.205} & \textbf{0.651} & \textbf{0.449} & \textbf{0.550} \\
\bottomrule
\end{tabular}
\end{minipage}
\hfill
% ===== Right Table =====
\begin{minipage}{0.44\textwidth}
\centering
\scriptsize
\setlength{\tabcolsep}{1pt}
\renewcommand{\arraystretch}{1.07}

\caption{\textbf{Results on Humanoid Robot with Qwen}}
\label{tab:humanoid_qwen}
\begin{tabular}{lcccccccc}
\toprule
\textbf{Model} & \textbf{PSS} & \textbf{TAC} & \textbf{RSS} & \textbf{MS} & \textbf{MA} & \textbf{TC} & \textbf{VQ} & \textbf{TS} \\
\midrule
\rowcolor{RowBlue}
\multicolumn{9}{l}{\textit{Open-source}} \\
Wan2.2\_A14B & 0.935 & 0.800 & 0.806 & 0.966 & 0.105 & 0.898 & 0.557 & 0.727 \\
LongCat-Video & 0.918 & 0.765 & 0.826 & 0.969 & 0.089 & 0.841 & 0.579 & 0.710 \\
Wan2.2\_5B & 0.880 & 0.758 & 0.791 & 0.963 & 0.152 & 0.819 & 0.544 & 0.681 \\
Wan2.1\_14B & 0.895 & 0.712 & 0.785 & 0.938 & 0.167 & 0.804 & 0.524 & 0.664 \\
Skyreels & 0.842 & 0.670 & 0.803 & 0.808 & 0.086 & 0.756 & 0.500 & 0.628 \\
LTX-Video & 0.852 & 0.670 & 0.808 & 0.628 & 0.050 & 0.761 & 0.445 & 0.603 \\
FramePack & 0.815 & 0.622 & 0.838 & 0.864 & 0.069 & 0.719 & 0.550 & 0.634 \\
CogVideoX-5B & 0.712 & 0.602 & 0.710 & 0.647 & 0.127 & 0.658 & 0.390 & 0.524 \\
\rowcolor{RowBlue}
\multicolumn{9}{l}{\textit{Closed-source}} \\
Wan 2.5 & 0.935 & 0.835 & 0.826 & 0.981 & 0.122 & 0.885 & 0.600 & 0.742 \\
Hailuo v2 & 0.952 & 0.848 & 0.796 & 0.970 & 0.133 & 0.891 & 0.548 & 0.719 \\
Veo 3 & 0.955 & 0.802 & 0.831 & 0.968 & 0.132 & 0.829 & 0.604 & 0.716 \\
Seedance 1.0 & 0.936 & 0.814 & 0.829 & 0.964 & 0.166 & 0.904 & 0.614 & 0.759 \\
\rowcolor{RowBlue}
\multicolumn{9}{l}{\textit{Robotics-specific}} \\
Cosmos 2.5 & 0.868 & 0.730 & 0.841 & 0.925 & 0.071 & 0.799 & 0.578 & 0.688 \\
DreamGen(gr1) & 0.858 & 0.682 & 0.823 & 0.885 & 0.079 & 0.770 & 0.542 & 0.656 \\
DreamGen(droid) & 0.842 & 0.736 & 0.739 & 0.843 & 0.137 & 0.788 & 0.478 & 0.633 \\
UnifoLM-WMA-0 & 0.168 & 0.285 & 0.349 & 0.512 & 0.112 & 0.226 & 0.115 & 0.170 \\
Vidar & 0.440 & 0.305 & 0.700 & 0.855 & 0.025 & 0.372 & 0.357 & 0.364 \\
\midrule
\textbf{Mean} & \textbf{0.806} & \textbf{0.674} & \textbf{0.770} & \textbf{0.862} & \textbf{0.113} & \textbf{0.740} & \textbf{0.501} & \textbf{0.620} \\
\bottomrule
\end{tabular}
\end{minipage}

\end{table*}
\vspace{-2mm}

% ============================================================ 009
\begin{table*}[t]
\centering

% ===== Left Table =====
\begin{minipage}{0.44\textwidth}
\centering
\scriptsize
\setlength{\tabcolsep}{1pt}
\renewcommand{\arraystretch}{1.07}

\caption{\textbf{Results on Single Arm Robot with Qwen}}
\label{tab:single_arm_qwen}
\begin{tabular}{lcccccccc}
\toprule
\textbf{Model} & \textbf{PSS} & \textbf{TAC} & \textbf{RSS} & \textbf{MS} & \textbf{MA} & \textbf{TC} & \textbf{VQ} & \textbf{TS} \\
\midrule
\rowcolor{RowBlue}
\multicolumn{9}{l}{\textit{Open-source}} \\
Wan2.2\_A14B & 0.815 & 0.802 & 0.755 & 0.942 & 0.263 & 0.809 & 0.568 & 0.688 \\
LongCat-Video & 0.750 & 0.712 & 0.812 & 0.945 & 0.298 & 0.736 & 0.660 & 0.698 \\
Wan2.2\_5B & 0.605 & 0.650 & 0.626 & 0.943 & 0.313 & 0.628 & 0.394 & 0.511 \\
Wan2.1\_14B & 0.700 & 0.616 & 0.638 & 0.849 & 0.282 & 0.659 & 0.379 & 0.519 \\
Skyreels & 0.760 & 0.699 & 0.731 & 0.911 & 0.286 & 0.732 & 0.492 & 0.612 \\
LTX-Video & 0.535 & 0.525 & 0.710 & 0.804 & 0.158 & 0.530 & 0.443 & 0.486 \\
FramePack & 0.400 & 0.500 & 0.674 & 0.888 & 0.104 & 0.394 & 0.413 & 0.403 \\
CogVideoX-5B & 0.460 & 0.460 & 0.553 & 0.815 & 0.256 & 0.460 & 0.288 & 0.374 \\
\rowcolor{RowBlue}
\multicolumn{9}{l}{\textit{Closed-source}} \\
Wan 2.5 & 0.895 & 0.918 & 0.747 & 0.969 & 0.412 & 0.895 & 0.589 & 0.742 \\
Hailuo v2 & 0.910 & 0.906 & 0.705 & 0.989 & 0.396 & 0.902 & 0.489 & 0.695 \\
Veo 3 & 0.898 & 0.891 & 0.756 & 0.977 & 0.362 & 0.890 & 0.594 & 0.742 \\
Seedance 1.0 & 0.852 & 0.818 & 0.747 & 0.967 & 0.306 & 0.835 & 0.580 & 0.707 \\
\rowcolor{RowBlue}
\multicolumn{9}{l}{\textit{Robotics-specific}} \\
Cosmos 2.5 & 0.795 & 0.765 & 0.751 & 0.888 & 0.206 & 0.776 & 0.516 & 0.646 \\
DreamGen(gr1) & 0.812 & 0.728 & 0.782 & 0.932 & 0.263 & 0.770 & 0.550 & 0.660 \\
DreamGen(droid) & 0.745 & 0.675 & 0.721 & 0.930 & 0.214 & 0.710 & 0.468 & 0.589 \\
UnifoLM-WMA-0 & 0.382 & 0.495 & 0.391 & 0.709 & 0.389 & 0.439 & 0.140 & 0.289 \\
Vidar & 0.220 & 0.200 & 0.736 & 0.929 & 0.068 & 0.210 & 0.479 & 0.344 \\
\midrule
\textbf{Mean} & \textbf{0.658} & \textbf{0.648} & \textbf{0.679} & \textbf{0.898} & \textbf{0.263} & \textbf{0.649} & \textbf{0.454} & \textbf{0.551} \\
\bottomrule
\end{tabular}
\end{minipage}
\hfill
% ===== Right Table =====
\begin{minipage}{0.44\textwidth}
\centering
\scriptsize
\setlength{\tabcolsep}{1pt}
\renewcommand{\arraystretch}{1.07}

\caption{\textbf{Results on Quadruped Robot with Qwen}}
\label{tab:quad_qwen}
\begin{tabular}{lcccccccc}
\toprule
\textbf{Model} & \textbf{PSS} & \textbf{TAC} & \textbf{RSS} & \textbf{MS} & \textbf{MA} & \textbf{TC} & \textbf{VQ} & \textbf{TS} \\
\midrule
\rowcolor{RowBlue}
\multicolumn{9}{l}{\textit{Open-source}} \\
Wan2.2\_A14B & 0.860 & 0.698 & 0.746 & 0.888 & 0.196 & 0.779 & 0.561 & 0.670 \\
LongCat-Video & 0.870 & 0.685 & 0.747 & 0.923 & 0.179 & 0.778 & 0.554 & 0.666 \\
Wan2.2\_5B & 0.858 & 0.635 & 0.721 & 0.921 & 0.273 & 0.746 & 0.529 & 0.637 \\
Wan2.1\_14B & 0.845 & 0.665 & 0.674 & 0.850 & 0.303 & 0.755 & 0.453 & 0.604 \\
Skyreels & 0.862 & 0.652 & 0.749 & 0.853 & 0.163 & 0.758 & 0.550 & 0.654 \\
LTX-Video & 0.818 & 0.690 & 0.709 & 0.676 & 0.122 & 0.754 & 0.423 & 0.588 \\
FramePack & 0.765 & 0.540 & 0.881 & 0.876 & 0.069 & 0.652 & 0.687 & 0.669 \\
CogVideoX-5B & 0.705 & 0.575 & 0.639 & 0.618 & 0.220 & 0.640 & 0.349 & 0.494 \\
\rowcolor{RowBlue}
\multicolumn{9}{l}{\textit{Closed-source}} \\
Wan 2.5 & 0.902 & 0.740 & 0.769 & 0.948 & 0.322 & 0.821 & 0.601 & 0.711 \\
Hailuo v2 & 0.888 & 0.722 & 0.654 & 0.961 & 0.354 & 0.805 & 0.458 & 0.631 \\
Veo 3 & 0.880 & 0.722 & 0.793 & 0.961 & 0.214 & 0.801 & 0.652 & 0.726 \\
Seedance 1.0 & 0.865 & 0.710 & 0.728 & 0.945 & 0.334 & 0.788 & 0.560 & 0.674 \\
\rowcolor{RowBlue}
\multicolumn{9}{l}{\textit{Robotics-specific}} \\
Cosmos 2.5 & 0.850 & 0.612 & 0.755 & 0.892 & 0.137 & 0.731 & 0.547 & 0.639 \\
DreamGen(gr1) & 0.852 & 0.672 & 0.701 & 0.788 & 0.164 & 0.762 & 0.460 & 0.611 \\
DreamGen(droid) & 0.745 & 0.592 & 0.724 & 0.854 & 0.160 & 0.669 & 0.500 & 0.584 \\
UnifoLM-WMA-0 & 0.310 & 0.300 & 0.399 & 0.497 & 0.132 & 0.305 & 0.197 & 0.251 \\
Vidar & 0.495 & 0.445 & 0.575 & 0.749 & 0.074 & 0.470 & 0.290 & 0.380 \\
\midrule
\textbf{Mean} & \textbf{0.782} & \textbf{0.612} & \textbf{0.705} & \textbf{0.825} & \textbf{0.198} & \textbf{0.701} & \textbf{0.494} & \textbf{0.597} \\
\bottomrule
\end{tabular}
\end{minipage}

\end{table*}
\vspace{-2mm}

\end{document}